\documentclass[11pt]{article}
\usepackage[margin=1in]{geometry}
\usepackage[T1]{fontenc}
\usepackage[utf8]{inputenc}
\usepackage{lmodern}
\usepackage{amsmath,amssymb,booktabs,microtype,url,xcolor,array,seqsplit}
\usepackage{caption}
\usepackage{placeins}
\usepackage{needspace}
\usepackage{listings}
\usepackage{pgfplots}\pgfplotsset{compat=1.18}
\usepackage{natbib}
\usepackage[hidelinks]{hyperref}
\makeatletter\g@addto@macro\UrlBreaks{\do\-}\makeatother   %
\hypersetup{pdftitle={Lessons from Luck: Do LLM-Written Lessons Carry Chance into an Agent's Next Decision?},
  pdfauthor={Kazuki Nakayashiki},
  pdfkeywords={agent memory, self-improving agents, outcome bias, LLM agents, registered study}}
\makeatletter
\newcommand{\V}[1]{\ifcsname v@#1\endcsname\csname v@#1\endcsname\else\errmessage{Undefined value: #1}\fi}
\makeatother
\newcommand{\lit}[1]{#1}

\expandafter\def\csname v@avail.pthree.full\endcsname{11}
\expandafter\def\csname v@avail.pthree.fullworlds\endcsname{254}
\expandafter\def\csname v@avail.pthree.less\endcsname{10}
\expandafter\def\csname v@avail.pthree.lessworlds\endcsname{2}
\expandafter\def\csname v@avail.ptwo.full\endcsname{11}
\expandafter\def\csname v@avail.ptwo.fullworlds\endcsname{252}
\expandafter\def\csname v@avail.ptwo.less\endcsname{10}
\expandafter\def\csname v@avail.ptwo.lessworlds\endcsname{4}
\expandafter\def\csname v@chain.block\endcsname{970581}
\expandafter\def\csname v@chain.blocktime\endcsname{2026-10-09 04:23:19 UTC}
\expandafter\def\csname v@chain.depositedat\endcsname{2026-10-09 04:03:33 UTC}
\expandafter\def\csname v@chain.doi\endcsname{10.5281/zenodo.23253562}
\expandafter\def\csname v@chain.lockfiles\endcsname{123,907}
\expandafter\def\csname v@chain.manifest\endcsname{\texttt{\seqsplit{0fc1f680ef7d241692e7726dc9dd5a7df45ea45905e6f13bb295255e2fe864be}}}
\expandafter\def\csname v@chain.packagefiles\endcsname{74}
\expandafter\def\csname v@design.capconfirm\endcsname{800}
\expandafter\def\csname v@design.capsmoke\endcsname{60}
\expandafter\def\csname v@design.endpoints\endcsname{11}
\expandafter\def\csname v@design.panelmin\endcsname{6}
\expandafter\def\csname v@design.panelnames\endcsname{Claude Opus 5.5, Claude Sonnet 5.5, Claude Haiku 5.5, GPT-6.1 Sol, GPT-6 Luna, Gemini 3.8 Flash, Grok 4.7, Qwen 3.8 Max, DeepSeek V4 Pro, Mistral Large 4 and Kimi K3}
\expandafter\def\csname v@design.readersperworld\endcsname{32}
\expandafter\def\csname v@design.seed\endcsname{20261009}
\expandafter\def\csname v@design.smokeworlds\endcsname{2}
\expandafter\def\csname v@design.worlds\endcsname{256}
\expandafter\def\csname v@design.worldshalf\endcsname{128}
\expandafter\def\csname v@design.writerrequests\endcsname{33,792}
\expandafter\def\csname v@ep.deepseek.a2rb.ci\endcsname{[$+0.222$, $+0.333$]}
\expandafter\def\csname v@ep.deepseek.a2rb.lower\endcsname{$+0.277$}
\expandafter\def\csname v@ep.deepseek.a2rb.lowerci\endcsname{[$+0.222$, $+0.333$]}
\expandafter\def\csname v@ep.deepseek.a2rb.mean\endcsname{$+0.277$}
\expandafter\def\csname v@ep.deepseek.a2rb.n\endcsname{256}
\expandafter\def\csname v@ep.deepseek.a2rb.pbraw\endcsname{$5.6\times10^{-18}$}
\expandafter\def\csname v@ep.deepseek.a2rb.praw\endcsname{$9.7\times10^{-20}$}
\expandafter\def\csname v@ep.deepseek.a2rb.upper\endcsname{$+0.277$}
\expandafter\def\csname v@ep.deepseek.a2rb.upperci\endcsname{[$+0.222$, $+0.333$]}
\expandafter\def\csname v@ep.deepseek.a3rb.ci\endcsname{[$+0.010$, $+0.053$]}
\expandafter\def\csname v@ep.deepseek.a3rb.lower\endcsname{$+0.031$}
\expandafter\def\csname v@ep.deepseek.a3rb.lowerci\endcsname{[$+0.010$, $+0.053$]}
\expandafter\def\csname v@ep.deepseek.a3rb.mean\endcsname{$+0.031$}
\expandafter\def\csname v@ep.deepseek.a3rb.n\endcsname{256}
\expandafter\def\csname v@ep.deepseek.a3rb.pbraw\endcsname{$0.011$}
\expandafter\def\csname v@ep.deepseek.a3rb.praw\endcsname{$0.0045$}
\expandafter\def\csname v@ep.deepseek.a3rb.upper\endcsname{$+0.031$}
\expandafter\def\csname v@ep.deepseek.a3rb.upperci\endcsname{[$+0.010$, $+0.053$]}
\expandafter\def\csname v@ep.deepseek.count.bad.failure.n\endcsname{0/256}
\expandafter\def\csname v@ep.deepseek.count.bad.failure.neutral\endcsname{0/256}
\expandafter\def\csname v@ep.deepseek.count.bad.failure.nx\endcsname{1/256}
\expandafter\def\csname v@ep.deepseek.count.bad.failure.planted\endcsname{11/256}
\expandafter\def\csname v@ep.deepseek.count.bad.failure.rb\endcsname{0/256}
\expandafter\def\csname v@ep.deepseek.count.bad.failure.rbd\endcsname{0/256}
\expandafter\def\csname v@ep.deepseek.count.bad.failure.rbdx\endcsname{0/256}
\expandafter\def\csname v@ep.deepseek.count.bad.failure.rbx\endcsname{5/256}
\expandafter\def\csname v@ep.deepseek.count.bad.success.n\endcsname{2/256}
\expandafter\def\csname v@ep.deepseek.count.bad.success.neutral\endcsname{0/256}
\expandafter\def\csname v@ep.deepseek.count.bad.success.nx\endcsname{0/256}
\expandafter\def\csname v@ep.deepseek.count.bad.success.planted\endcsname{53/256}
\expandafter\def\csname v@ep.deepseek.count.bad.success.rb\endcsname{8/256}
\expandafter\def\csname v@ep.deepseek.count.bad.success.rbd\endcsname{0/256}
\expandafter\def\csname v@ep.deepseek.count.bad.success.rbdx\endcsname{0/256}
\expandafter\def\csname v@ep.deepseek.count.bad.success.rbx\endcsname{0/256}
\expandafter\def\csname v@ep.deepseek.count.good.failure.n\endcsname{8/256}
\expandafter\def\csname v@ep.deepseek.count.good.failure.neutral\endcsname{0/256}
\expandafter\def\csname v@ep.deepseek.count.good.failure.nx\endcsname{0/256}
\expandafter\def\csname v@ep.deepseek.count.good.failure.planted\endcsname{128/256}
\expandafter\def\csname v@ep.deepseek.count.good.failure.rb\endcsname{71/256}
\expandafter\def\csname v@ep.deepseek.count.good.failure.rbd\endcsname{41/256}
\expandafter\def\csname v@ep.deepseek.count.good.failure.rbdx\endcsname{0/256}
\expandafter\def\csname v@ep.deepseek.count.good.failure.rbx\endcsname{0/256}
\expandafter\def\csname v@ep.deepseek.count.good.success.n\endcsname{0/256}
\expandafter\def\csname v@ep.deepseek.count.good.success.neutral\endcsname{0/256}
\expandafter\def\csname v@ep.deepseek.count.good.success.nx\endcsname{1/256}
\expandafter\def\csname v@ep.deepseek.count.good.success.planted\endcsname{23/256}
\expandafter\def\csname v@ep.deepseek.count.good.success.rb\endcsname{0/256}
\expandafter\def\csname v@ep.deepseek.count.good.success.rbd\endcsname{0/256}
\expandafter\def\csname v@ep.deepseek.count.good.success.rbdx\endcsname{0/256}
\expandafter\def\csname v@ep.deepseek.count.good.success.rbx\endcsname{0/256}
\expandafter\def\csname v@ep.deepseek.id\endcsname{\texttt{deepseek/deepseek-v4-pro-0813}}
\expandafter\def\csname v@ep.deepseek.k.bad.failure.n\endcsname{0}
\expandafter\def\csname v@ep.deepseek.k.bad.failure.neutral\endcsname{0}
\expandafter\def\csname v@ep.deepseek.k.bad.failure.nx\endcsname{1}
\expandafter\def\csname v@ep.deepseek.k.bad.failure.planted\endcsname{11}
\expandafter\def\csname v@ep.deepseek.k.bad.failure.rb\endcsname{0}
\expandafter\def\csname v@ep.deepseek.k.bad.failure.rbd\endcsname{0}
\expandafter\def\csname v@ep.deepseek.k.bad.failure.rbdx\endcsname{0}
\expandafter\def\csname v@ep.deepseek.k.bad.failure.rbx\endcsname{5}
\expandafter\def\csname v@ep.deepseek.k.bad.success.n\endcsname{2}
\expandafter\def\csname v@ep.deepseek.k.bad.success.neutral\endcsname{0}
\expandafter\def\csname v@ep.deepseek.k.bad.success.nx\endcsname{0}
\expandafter\def\csname v@ep.deepseek.k.bad.success.planted\endcsname{53}
\expandafter\def\csname v@ep.deepseek.k.bad.success.rb\endcsname{8}
\expandafter\def\csname v@ep.deepseek.k.bad.success.rbd\endcsname{0}
\expandafter\def\csname v@ep.deepseek.k.bad.success.rbdx\endcsname{0}
\expandafter\def\csname v@ep.deepseek.k.bad.success.rbx\endcsname{0}
\expandafter\def\csname v@ep.deepseek.k.good.failure.n\endcsname{8}
\expandafter\def\csname v@ep.deepseek.k.good.failure.neutral\endcsname{0}
\expandafter\def\csname v@ep.deepseek.k.good.failure.nx\endcsname{0}
\expandafter\def\csname v@ep.deepseek.k.good.failure.planted\endcsname{128}
\expandafter\def\csname v@ep.deepseek.k.good.failure.rb\endcsname{71}
\expandafter\def\csname v@ep.deepseek.k.good.failure.rbd\endcsname{41}
\expandafter\def\csname v@ep.deepseek.k.good.failure.rbdx\endcsname{0}
\expandafter\def\csname v@ep.deepseek.k.good.failure.rbx\endcsname{0}
\expandafter\def\csname v@ep.deepseek.k.good.success.n\endcsname{0}
\expandafter\def\csname v@ep.deepseek.k.good.success.neutral\endcsname{0}
\expandafter\def\csname v@ep.deepseek.k.good.success.nx\endcsname{1}
\expandafter\def\csname v@ep.deepseek.k.good.success.planted\endcsname{23}
\expandafter\def\csname v@ep.deepseek.k.good.success.rb\endcsname{0}
\expandafter\def\csname v@ep.deepseek.k.good.success.rbd\endcsname{0}
\expandafter\def\csname v@ep.deepseek.k.good.success.rbdx\endcsname{0}
\expandafter\def\csname v@ep.deepseek.k.good.success.rbx\endcsname{0}
\expandafter\def\csname v@ep.deepseek.l2brb.ci\endcsname{[$-0.014$, $+0.014$]}
\expandafter\def\csname v@ep.deepseek.l2brb.mean\endcsname{$0.000$}
\expandafter\def\csname v@ep.deepseek.l2brb.n\endcsname{256}
\expandafter\def\csname v@ep.deepseek.l2brb.pbraw\endcsname{$1.000$}
\expandafter\def\csname v@ep.deepseek.l2brb.praw\endcsname{$1.000$}
\expandafter\def\csname v@ep.deepseek.l2rb.ci\endcsname{[$+0.222$, $+0.333$]}
\expandafter\def\csname v@ep.deepseek.l2rb.lower\endcsname{$+0.277$}
\expandafter\def\csname v@ep.deepseek.l2rb.lowerci\endcsname{[$+0.222$, $+0.333$]}
\expandafter\def\csname v@ep.deepseek.l2rb.mean\endcsname{$+0.277$}
\expandafter\def\csname v@ep.deepseek.l2rb.n\endcsname{256}
\expandafter\def\csname v@ep.deepseek.l2rb.pbraw\endcsname{$5.6\times10^{-18}$}
\expandafter\def\csname v@ep.deepseek.l2rb.praw\endcsname{$9.7\times10^{-20}$}
\expandafter\def\csname v@ep.deepseek.l2rb.upper\endcsname{$+0.277$}
\expandafter\def\csname v@ep.deepseek.l2rb.upperci\endcsname{[$+0.222$, $+0.333$]}
\expandafter\def\csname v@ep.deepseek.l3rb.ci\endcsname{[$+0.010$, $+0.053$]}
\expandafter\def\csname v@ep.deepseek.l3rb.lower\endcsname{$+0.031$}
\expandafter\def\csname v@ep.deepseek.l3rb.lowerci\endcsname{[$+0.010$, $+0.053$]}
\expandafter\def\csname v@ep.deepseek.l3rb.mean\endcsname{$+0.031$}
\expandafter\def\csname v@ep.deepseek.l3rb.n\endcsname{256}
\expandafter\def\csname v@ep.deepseek.l3rb.pbraw\endcsname{$0.011$}
\expandafter\def\csname v@ep.deepseek.l3rb.praw\endcsname{$0.0045$}
\expandafter\def\csname v@ep.deepseek.l3rb.upper\endcsname{$+0.031$}
\expandafter\def\csname v@ep.deepseek.l3rb.upperci\endcsname{[$+0.010$, $+0.053$]}
\expandafter\def\csname v@ep.deepseek.n.bad.failure.n\endcsname{256}
\expandafter\def\csname v@ep.deepseek.n.bad.failure.neutral\endcsname{256}
\expandafter\def\csname v@ep.deepseek.n.bad.failure.nx\endcsname{256}
\expandafter\def\csname v@ep.deepseek.n.bad.failure.planted\endcsname{256}
\expandafter\def\csname v@ep.deepseek.n.bad.failure.rb\endcsname{256}
\expandafter\def\csname v@ep.deepseek.n.bad.failure.rbd\endcsname{256}
\expandafter\def\csname v@ep.deepseek.n.bad.failure.rbdx\endcsname{256}
\expandafter\def\csname v@ep.deepseek.n.bad.failure.rbx\endcsname{256}
\expandafter\def\csname v@ep.deepseek.n.bad.success.n\endcsname{256}
\expandafter\def\csname v@ep.deepseek.n.bad.success.neutral\endcsname{256}
\expandafter\def\csname v@ep.deepseek.n.bad.success.nx\endcsname{256}
\expandafter\def\csname v@ep.deepseek.n.bad.success.planted\endcsname{256}
\expandafter\def\csname v@ep.deepseek.n.bad.success.rb\endcsname{256}
\expandafter\def\csname v@ep.deepseek.n.bad.success.rbd\endcsname{256}
\expandafter\def\csname v@ep.deepseek.n.bad.success.rbdx\endcsname{256}
\expandafter\def\csname v@ep.deepseek.n.bad.success.rbx\endcsname{256}
\expandafter\def\csname v@ep.deepseek.n.good.failure.n\endcsname{256}
\expandafter\def\csname v@ep.deepseek.n.good.failure.neutral\endcsname{256}
\expandafter\def\csname v@ep.deepseek.n.good.failure.nx\endcsname{256}
\expandafter\def\csname v@ep.deepseek.n.good.failure.planted\endcsname{256}
\expandafter\def\csname v@ep.deepseek.n.good.failure.rb\endcsname{256}
\expandafter\def\csname v@ep.deepseek.n.good.failure.rbd\endcsname{256}
\expandafter\def\csname v@ep.deepseek.n.good.failure.rbdx\endcsname{256}
\expandafter\def\csname v@ep.deepseek.n.good.failure.rbx\endcsname{256}
\expandafter\def\csname v@ep.deepseek.n.good.success.n\endcsname{256}
\expandafter\def\csname v@ep.deepseek.n.good.success.neutral\endcsname{256}
\expandafter\def\csname v@ep.deepseek.n.good.success.nx\endcsname{256}
\expandafter\def\csname v@ep.deepseek.n.good.success.planted\endcsname{256}
\expandafter\def\csname v@ep.deepseek.n.good.success.rb\endcsname{256}
\expandafter\def\csname v@ep.deepseek.n.good.success.rbd\endcsname{256}
\expandafter\def\csname v@ep.deepseek.n.good.success.rbdx\endcsname{256}
\expandafter\def\csname v@ep.deepseek.n.good.success.rbx\endcsname{256}
\expandafter\def\csname v@ep.deepseek.name\endcsname{DeepSeek V4 Pro}
\expandafter\def\csname v@ep.deepseek.namesworse\endcsname{38\%}
\expandafter\def\csname v@ep.deepseek.p2rb.ci\endcsname{[$+0.222$, $+0.333$]}
\expandafter\def\csname v@ep.deepseek.p2rb.lower\endcsname{$+0.277$}
\expandafter\def\csname v@ep.deepseek.p2rb.lowerci\endcsname{[$+0.222$, $+0.333$]}
\expandafter\def\csname v@ep.deepseek.p2rb.mean\endcsname{$+0.277$}
\expandafter\def\csname v@ep.deepseek.p2rb.n\endcsname{256}
\expandafter\def\csname v@ep.deepseek.p2rb.pb\endcsname{$3.0\times10^{-10}$}
\expandafter\def\csname v@ep.deepseek.p2rb.pbraw\endcsname{$1.5\times10^{-10}$}
\expandafter\def\csname v@ep.deepseek.p2rb.praw\endcsname{$9.7\times10^{-20}$}
\expandafter\def\csname v@ep.deepseek.p2rb.pt\endcsname{$1.9\times10^{-19}$}
\expandafter\def\csname v@ep.deepseek.p2rb.upper\endcsname{$+0.277$}
\expandafter\def\csname v@ep.deepseek.p2rb.upperci\endcsname{[$+0.222$, $+0.333$]}
\expandafter\def\csname v@ep.deepseek.p3mrb.ci\endcsname{[$+0.010$, $+0.053$]}
\expandafter\def\csname v@ep.deepseek.p3mrb.lower\endcsname{$+0.031$}
\expandafter\def\csname v@ep.deepseek.p3mrb.lowerci\endcsname{[$+0.010$, $+0.053$]}
\expandafter\def\csname v@ep.deepseek.p3mrb.mean\endcsname{$+0.031$}
\expandafter\def\csname v@ep.deepseek.p3mrb.n\endcsname{256}
\expandafter\def\csname v@ep.deepseek.p3mrb.pb\endcsname{$0.078$}
\expandafter\def\csname v@ep.deepseek.p3mrb.pbraw\endcsname{$0.078$}
\expandafter\def\csname v@ep.deepseek.p3mrb.praw\endcsname{$0.0045$}
\expandafter\def\csname v@ep.deepseek.p3mrb.pt\endcsname{$0.0045$}
\expandafter\def\csname v@ep.deepseek.p3mrb.upper\endcsname{$+0.031$}
\expandafter\def\csname v@ep.deepseek.p3mrb.upperci\endcsname{[$+0.010$, $+0.053$]}
\expandafter\def\csname v@ep.deepseek.planted.k\endcsname{215}
\expandafter\def\csname v@ep.deepseek.planted.n\endcsname{1,024}
\expandafter\def\csname v@ep.deepseek.rate.bad.failure.n\endcsname{0.0\%}
\expandafter\def\csname v@ep.deepseek.rate.bad.failure.neutral\endcsname{0.0\%}
\expandafter\def\csname v@ep.deepseek.rate.bad.failure.nx\endcsname{0.4\%}
\expandafter\def\csname v@ep.deepseek.rate.bad.failure.planted\endcsname{4.3\%}
\expandafter\def\csname v@ep.deepseek.rate.bad.failure.rb\endcsname{0.0\%}
\expandafter\def\csname v@ep.deepseek.rate.bad.failure.rbd\endcsname{0.0\%}
\expandafter\def\csname v@ep.deepseek.rate.bad.failure.rbdx\endcsname{0.0\%}
\expandafter\def\csname v@ep.deepseek.rate.bad.failure.rbx\endcsname{2.0\%}
\expandafter\def\csname v@ep.deepseek.rate.bad.success.n\endcsname{0.8\%}
\expandafter\def\csname v@ep.deepseek.rate.bad.success.neutral\endcsname{0.0\%}
\expandafter\def\csname v@ep.deepseek.rate.bad.success.nx\endcsname{0.0\%}
\expandafter\def\csname v@ep.deepseek.rate.bad.success.planted\endcsname{20.7\%}
\expandafter\def\csname v@ep.deepseek.rate.bad.success.rb\endcsname{3.1\%}
\expandafter\def\csname v@ep.deepseek.rate.bad.success.rbd\endcsname{0.0\%}
\expandafter\def\csname v@ep.deepseek.rate.bad.success.rbdx\endcsname{0.0\%}
\expandafter\def\csname v@ep.deepseek.rate.bad.success.rbx\endcsname{0.0\%}
\expandafter\def\csname v@ep.deepseek.rate.good.failure.n\endcsname{3.1\%}
\expandafter\def\csname v@ep.deepseek.rate.good.failure.neutral\endcsname{0.0\%}
\expandafter\def\csname v@ep.deepseek.rate.good.failure.nx\endcsname{0.0\%}
\expandafter\def\csname v@ep.deepseek.rate.good.failure.planted\endcsname{50.0\%}
\expandafter\def\csname v@ep.deepseek.rate.good.failure.rb\endcsname{27.7\%}
\expandafter\def\csname v@ep.deepseek.rate.good.failure.rbd\endcsname{16.0\%}
\expandafter\def\csname v@ep.deepseek.rate.good.failure.rbdx\endcsname{0.0\%}
\expandafter\def\csname v@ep.deepseek.rate.good.failure.rbx\endcsname{0.0\%}
\expandafter\def\csname v@ep.deepseek.rate.good.success.n\endcsname{0.0\%}
\expandafter\def\csname v@ep.deepseek.rate.good.success.neutral\endcsname{0.0\%}
\expandafter\def\csname v@ep.deepseek.rate.good.success.nx\endcsname{0.4\%}
\expandafter\def\csname v@ep.deepseek.rate.good.success.planted\endcsname{9.0\%}
\expandafter\def\csname v@ep.deepseek.rate.good.success.rb\endcsname{0.0\%}
\expandafter\def\csname v@ep.deepseek.rate.good.success.rbd\endcsname{0.0\%}
\expandafter\def\csname v@ep.deepseek.rate.good.success.rbdx\endcsname{0.0\%}
\expandafter\def\csname v@ep.deepseek.rate.good.success.rbx\endcsname{0.0\%}
\expandafter\def\csname v@ep.deepseek.sus.ci\endcsname{[$+0.184$, $+0.236$]}
\expandafter\def\csname v@ep.deepseek.sus.mean\endcsname{$+0.210$}
\expandafter\def\csname v@ep.deepseek.sus.n\endcsname{256}
\expandafter\def\csname v@ep.deepseek.sus.pbraw\endcsname{$5.5\times10^{-16}$}
\expandafter\def\csname v@ep.deepseek.sus.praw\endcsname{$1.1\times10^{-39}$}
\expandafter\def\csname v@ep.deepseek.word.pthree\endcsname{\emph{bounded below 0.10}}
\expandafter\def\csname v@ep.deepseek.word.ptwo\endcsname{\emph{luck-carrying}}
\expandafter\def\csname v@ep.deepseek.word.sus\endcsname{\emph{increment above 0.10}}
\expandafter\def\csname v@ep.gemini.a2rb.ci\endcsname{[$+0.084$, $+0.166$]}
\expandafter\def\csname v@ep.gemini.a2rb.lower\endcsname{$+0.125$}
\expandafter\def\csname v@ep.gemini.a2rb.lowerci\endcsname{[$+0.084$, $+0.166$]}
\expandafter\def\csname v@ep.gemini.a2rb.mean\endcsname{$+0.125$}
\expandafter\def\csname v@ep.gemini.a2rb.n\endcsname{256}
\expandafter\def\csname v@ep.gemini.a2rb.pbraw\endcsname{$1.7\times10^{-8}$}
\expandafter\def\csname v@ep.gemini.a2rb.praw\endcsname{$5.6\times10^{-9}$}
\expandafter\def\csname v@ep.gemini.a2rb.upper\endcsname{$+0.125$}
\expandafter\def\csname v@ep.gemini.a2rb.upperci\endcsname{[$+0.084$, $+0.166$]}
\expandafter\def\csname v@ep.gemini.a3rb.ci\endcsname{[$-0.014$, $+0.014$]}
\expandafter\def\csname v@ep.gemini.a3rb.lower\endcsname{$0.000$}
\expandafter\def\csname v@ep.gemini.a3rb.lowerci\endcsname{[$-0.014$, $+0.014$]}
\expandafter\def\csname v@ep.gemini.a3rb.mean\endcsname{$0.000$}
\expandafter\def\csname v@ep.gemini.a3rb.n\endcsname{256}
\expandafter\def\csname v@ep.gemini.a3rb.pbraw\endcsname{$1.000$}
\expandafter\def\csname v@ep.gemini.a3rb.praw\endcsname{$1.000$}
\expandafter\def\csname v@ep.gemini.a3rb.upper\endcsname{$0.000$}
\expandafter\def\csname v@ep.gemini.a3rb.upperci\endcsname{[$-0.014$, $+0.014$]}
\expandafter\def\csname v@ep.gemini.count.bad.failure.n\endcsname{0/256}
\expandafter\def\csname v@ep.gemini.count.bad.failure.neutral\endcsname{0/256}
\expandafter\def\csname v@ep.gemini.count.bad.failure.nx\endcsname{0/256}
\expandafter\def\csname v@ep.gemini.count.bad.failure.planted\endcsname{164/256}
\expandafter\def\csname v@ep.gemini.count.bad.failure.rb\endcsname{0/256}
\expandafter\def\csname v@ep.gemini.count.bad.failure.rbd\endcsname{0/256}
\expandafter\def\csname v@ep.gemini.count.bad.failure.rbdx\endcsname{0/256}
\expandafter\def\csname v@ep.gemini.count.bad.failure.rbx\endcsname{0/256}
\expandafter\def\csname v@ep.gemini.count.bad.success.n\endcsname{0/256}
\expandafter\def\csname v@ep.gemini.count.bad.success.neutral\endcsname{0/256}
\expandafter\def\csname v@ep.gemini.count.bad.success.nx\endcsname{0/256}
\expandafter\def\csname v@ep.gemini.count.bad.success.planted\endcsname{75/256}
\expandafter\def\csname v@ep.gemini.count.bad.success.rb\endcsname{0/256}
\expandafter\def\csname v@ep.gemini.count.bad.success.rbd\endcsname{0/256}
\expandafter\def\csname v@ep.gemini.count.bad.success.rbdx\endcsname{0/256}
\expandafter\def\csname v@ep.gemini.count.bad.success.rbx\endcsname{0/256}
\expandafter\def\csname v@ep.gemini.count.good.failure.n\endcsname{1/256}
\expandafter\def\csname v@ep.gemini.count.good.failure.neutral\endcsname{0/256}
\expandafter\def\csname v@ep.gemini.count.good.failure.nx\endcsname{0/256}
\expandafter\def\csname v@ep.gemini.count.good.failure.planted\endcsname{106/256}
\expandafter\def\csname v@ep.gemini.count.good.failure.rb\endcsname{32/256}
\expandafter\def\csname v@ep.gemini.count.good.failure.rbd\endcsname{0/255}
\expandafter\def\csname v@ep.gemini.count.good.failure.rbdx\endcsname{0/256}
\expandafter\def\csname v@ep.gemini.count.good.failure.rbx\endcsname{0/256}
\expandafter\def\csname v@ep.gemini.count.good.success.n\endcsname{0/256}
\expandafter\def\csname v@ep.gemini.count.good.success.neutral\endcsname{0/256}
\expandafter\def\csname v@ep.gemini.count.good.success.nx\endcsname{0/256}
\expandafter\def\csname v@ep.gemini.count.good.success.planted\endcsname{214/256}
\expandafter\def\csname v@ep.gemini.count.good.success.rb\endcsname{0/256}
\expandafter\def\csname v@ep.gemini.count.good.success.rbd\endcsname{0/256}
\expandafter\def\csname v@ep.gemini.count.good.success.rbdx\endcsname{0/255}
\expandafter\def\csname v@ep.gemini.count.good.success.rbx\endcsname{0/256}
\expandafter\def\csname v@ep.gemini.id\endcsname{\texttt{google/gemini-3.8-flash}}
\expandafter\def\csname v@ep.gemini.k.bad.failure.n\endcsname{0}
\expandafter\def\csname v@ep.gemini.k.bad.failure.neutral\endcsname{0}
\expandafter\def\csname v@ep.gemini.k.bad.failure.nx\endcsname{0}
\expandafter\def\csname v@ep.gemini.k.bad.failure.planted\endcsname{164}
\expandafter\def\csname v@ep.gemini.k.bad.failure.rb\endcsname{0}
\expandafter\def\csname v@ep.gemini.k.bad.failure.rbd\endcsname{0}
\expandafter\def\csname v@ep.gemini.k.bad.failure.rbdx\endcsname{0}
\expandafter\def\csname v@ep.gemini.k.bad.failure.rbx\endcsname{0}
\expandafter\def\csname v@ep.gemini.k.bad.success.n\endcsname{0}
\expandafter\def\csname v@ep.gemini.k.bad.success.neutral\endcsname{0}
\expandafter\def\csname v@ep.gemini.k.bad.success.nx\endcsname{0}
\expandafter\def\csname v@ep.gemini.k.bad.success.planted\endcsname{75}
\expandafter\def\csname v@ep.gemini.k.bad.success.rb\endcsname{0}
\expandafter\def\csname v@ep.gemini.k.bad.success.rbd\endcsname{0}
\expandafter\def\csname v@ep.gemini.k.bad.success.rbdx\endcsname{0}
\expandafter\def\csname v@ep.gemini.k.bad.success.rbx\endcsname{0}
\expandafter\def\csname v@ep.gemini.k.good.failure.n\endcsname{1}
\expandafter\def\csname v@ep.gemini.k.good.failure.neutral\endcsname{0}
\expandafter\def\csname v@ep.gemini.k.good.failure.nx\endcsname{0}
\expandafter\def\csname v@ep.gemini.k.good.failure.planted\endcsname{106}
\expandafter\def\csname v@ep.gemini.k.good.failure.rb\endcsname{32}
\expandafter\def\csname v@ep.gemini.k.good.failure.rbd\endcsname{0}
\expandafter\def\csname v@ep.gemini.k.good.failure.rbdx\endcsname{0}
\expandafter\def\csname v@ep.gemini.k.good.failure.rbx\endcsname{0}
\expandafter\def\csname v@ep.gemini.k.good.success.n\endcsname{0}
\expandafter\def\csname v@ep.gemini.k.good.success.neutral\endcsname{0}
\expandafter\def\csname v@ep.gemini.k.good.success.nx\endcsname{0}
\expandafter\def\csname v@ep.gemini.k.good.success.planted\endcsname{214}
\expandafter\def\csname v@ep.gemini.k.good.success.rb\endcsname{0}
\expandafter\def\csname v@ep.gemini.k.good.success.rbd\endcsname{0}
\expandafter\def\csname v@ep.gemini.k.good.success.rbdx\endcsname{0}
\expandafter\def\csname v@ep.gemini.k.good.success.rbx\endcsname{0}
\expandafter\def\csname v@ep.gemini.l2brb.ci\endcsname{[$-0.014$, $+0.014$]}
\expandafter\def\csname v@ep.gemini.l2brb.mean\endcsname{$0.000$}
\expandafter\def\csname v@ep.gemini.l2brb.n\endcsname{256}
\expandafter\def\csname v@ep.gemini.l2brb.pbraw\endcsname{$1.000$}
\expandafter\def\csname v@ep.gemini.l2brb.praw\endcsname{$1.000$}
\expandafter\def\csname v@ep.gemini.l2rb.ci\endcsname{[$+0.084$, $+0.166$]}
\expandafter\def\csname v@ep.gemini.l2rb.lower\endcsname{$+0.125$}
\expandafter\def\csname v@ep.gemini.l2rb.lowerci\endcsname{[$+0.084$, $+0.166$]}
\expandafter\def\csname v@ep.gemini.l2rb.mean\endcsname{$+0.125$}
\expandafter\def\csname v@ep.gemini.l2rb.n\endcsname{256}
\expandafter\def\csname v@ep.gemini.l2rb.pbraw\endcsname{$1.7\times10^{-8}$}
\expandafter\def\csname v@ep.gemini.l2rb.praw\endcsname{$5.6\times10^{-9}$}
\expandafter\def\csname v@ep.gemini.l2rb.upper\endcsname{$+0.125$}
\expandafter\def\csname v@ep.gemini.l2rb.upperci\endcsname{[$+0.084$, $+0.166$]}
\expandafter\def\csname v@ep.gemini.l3rb.ci\endcsname{[$-0.014$, $+0.014$]}
\expandafter\def\csname v@ep.gemini.l3rb.lower\endcsname{$0.000$}
\expandafter\def\csname v@ep.gemini.l3rb.lowerci\endcsname{[$-0.014$, $+0.014$]}
\expandafter\def\csname v@ep.gemini.l3rb.mean\endcsname{$0.000$}
\expandafter\def\csname v@ep.gemini.l3rb.n\endcsname{256}
\expandafter\def\csname v@ep.gemini.l3rb.pbraw\endcsname{$1.000$}
\expandafter\def\csname v@ep.gemini.l3rb.praw\endcsname{$1.000$}
\expandafter\def\csname v@ep.gemini.l3rb.upper\endcsname{$0.000$}
\expandafter\def\csname v@ep.gemini.l3rb.upperci\endcsname{[$-0.014$, $+0.014$]}
\expandafter\def\csname v@ep.gemini.n.bad.failure.n\endcsname{256}
\expandafter\def\csname v@ep.gemini.n.bad.failure.neutral\endcsname{256}
\expandafter\def\csname v@ep.gemini.n.bad.failure.nx\endcsname{256}
\expandafter\def\csname v@ep.gemini.n.bad.failure.planted\endcsname{256}
\expandafter\def\csname v@ep.gemini.n.bad.failure.rb\endcsname{256}
\expandafter\def\csname v@ep.gemini.n.bad.failure.rbd\endcsname{256}
\expandafter\def\csname v@ep.gemini.n.bad.failure.rbdx\endcsname{256}
\expandafter\def\csname v@ep.gemini.n.bad.failure.rbx\endcsname{256}
\expandafter\def\csname v@ep.gemini.n.bad.success.n\endcsname{256}
\expandafter\def\csname v@ep.gemini.n.bad.success.neutral\endcsname{256}
\expandafter\def\csname v@ep.gemini.n.bad.success.nx\endcsname{256}
\expandafter\def\csname v@ep.gemini.n.bad.success.planted\endcsname{256}
\expandafter\def\csname v@ep.gemini.n.bad.success.rb\endcsname{256}
\expandafter\def\csname v@ep.gemini.n.bad.success.rbd\endcsname{256}
\expandafter\def\csname v@ep.gemini.n.bad.success.rbdx\endcsname{256}
\expandafter\def\csname v@ep.gemini.n.bad.success.rbx\endcsname{256}
\expandafter\def\csname v@ep.gemini.n.good.failure.n\endcsname{256}
\expandafter\def\csname v@ep.gemini.n.good.failure.neutral\endcsname{256}
\expandafter\def\csname v@ep.gemini.n.good.failure.nx\endcsname{256}
\expandafter\def\csname v@ep.gemini.n.good.failure.planted\endcsname{256}
\expandafter\def\csname v@ep.gemini.n.good.failure.rb\endcsname{256}
\expandafter\def\csname v@ep.gemini.n.good.failure.rbd\endcsname{255}
\expandafter\def\csname v@ep.gemini.n.good.failure.rbdx\endcsname{256}
\expandafter\def\csname v@ep.gemini.n.good.failure.rbx\endcsname{256}
\expandafter\def\csname v@ep.gemini.n.good.success.n\endcsname{256}
\expandafter\def\csname v@ep.gemini.n.good.success.neutral\endcsname{256}
\expandafter\def\csname v@ep.gemini.n.good.success.nx\endcsname{256}
\expandafter\def\csname v@ep.gemini.n.good.success.planted\endcsname{256}
\expandafter\def\csname v@ep.gemini.n.good.success.rb\endcsname{256}
\expandafter\def\csname v@ep.gemini.n.good.success.rbd\endcsname{256}
\expandafter\def\csname v@ep.gemini.n.good.success.rbdx\endcsname{255}
\expandafter\def\csname v@ep.gemini.n.good.success.rbx\endcsname{256}
\expandafter\def\csname v@ep.gemini.name\endcsname{Gemini 3.8 Flash}
\expandafter\def\csname v@ep.gemini.namesworse\endcsname{26\%}
\expandafter\def\csname v@ep.gemini.p2rb.ci\endcsname{[$+0.084$, $+0.166$]}
\expandafter\def\csname v@ep.gemini.p2rb.lower\endcsname{$+0.125$}
\expandafter\def\csname v@ep.gemini.p2rb.lowerci\endcsname{[$+0.084$, $+0.166$]}
\expandafter\def\csname v@ep.gemini.p2rb.mean\endcsname{$+0.125$}
\expandafter\def\csname v@ep.gemini.p2rb.n\endcsname{256}
\expandafter\def\csname v@ep.gemini.p2rb.pb\endcsname{$7.5\times10^{-5}$}
\expandafter\def\csname v@ep.gemini.p2rb.pbraw\endcsname{$3.8\times10^{-5}$}
\expandafter\def\csname v@ep.gemini.p2rb.praw\endcsname{$5.6\times10^{-9}$}
\expandafter\def\csname v@ep.gemini.p2rb.pt\endcsname{$1.1\times10^{-8}$}
\expandafter\def\csname v@ep.gemini.p2rb.upper\endcsname{$+0.125$}
\expandafter\def\csname v@ep.gemini.p2rb.upperci\endcsname{[$+0.084$, $+0.166$]}
\expandafter\def\csname v@ep.gemini.p3mrb.ci\endcsname{[$-0.029$, $+0.029$]}
\expandafter\def\csname v@ep.gemini.p3mrb.lower\endcsname{$0.000$}
\expandafter\def\csname v@ep.gemini.p3mrb.lowerci\endcsname{[$-0.029$, $+0.029$]}
\expandafter\def\csname v@ep.gemini.p3mrb.mean\endcsname{$0.000$}
\expandafter\def\csname v@ep.gemini.p3mrb.n\endcsname{256}
\expandafter\def\csname v@ep.gemini.p3mrb.pb\endcsname{$1.000$}
\expandafter\def\csname v@ep.gemini.p3mrb.pbraw\endcsname{$1.000$}
\expandafter\def\csname v@ep.gemini.p3mrb.praw\endcsname{$1.000$}
\expandafter\def\csname v@ep.gemini.p3mrb.pt\endcsname{$1.000$}
\expandafter\def\csname v@ep.gemini.p3mrb.upper\endcsname{$0.000$}
\expandafter\def\csname v@ep.gemini.p3mrb.upperci\endcsname{[$-0.029$, $+0.029$]}
\expandafter\def\csname v@ep.gemini.planted.k\endcsname{559}
\expandafter\def\csname v@ep.gemini.planted.n\endcsname{1,024}
\expandafter\def\csname v@ep.gemini.rate.bad.failure.n\endcsname{0.0\%}
\expandafter\def\csname v@ep.gemini.rate.bad.failure.neutral\endcsname{0.0\%}
\expandafter\def\csname v@ep.gemini.rate.bad.failure.nx\endcsname{0.0\%}
\expandafter\def\csname v@ep.gemini.rate.bad.failure.planted\endcsname{64.1\%}
\expandafter\def\csname v@ep.gemini.rate.bad.failure.rb\endcsname{0.0\%}
\expandafter\def\csname v@ep.gemini.rate.bad.failure.rbd\endcsname{0.0\%}
\expandafter\def\csname v@ep.gemini.rate.bad.failure.rbdx\endcsname{0.0\%}
\expandafter\def\csname v@ep.gemini.rate.bad.failure.rbx\endcsname{0.0\%}
\expandafter\def\csname v@ep.gemini.rate.bad.success.n\endcsname{0.0\%}
\expandafter\def\csname v@ep.gemini.rate.bad.success.neutral\endcsname{0.0\%}
\expandafter\def\csname v@ep.gemini.rate.bad.success.nx\endcsname{0.0\%}
\expandafter\def\csname v@ep.gemini.rate.bad.success.planted\endcsname{29.3\%}
\expandafter\def\csname v@ep.gemini.rate.bad.success.rb\endcsname{0.0\%}
\expandafter\def\csname v@ep.gemini.rate.bad.success.rbd\endcsname{0.0\%}
\expandafter\def\csname v@ep.gemini.rate.bad.success.rbdx\endcsname{0.0\%}
\expandafter\def\csname v@ep.gemini.rate.bad.success.rbx\endcsname{0.0\%}
\expandafter\def\csname v@ep.gemini.rate.good.failure.n\endcsname{0.4\%}
\expandafter\def\csname v@ep.gemini.rate.good.failure.neutral\endcsname{0.0\%}
\expandafter\def\csname v@ep.gemini.rate.good.failure.nx\endcsname{0.0\%}
\expandafter\def\csname v@ep.gemini.rate.good.failure.planted\endcsname{41.4\%}
\expandafter\def\csname v@ep.gemini.rate.good.failure.rb\endcsname{12.5\%}
\expandafter\def\csname v@ep.gemini.rate.good.failure.rbd\endcsname{0.0\%}
\expandafter\def\csname v@ep.gemini.rate.good.failure.rbdx\endcsname{0.0\%}
\expandafter\def\csname v@ep.gemini.rate.good.failure.rbx\endcsname{0.0\%}
\expandafter\def\csname v@ep.gemini.rate.good.success.n\endcsname{0.0\%}
\expandafter\def\csname v@ep.gemini.rate.good.success.neutral\endcsname{0.0\%}
\expandafter\def\csname v@ep.gemini.rate.good.success.nx\endcsname{0.0\%}
\expandafter\def\csname v@ep.gemini.rate.good.success.planted\endcsname{83.6\%}
\expandafter\def\csname v@ep.gemini.rate.good.success.rb\endcsname{0.0\%}
\expandafter\def\csname v@ep.gemini.rate.good.success.rbd\endcsname{0.0\%}
\expandafter\def\csname v@ep.gemini.rate.good.success.rbdx\endcsname{0.0\%}
\expandafter\def\csname v@ep.gemini.rate.good.success.rbx\endcsname{0.0\%}
\expandafter\def\csname v@ep.gemini.sus.ci\endcsname{[$+0.514$, $+0.578$]}
\expandafter\def\csname v@ep.gemini.sus.mean\endcsname{$+0.546$}
\expandafter\def\csname v@ep.gemini.sus.n\endcsname{256}
\expandafter\def\csname v@ep.gemini.sus.pbraw\endcsname{$9.0\times10^{-38}$}
\expandafter\def\csname v@ep.gemini.sus.praw\endcsname{$3.3\times10^{-95}$}
\expandafter\def\csname v@ep.gemini.word.pthree\endcsname{\emph{bounded below 0.10}}
\expandafter\def\csname v@ep.gemini.word.ptwo\endcsname{\emph{luck-carrying}}
\expandafter\def\csname v@ep.gemini.word.sus\endcsname{\emph{increment above 0.10}}
\expandafter\def\csname v@ep.grok.a2rb.ci\endcsname{[$+0.068$, $+0.143$]}
\expandafter\def\csname v@ep.grok.a2rb.lower\endcsname{$+0.105$}
\expandafter\def\csname v@ep.grok.a2rb.lowerci\endcsname{[$+0.068$, $+0.143$]}
\expandafter\def\csname v@ep.grok.a2rb.mean\endcsname{$+0.105$}
\expandafter\def\csname v@ep.grok.a2rb.n\endcsname{256}
\expandafter\def\csname v@ep.grok.a2rb.pbraw\endcsname{$2.7\times10^{-7}$}
\expandafter\def\csname v@ep.grok.a2rb.praw\endcsname{$1.0\times10^{-7}$}
\expandafter\def\csname v@ep.grok.a2rb.upper\endcsname{$+0.105$}
\expandafter\def\csname v@ep.grok.a2rb.upperci\endcsname{[$+0.068$, $+0.143$]}
\expandafter\def\csname v@ep.grok.a3rb.ci\endcsname{[$-0.014$, $+0.014$]}
\expandafter\def\csname v@ep.grok.a3rb.lower\endcsname{$0.000$}
\expandafter\def\csname v@ep.grok.a3rb.lowerci\endcsname{[$-0.014$, $+0.014$]}
\expandafter\def\csname v@ep.grok.a3rb.mean\endcsname{$0.000$}
\expandafter\def\csname v@ep.grok.a3rb.n\endcsname{256}
\expandafter\def\csname v@ep.grok.a3rb.pbraw\endcsname{$1.000$}
\expandafter\def\csname v@ep.grok.a3rb.praw\endcsname{$1.000$}
\expandafter\def\csname v@ep.grok.a3rb.upper\endcsname{$0.000$}
\expandafter\def\csname v@ep.grok.a3rb.upperci\endcsname{[$-0.014$, $+0.014$]}
\expandafter\def\csname v@ep.grok.count.bad.failure.n\endcsname{0/256}
\expandafter\def\csname v@ep.grok.count.bad.failure.neutral\endcsname{0/256}
\expandafter\def\csname v@ep.grok.count.bad.failure.nx\endcsname{0/256}
\expandafter\def\csname v@ep.grok.count.bad.failure.planted\endcsname{1/256}
\expandafter\def\csname v@ep.grok.count.bad.failure.rb\endcsname{0/256}
\expandafter\def\csname v@ep.grok.count.bad.failure.rbd\endcsname{0/256}
\expandafter\def\csname v@ep.grok.count.bad.failure.rbdx\endcsname{0/256}
\expandafter\def\csname v@ep.grok.count.bad.failure.rbx\endcsname{0/256}
\expandafter\def\csname v@ep.grok.count.bad.success.n\endcsname{0/256}
\expandafter\def\csname v@ep.grok.count.bad.success.neutral\endcsname{0/256}
\expandafter\def\csname v@ep.grok.count.bad.success.nx\endcsname{0/256}
\expandafter\def\csname v@ep.grok.count.bad.success.planted\endcsname{1/256}
\expandafter\def\csname v@ep.grok.count.bad.success.rb\endcsname{0/256}
\expandafter\def\csname v@ep.grok.count.bad.success.rbd\endcsname{0/256}
\expandafter\def\csname v@ep.grok.count.bad.success.rbdx\endcsname{0/256}
\expandafter\def\csname v@ep.grok.count.bad.success.rbx\endcsname{0/256}
\expandafter\def\csname v@ep.grok.count.good.failure.n\endcsname{16/256}
\expandafter\def\csname v@ep.grok.count.good.failure.neutral\endcsname{0/256}
\expandafter\def\csname v@ep.grok.count.good.failure.nx\endcsname{1/256}
\expandafter\def\csname v@ep.grok.count.good.failure.planted\endcsname{2/256}
\expandafter\def\csname v@ep.grok.count.good.failure.rb\endcsname{27/256}
\expandafter\def\csname v@ep.grok.count.good.failure.rbd\endcsname{5/256}
\expandafter\def\csname v@ep.grok.count.good.failure.rbdx\endcsname{0/256}
\expandafter\def\csname v@ep.grok.count.good.failure.rbx\endcsname{0/256}
\expandafter\def\csname v@ep.grok.count.good.success.n\endcsname{0/256}
\expandafter\def\csname v@ep.grok.count.good.success.neutral\endcsname{0/256}
\expandafter\def\csname v@ep.grok.count.good.success.nx\endcsname{1/256}
\expandafter\def\csname v@ep.grok.count.good.success.planted\endcsname{0/256}
\expandafter\def\csname v@ep.grok.count.good.success.rb\endcsname{0/256}
\expandafter\def\csname v@ep.grok.count.good.success.rbd\endcsname{0/256}
\expandafter\def\csname v@ep.grok.count.good.success.rbdx\endcsname{0/256}
\expandafter\def\csname v@ep.grok.count.good.success.rbx\endcsname{1/256}
\expandafter\def\csname v@ep.grok.id\endcsname{\texttt{x-ai/grok-4.7}}
\expandafter\def\csname v@ep.grok.k.bad.failure.n\endcsname{0}
\expandafter\def\csname v@ep.grok.k.bad.failure.neutral\endcsname{0}
\expandafter\def\csname v@ep.grok.k.bad.failure.nx\endcsname{0}
\expandafter\def\csname v@ep.grok.k.bad.failure.planted\endcsname{1}
\expandafter\def\csname v@ep.grok.k.bad.failure.rb\endcsname{0}
\expandafter\def\csname v@ep.grok.k.bad.failure.rbd\endcsname{0}
\expandafter\def\csname v@ep.grok.k.bad.failure.rbdx\endcsname{0}
\expandafter\def\csname v@ep.grok.k.bad.failure.rbx\endcsname{0}
\expandafter\def\csname v@ep.grok.k.bad.success.n\endcsname{0}
\expandafter\def\csname v@ep.grok.k.bad.success.neutral\endcsname{0}
\expandafter\def\csname v@ep.grok.k.bad.success.nx\endcsname{0}
\expandafter\def\csname v@ep.grok.k.bad.success.planted\endcsname{1}
\expandafter\def\csname v@ep.grok.k.bad.success.rb\endcsname{0}
\expandafter\def\csname v@ep.grok.k.bad.success.rbd\endcsname{0}
\expandafter\def\csname v@ep.grok.k.bad.success.rbdx\endcsname{0}
\expandafter\def\csname v@ep.grok.k.bad.success.rbx\endcsname{0}
\expandafter\def\csname v@ep.grok.k.good.failure.n\endcsname{16}
\expandafter\def\csname v@ep.grok.k.good.failure.neutral\endcsname{0}
\expandafter\def\csname v@ep.grok.k.good.failure.nx\endcsname{1}
\expandafter\def\csname v@ep.grok.k.good.failure.planted\endcsname{2}
\expandafter\def\csname v@ep.grok.k.good.failure.rb\endcsname{27}
\expandafter\def\csname v@ep.grok.k.good.failure.rbd\endcsname{5}
\expandafter\def\csname v@ep.grok.k.good.failure.rbdx\endcsname{0}
\expandafter\def\csname v@ep.grok.k.good.failure.rbx\endcsname{0}
\expandafter\def\csname v@ep.grok.k.good.success.n\endcsname{0}
\expandafter\def\csname v@ep.grok.k.good.success.neutral\endcsname{0}
\expandafter\def\csname v@ep.grok.k.good.success.nx\endcsname{1}
\expandafter\def\csname v@ep.grok.k.good.success.planted\endcsname{0}
\expandafter\def\csname v@ep.grok.k.good.success.rb\endcsname{0}
\expandafter\def\csname v@ep.grok.k.good.success.rbd\endcsname{0}
\expandafter\def\csname v@ep.grok.k.good.success.rbdx\endcsname{0}
\expandafter\def\csname v@ep.grok.k.good.success.rbx\endcsname{1}
\expandafter\def\csname v@ep.grok.l2brb.ci\endcsname{[$-0.004$, $+0.012$]}
\expandafter\def\csname v@ep.grok.l2brb.mean\endcsname{$+0.004$}
\expandafter\def\csname v@ep.grok.l2brb.n\endcsname{256}
\expandafter\def\csname v@ep.grok.l2brb.pbraw\endcsname{$0.571$}
\expandafter\def\csname v@ep.grok.l2brb.praw\endcsname{$0.318$}
\expandafter\def\csname v@ep.grok.l2rb.ci\endcsname{[$+0.068$, $+0.143$]}
\expandafter\def\csname v@ep.grok.l2rb.lower\endcsname{$+0.105$}
\expandafter\def\csname v@ep.grok.l2rb.lowerci\endcsname{[$+0.068$, $+0.143$]}
\expandafter\def\csname v@ep.grok.l2rb.mean\endcsname{$+0.105$}
\expandafter\def\csname v@ep.grok.l2rb.n\endcsname{256}
\expandafter\def\csname v@ep.grok.l2rb.pbraw\endcsname{$2.7\times10^{-7}$}
\expandafter\def\csname v@ep.grok.l2rb.praw\endcsname{$1.0\times10^{-7}$}
\expandafter\def\csname v@ep.grok.l2rb.upper\endcsname{$+0.105$}
\expandafter\def\csname v@ep.grok.l2rb.upperci\endcsname{[$+0.068$, $+0.143$]}
\expandafter\def\csname v@ep.grok.l3rb.ci\endcsname{[$-0.014$, $+0.014$]}
\expandafter\def\csname v@ep.grok.l3rb.lower\endcsname{$0.000$}
\expandafter\def\csname v@ep.grok.l3rb.lowerci\endcsname{[$-0.014$, $+0.014$]}
\expandafter\def\csname v@ep.grok.l3rb.mean\endcsname{$0.000$}
\expandafter\def\csname v@ep.grok.l3rb.n\endcsname{256}
\expandafter\def\csname v@ep.grok.l3rb.pbraw\endcsname{$1.000$}
\expandafter\def\csname v@ep.grok.l3rb.praw\endcsname{$1.000$}
\expandafter\def\csname v@ep.grok.l3rb.upper\endcsname{$0.000$}
\expandafter\def\csname v@ep.grok.l3rb.upperci\endcsname{[$-0.014$, $+0.014$]}
\expandafter\def\csname v@ep.grok.n.bad.failure.n\endcsname{256}
\expandafter\def\csname v@ep.grok.n.bad.failure.neutral\endcsname{256}
\expandafter\def\csname v@ep.grok.n.bad.failure.nx\endcsname{256}
\expandafter\def\csname v@ep.grok.n.bad.failure.planted\endcsname{256}
\expandafter\def\csname v@ep.grok.n.bad.failure.rb\endcsname{256}
\expandafter\def\csname v@ep.grok.n.bad.failure.rbd\endcsname{256}
\expandafter\def\csname v@ep.grok.n.bad.failure.rbdx\endcsname{256}
\expandafter\def\csname v@ep.grok.n.bad.failure.rbx\endcsname{256}
\expandafter\def\csname v@ep.grok.n.bad.success.n\endcsname{256}
\expandafter\def\csname v@ep.grok.n.bad.success.neutral\endcsname{256}
\expandafter\def\csname v@ep.grok.n.bad.success.nx\endcsname{256}
\expandafter\def\csname v@ep.grok.n.bad.success.planted\endcsname{256}
\expandafter\def\csname v@ep.grok.n.bad.success.rb\endcsname{256}
\expandafter\def\csname v@ep.grok.n.bad.success.rbd\endcsname{256}
\expandafter\def\csname v@ep.grok.n.bad.success.rbdx\endcsname{256}
\expandafter\def\csname v@ep.grok.n.bad.success.rbx\endcsname{256}
\expandafter\def\csname v@ep.grok.n.good.failure.n\endcsname{256}
\expandafter\def\csname v@ep.grok.n.good.failure.neutral\endcsname{256}
\expandafter\def\csname v@ep.grok.n.good.failure.nx\endcsname{256}
\expandafter\def\csname v@ep.grok.n.good.failure.planted\endcsname{256}
\expandafter\def\csname v@ep.grok.n.good.failure.rb\endcsname{256}
\expandafter\def\csname v@ep.grok.n.good.failure.rbd\endcsname{256}
\expandafter\def\csname v@ep.grok.n.good.failure.rbdx\endcsname{256}
\expandafter\def\csname v@ep.grok.n.good.failure.rbx\endcsname{256}
\expandafter\def\csname v@ep.grok.n.good.success.n\endcsname{256}
\expandafter\def\csname v@ep.grok.n.good.success.neutral\endcsname{256}
\expandafter\def\csname v@ep.grok.n.good.success.nx\endcsname{256}
\expandafter\def\csname v@ep.grok.n.good.success.planted\endcsname{256}
\expandafter\def\csname v@ep.grok.n.good.success.rb\endcsname{256}
\expandafter\def\csname v@ep.grok.n.good.success.rbd\endcsname{256}
\expandafter\def\csname v@ep.grok.n.good.success.rbdx\endcsname{256}
\expandafter\def\csname v@ep.grok.n.good.success.rbx\endcsname{256}
\expandafter\def\csname v@ep.grok.name\endcsname{Grok 4.7}
\expandafter\def\csname v@ep.grok.namesworse\endcsname{0\%}
\expandafter\def\csname v@ep.grok.p2rb.ci\endcsname{[$+0.068$, $+0.143$]}
\expandafter\def\csname v@ep.grok.p2rb.lower\endcsname{$+0.105$}
\expandafter\def\csname v@ep.grok.p2rb.lowerci\endcsname{[$+0.068$, $+0.143$]}
\expandafter\def\csname v@ep.grok.p2rb.mean\endcsname{$+0.105$}
\expandafter\def\csname v@ep.grok.p2rb.n\endcsname{256}
\expandafter\def\csname v@ep.grok.p2rb.pb\endcsname{$3.7\times10^{-4}$}
\expandafter\def\csname v@ep.grok.p2rb.pbraw\endcsname{$1.8\times10^{-4}$}
\expandafter\def\csname v@ep.grok.p2rb.praw\endcsname{$1.0\times10^{-7}$}
\expandafter\def\csname v@ep.grok.p2rb.pt\endcsname{$2.0\times10^{-7}$}
\expandafter\def\csname v@ep.grok.p2rb.upper\endcsname{$+0.105$}
\expandafter\def\csname v@ep.grok.p2rb.upperci\endcsname{[$+0.068$, $+0.143$]}
\expandafter\def\csname v@ep.grok.p3mrb.ci\endcsname{[$-0.029$, $+0.029$]}
\expandafter\def\csname v@ep.grok.p3mrb.lower\endcsname{$0.000$}
\expandafter\def\csname v@ep.grok.p3mrb.lowerci\endcsname{[$-0.029$, $+0.029$]}
\expandafter\def\csname v@ep.grok.p3mrb.mean\endcsname{$0.000$}
\expandafter\def\csname v@ep.grok.p3mrb.n\endcsname{256}
\expandafter\def\csname v@ep.grok.p3mrb.pb\endcsname{$1.000$}
\expandafter\def\csname v@ep.grok.p3mrb.pbraw\endcsname{$1.000$}
\expandafter\def\csname v@ep.grok.p3mrb.praw\endcsname{$1.000$}
\expandafter\def\csname v@ep.grok.p3mrb.pt\endcsname{$1.000$}
\expandafter\def\csname v@ep.grok.p3mrb.upper\endcsname{$0.000$}
\expandafter\def\csname v@ep.grok.p3mrb.upperci\endcsname{[$-0.029$, $+0.029$]}
\expandafter\def\csname v@ep.grok.planted.k\endcsname{4}
\expandafter\def\csname v@ep.grok.planted.n\endcsname{1,024}
\expandafter\def\csname v@ep.grok.rate.bad.failure.n\endcsname{0.0\%}
\expandafter\def\csname v@ep.grok.rate.bad.failure.neutral\endcsname{0.0\%}
\expandafter\def\csname v@ep.grok.rate.bad.failure.nx\endcsname{0.0\%}
\expandafter\def\csname v@ep.grok.rate.bad.failure.planted\endcsname{0.4\%}
\expandafter\def\csname v@ep.grok.rate.bad.failure.rb\endcsname{0.0\%}
\expandafter\def\csname v@ep.grok.rate.bad.failure.rbd\endcsname{0.0\%}
\expandafter\def\csname v@ep.grok.rate.bad.failure.rbdx\endcsname{0.0\%}
\expandafter\def\csname v@ep.grok.rate.bad.failure.rbx\endcsname{0.0\%}
\expandafter\def\csname v@ep.grok.rate.bad.success.n\endcsname{0.0\%}
\expandafter\def\csname v@ep.grok.rate.bad.success.neutral\endcsname{0.0\%}
\expandafter\def\csname v@ep.grok.rate.bad.success.nx\endcsname{0.0\%}
\expandafter\def\csname v@ep.grok.rate.bad.success.planted\endcsname{0.4\%}
\expandafter\def\csname v@ep.grok.rate.bad.success.rb\endcsname{0.0\%}
\expandafter\def\csname v@ep.grok.rate.bad.success.rbd\endcsname{0.0\%}
\expandafter\def\csname v@ep.grok.rate.bad.success.rbdx\endcsname{0.0\%}
\expandafter\def\csname v@ep.grok.rate.bad.success.rbx\endcsname{0.0\%}
\expandafter\def\csname v@ep.grok.rate.good.failure.n\endcsname{6.3\%}
\expandafter\def\csname v@ep.grok.rate.good.failure.neutral\endcsname{0.0\%}
\expandafter\def\csname v@ep.grok.rate.good.failure.nx\endcsname{0.4\%}
\expandafter\def\csname v@ep.grok.rate.good.failure.planted\endcsname{0.8\%}
\expandafter\def\csname v@ep.grok.rate.good.failure.rb\endcsname{10.5\%}
\expandafter\def\csname v@ep.grok.rate.good.failure.rbd\endcsname{2.0\%}
\expandafter\def\csname v@ep.grok.rate.good.failure.rbdx\endcsname{0.0\%}
\expandafter\def\csname v@ep.grok.rate.good.failure.rbx\endcsname{0.0\%}
\expandafter\def\csname v@ep.grok.rate.good.success.n\endcsname{0.0\%}
\expandafter\def\csname v@ep.grok.rate.good.success.neutral\endcsname{0.0\%}
\expandafter\def\csname v@ep.grok.rate.good.success.nx\endcsname{0.4\%}
\expandafter\def\csname v@ep.grok.rate.good.success.planted\endcsname{0.0\%}
\expandafter\def\csname v@ep.grok.rate.good.success.rb\endcsname{0.0\%}
\expandafter\def\csname v@ep.grok.rate.good.success.rbd\endcsname{0.0\%}
\expandafter\def\csname v@ep.grok.rate.good.success.rbdx\endcsname{0.0\%}
\expandafter\def\csname v@ep.grok.rate.good.success.rbx\endcsname{0.4\%}
\expandafter\def\csname v@ep.grok.sus.ci\endcsname{[$0.000$, $+0.008$]}
\expandafter\def\csname v@ep.grok.sus.mean\endcsname{$+0.004$}
\expandafter\def\csname v@ep.grok.sus.n\endcsname{256}
\expandafter\def\csname v@ep.grok.sus.pbraw\endcsname{$0.503$}
\expandafter\def\csname v@ep.grok.sus.praw\endcsname{$0.045$}
\expandafter\def\csname v@ep.grok.word.pthree\endcsname{\emph{bounded below 0.10}}
\expandafter\def\csname v@ep.grok.word.ptwo\endcsname{\emph{luck-carrying}}
\expandafter\def\csname v@ep.grok.word.sus\endcsname{\emph{increment below 0.10}}
\expandafter\def\csname v@ep.haiku.a2rb.ci\endcsname{[$+0.024$, $+0.078$]}
\expandafter\def\csname v@ep.haiku.a2rb.lower\endcsname{$+0.051$}
\expandafter\def\csname v@ep.haiku.a2rb.lowerci\endcsname{[$+0.024$, $+0.078$]}
\expandafter\def\csname v@ep.haiku.a2rb.mean\endcsname{$+0.051$}
\expandafter\def\csname v@ep.haiku.a2rb.n\endcsname{256}
\expandafter\def\csname v@ep.haiku.a2rb.pbraw\endcsname{$6.9\times10^{-4}$}
\expandafter\def\csname v@ep.haiku.a2rb.praw\endcsname{$2.7\times10^{-4}$}
\expandafter\def\csname v@ep.haiku.a2rb.upper\endcsname{$+0.051$}
\expandafter\def\csname v@ep.haiku.a2rb.upperci\endcsname{[$+0.024$, $+0.078$]}
\expandafter\def\csname v@ep.haiku.a3rb.ci\endcsname{[$-0.014$, $+0.014$]}
\expandafter\def\csname v@ep.haiku.a3rb.lower\endcsname{$0.000$}
\expandafter\def\csname v@ep.haiku.a3rb.lowerci\endcsname{[$-0.014$, $+0.014$]}
\expandafter\def\csname v@ep.haiku.a3rb.mean\endcsname{$0.000$}
\expandafter\def\csname v@ep.haiku.a3rb.n\endcsname{256}
\expandafter\def\csname v@ep.haiku.a3rb.pbraw\endcsname{$1.000$}
\expandafter\def\csname v@ep.haiku.a3rb.praw\endcsname{$1.000$}
\expandafter\def\csname v@ep.haiku.a3rb.upper\endcsname{$0.000$}
\expandafter\def\csname v@ep.haiku.a3rb.upperci\endcsname{[$-0.014$, $+0.014$]}
\expandafter\def\csname v@ep.haiku.count.bad.failure.n\endcsname{0/256}
\expandafter\def\csname v@ep.haiku.count.bad.failure.neutral\endcsname{0/256}
\expandafter\def\csname v@ep.haiku.count.bad.failure.nx\endcsname{0/256}
\expandafter\def\csname v@ep.haiku.count.bad.failure.planted\endcsname{0/256}
\expandafter\def\csname v@ep.haiku.count.bad.failure.rb\endcsname{0/256}
\expandafter\def\csname v@ep.haiku.count.bad.failure.rbd\endcsname{0/256}
\expandafter\def\csname v@ep.haiku.count.bad.failure.rbdx\endcsname{0/256}
\expandafter\def\csname v@ep.haiku.count.bad.failure.rbx\endcsname{0/256}
\expandafter\def\csname v@ep.haiku.count.bad.success.n\endcsname{0/256}
\expandafter\def\csname v@ep.haiku.count.bad.success.neutral\endcsname{0/256}
\expandafter\def\csname v@ep.haiku.count.bad.success.nx\endcsname{0/256}
\expandafter\def\csname v@ep.haiku.count.bad.success.planted\endcsname{0/256}
\expandafter\def\csname v@ep.haiku.count.bad.success.rb\endcsname{0/256}
\expandafter\def\csname v@ep.haiku.count.bad.success.rbd\endcsname{0/256}
\expandafter\def\csname v@ep.haiku.count.bad.success.rbdx\endcsname{0/256}
\expandafter\def\csname v@ep.haiku.count.bad.success.rbx\endcsname{0/256}
\expandafter\def\csname v@ep.haiku.count.good.failure.n\endcsname{4/256}
\expandafter\def\csname v@ep.haiku.count.good.failure.neutral\endcsname{0/256}
\expandafter\def\csname v@ep.haiku.count.good.failure.nx\endcsname{3/256}
\expandafter\def\csname v@ep.haiku.count.good.failure.planted\endcsname{0/256}
\expandafter\def\csname v@ep.haiku.count.good.failure.rb\endcsname{13/256}
\expandafter\def\csname v@ep.haiku.count.good.failure.rbd\endcsname{5/256}
\expandafter\def\csname v@ep.haiku.count.good.failure.rbdx\endcsname{4/256}
\expandafter\def\csname v@ep.haiku.count.good.failure.rbx\endcsname{2/256}
\expandafter\def\csname v@ep.haiku.count.good.success.n\endcsname{0/256}
\expandafter\def\csname v@ep.haiku.count.good.success.neutral\endcsname{0/256}
\expandafter\def\csname v@ep.haiku.count.good.success.nx\endcsname{0/256}
\expandafter\def\csname v@ep.haiku.count.good.success.planted\endcsname{0/256}
\expandafter\def\csname v@ep.haiku.count.good.success.rb\endcsname{1/256}
\expandafter\def\csname v@ep.haiku.count.good.success.rbd\endcsname{0/256}
\expandafter\def\csname v@ep.haiku.count.good.success.rbdx\endcsname{1/256}
\expandafter\def\csname v@ep.haiku.count.good.success.rbx\endcsname{0/256}
\expandafter\def\csname v@ep.haiku.id\endcsname{\texttt{anthropic/claude-haiku-5.5}}
\expandafter\def\csname v@ep.haiku.k.bad.failure.n\endcsname{0}
\expandafter\def\csname v@ep.haiku.k.bad.failure.neutral\endcsname{0}
\expandafter\def\csname v@ep.haiku.k.bad.failure.nx\endcsname{0}
\expandafter\def\csname v@ep.haiku.k.bad.failure.planted\endcsname{0}
\expandafter\def\csname v@ep.haiku.k.bad.failure.rb\endcsname{0}
\expandafter\def\csname v@ep.haiku.k.bad.failure.rbd\endcsname{0}
\expandafter\def\csname v@ep.haiku.k.bad.failure.rbdx\endcsname{0}
\expandafter\def\csname v@ep.haiku.k.bad.failure.rbx\endcsname{0}
\expandafter\def\csname v@ep.haiku.k.bad.success.n\endcsname{0}
\expandafter\def\csname v@ep.haiku.k.bad.success.neutral\endcsname{0}
\expandafter\def\csname v@ep.haiku.k.bad.success.nx\endcsname{0}
\expandafter\def\csname v@ep.haiku.k.bad.success.planted\endcsname{0}
\expandafter\def\csname v@ep.haiku.k.bad.success.rb\endcsname{0}
\expandafter\def\csname v@ep.haiku.k.bad.success.rbd\endcsname{0}
\expandafter\def\csname v@ep.haiku.k.bad.success.rbdx\endcsname{0}
\expandafter\def\csname v@ep.haiku.k.bad.success.rbx\endcsname{0}
\expandafter\def\csname v@ep.haiku.k.good.failure.n\endcsname{4}
\expandafter\def\csname v@ep.haiku.k.good.failure.neutral\endcsname{0}
\expandafter\def\csname v@ep.haiku.k.good.failure.nx\endcsname{3}
\expandafter\def\csname v@ep.haiku.k.good.failure.planted\endcsname{0}
\expandafter\def\csname v@ep.haiku.k.good.failure.rb\endcsname{13}
\expandafter\def\csname v@ep.haiku.k.good.failure.rbd\endcsname{5}
\expandafter\def\csname v@ep.haiku.k.good.failure.rbdx\endcsname{4}
\expandafter\def\csname v@ep.haiku.k.good.failure.rbx\endcsname{2}
\expandafter\def\csname v@ep.haiku.k.good.success.n\endcsname{0}
\expandafter\def\csname v@ep.haiku.k.good.success.neutral\endcsname{0}
\expandafter\def\csname v@ep.haiku.k.good.success.nx\endcsname{0}
\expandafter\def\csname v@ep.haiku.k.good.success.planted\endcsname{0}
\expandafter\def\csname v@ep.haiku.k.good.success.rb\endcsname{1}
\expandafter\def\csname v@ep.haiku.k.good.success.rbd\endcsname{0}
\expandafter\def\csname v@ep.haiku.k.good.success.rbdx\endcsname{1}
\expandafter\def\csname v@ep.haiku.k.good.success.rbx\endcsname{0}
\expandafter\def\csname v@ep.haiku.l2brb.ci\endcsname{[$-0.012$, $+0.004$]}
\expandafter\def\csname v@ep.haiku.l2brb.mean\endcsname{$-0.004$}
\expandafter\def\csname v@ep.haiku.l2brb.n\endcsname{256}
\expandafter\def\csname v@ep.haiku.l2brb.pbraw\endcsname{$1.000$}
\expandafter\def\csname v@ep.haiku.l2brb.praw\endcsname{$0.318$}
\expandafter\def\csname v@ep.haiku.l2rb.ci\endcsname{[$+0.014$, $+0.072$]}
\expandafter\def\csname v@ep.haiku.l2rb.lower\endcsname{$+0.043$}
\expandafter\def\csname v@ep.haiku.l2rb.lowerci\endcsname{[$+0.014$, $+0.072$]}
\expandafter\def\csname v@ep.haiku.l2rb.mean\endcsname{$+0.043$}
\expandafter\def\csname v@ep.haiku.l2rb.n\endcsname{256}
\expandafter\def\csname v@ep.haiku.l2rb.pbraw\endcsname{$0.011$}
\expandafter\def\csname v@ep.haiku.l2rb.praw\endcsname{$0.0043$}
\expandafter\def\csname v@ep.haiku.l2rb.upper\endcsname{$+0.043$}
\expandafter\def\csname v@ep.haiku.l2rb.upperci\endcsname{[$+0.014$, $+0.072$]}
\expandafter\def\csname v@ep.haiku.l3rb.ci\endcsname{[$-0.014$, $+0.014$]}
\expandafter\def\csname v@ep.haiku.l3rb.lower\endcsname{$0.000$}
\expandafter\def\csname v@ep.haiku.l3rb.lowerci\endcsname{[$-0.014$, $+0.014$]}
\expandafter\def\csname v@ep.haiku.l3rb.mean\endcsname{$0.000$}
\expandafter\def\csname v@ep.haiku.l3rb.n\endcsname{256}
\expandafter\def\csname v@ep.haiku.l3rb.pbraw\endcsname{$1.000$}
\expandafter\def\csname v@ep.haiku.l3rb.praw\endcsname{$1.000$}
\expandafter\def\csname v@ep.haiku.l3rb.upper\endcsname{$0.000$}
\expandafter\def\csname v@ep.haiku.l3rb.upperci\endcsname{[$-0.014$, $+0.014$]}
\expandafter\def\csname v@ep.haiku.n.bad.failure.n\endcsname{256}
\expandafter\def\csname v@ep.haiku.n.bad.failure.neutral\endcsname{256}
\expandafter\def\csname v@ep.haiku.n.bad.failure.nx\endcsname{256}
\expandafter\def\csname v@ep.haiku.n.bad.failure.planted\endcsname{256}
\expandafter\def\csname v@ep.haiku.n.bad.failure.rb\endcsname{256}
\expandafter\def\csname v@ep.haiku.n.bad.failure.rbd\endcsname{256}
\expandafter\def\csname v@ep.haiku.n.bad.failure.rbdx\endcsname{256}
\expandafter\def\csname v@ep.haiku.n.bad.failure.rbx\endcsname{256}
\expandafter\def\csname v@ep.haiku.n.bad.success.n\endcsname{256}
\expandafter\def\csname v@ep.haiku.n.bad.success.neutral\endcsname{256}
\expandafter\def\csname v@ep.haiku.n.bad.success.nx\endcsname{256}
\expandafter\def\csname v@ep.haiku.n.bad.success.planted\endcsname{256}
\expandafter\def\csname v@ep.haiku.n.bad.success.rb\endcsname{256}
\expandafter\def\csname v@ep.haiku.n.bad.success.rbd\endcsname{256}
\expandafter\def\csname v@ep.haiku.n.bad.success.rbdx\endcsname{256}
\expandafter\def\csname v@ep.haiku.n.bad.success.rbx\endcsname{256}
\expandafter\def\csname v@ep.haiku.n.good.failure.n\endcsname{256}
\expandafter\def\csname v@ep.haiku.n.good.failure.neutral\endcsname{256}
\expandafter\def\csname v@ep.haiku.n.good.failure.nx\endcsname{256}
\expandafter\def\csname v@ep.haiku.n.good.failure.planted\endcsname{256}
\expandafter\def\csname v@ep.haiku.n.good.failure.rb\endcsname{256}
\expandafter\def\csname v@ep.haiku.n.good.failure.rbd\endcsname{256}
\expandafter\def\csname v@ep.haiku.n.good.failure.rbdx\endcsname{256}
\expandafter\def\csname v@ep.haiku.n.good.failure.rbx\endcsname{256}
\expandafter\def\csname v@ep.haiku.n.good.success.n\endcsname{256}
\expandafter\def\csname v@ep.haiku.n.good.success.neutral\endcsname{256}
\expandafter\def\csname v@ep.haiku.n.good.success.nx\endcsname{256}
\expandafter\def\csname v@ep.haiku.n.good.success.planted\endcsname{256}
\expandafter\def\csname v@ep.haiku.n.good.success.rb\endcsname{256}
\expandafter\def\csname v@ep.haiku.n.good.success.rbd\endcsname{256}
\expandafter\def\csname v@ep.haiku.n.good.success.rbdx\endcsname{256}
\expandafter\def\csname v@ep.haiku.n.good.success.rbx\endcsname{256}
\expandafter\def\csname v@ep.haiku.name\endcsname{Claude Haiku 5.5}
\expandafter\def\csname v@ep.haiku.namesworse\endcsname{10\%}
\expandafter\def\csname v@ep.haiku.p2rb.ci\endcsname{[$+0.019$, $+0.075$]}
\expandafter\def\csname v@ep.haiku.p2rb.lower\endcsname{$+0.047$}
\expandafter\def\csname v@ep.haiku.p2rb.lowerci\endcsname{[$+0.019$, $+0.075$]}
\expandafter\def\csname v@ep.haiku.p2rb.mean\endcsname{$+0.047$}
\expandafter\def\csname v@ep.haiku.p2rb.n\endcsname{256}
\expandafter\def\csname v@ep.haiku.p2rb.pb\endcsname{$0.051$}
\expandafter\def\csname v@ep.haiku.p2rb.pbraw\endcsname{$0.025$}
\expandafter\def\csname v@ep.haiku.p2rb.praw\endcsname{$0.0012$}
\expandafter\def\csname v@ep.haiku.p2rb.pt\endcsname{$0.0025$}
\expandafter\def\csname v@ep.haiku.p2rb.upper\endcsname{$+0.047$}
\expandafter\def\csname v@ep.haiku.p2rb.upperci\endcsname{[$+0.019$, $+0.075$]}
\expandafter\def\csname v@ep.haiku.p3mrb.ci\endcsname{[$-0.029$, $+0.029$]}
\expandafter\def\csname v@ep.haiku.p3mrb.lower\endcsname{$0.000$}
\expandafter\def\csname v@ep.haiku.p3mrb.lowerci\endcsname{[$-0.029$, $+0.029$]}
\expandafter\def\csname v@ep.haiku.p3mrb.mean\endcsname{$0.000$}
\expandafter\def\csname v@ep.haiku.p3mrb.n\endcsname{256}
\expandafter\def\csname v@ep.haiku.p3mrb.pb\endcsname{$1.000$}
\expandafter\def\csname v@ep.haiku.p3mrb.pbraw\endcsname{$1.000$}
\expandafter\def\csname v@ep.haiku.p3mrb.praw\endcsname{$1.000$}
\expandafter\def\csname v@ep.haiku.p3mrb.pt\endcsname{$1.000$}
\expandafter\def\csname v@ep.haiku.p3mrb.upper\endcsname{$0.000$}
\expandafter\def\csname v@ep.haiku.p3mrb.upperci\endcsname{[$-0.029$, $+0.029$]}
\expandafter\def\csname v@ep.haiku.planted.k\endcsname{0}
\expandafter\def\csname v@ep.haiku.planted.n\endcsname{1,024}
\expandafter\def\csname v@ep.haiku.rate.bad.failure.n\endcsname{0.0\%}
\expandafter\def\csname v@ep.haiku.rate.bad.failure.neutral\endcsname{0.0\%}
\expandafter\def\csname v@ep.haiku.rate.bad.failure.nx\endcsname{0.0\%}
\expandafter\def\csname v@ep.haiku.rate.bad.failure.planted\endcsname{0.0\%}
\expandafter\def\csname v@ep.haiku.rate.bad.failure.rb\endcsname{0.0\%}
\expandafter\def\csname v@ep.haiku.rate.bad.failure.rbd\endcsname{0.0\%}
\expandafter\def\csname v@ep.haiku.rate.bad.failure.rbdx\endcsname{0.0\%}
\expandafter\def\csname v@ep.haiku.rate.bad.failure.rbx\endcsname{0.0\%}
\expandafter\def\csname v@ep.haiku.rate.bad.success.n\endcsname{0.0\%}
\expandafter\def\csname v@ep.haiku.rate.bad.success.neutral\endcsname{0.0\%}
\expandafter\def\csname v@ep.haiku.rate.bad.success.nx\endcsname{0.0\%}
\expandafter\def\csname v@ep.haiku.rate.bad.success.planted\endcsname{0.0\%}
\expandafter\def\csname v@ep.haiku.rate.bad.success.rb\endcsname{0.0\%}
\expandafter\def\csname v@ep.haiku.rate.bad.success.rbd\endcsname{0.0\%}
\expandafter\def\csname v@ep.haiku.rate.bad.success.rbdx\endcsname{0.0\%}
\expandafter\def\csname v@ep.haiku.rate.bad.success.rbx\endcsname{0.0\%}
\expandafter\def\csname v@ep.haiku.rate.good.failure.n\endcsname{1.6\%}
\expandafter\def\csname v@ep.haiku.rate.good.failure.neutral\endcsname{0.0\%}
\expandafter\def\csname v@ep.haiku.rate.good.failure.nx\endcsname{1.2\%}
\expandafter\def\csname v@ep.haiku.rate.good.failure.planted\endcsname{0.0\%}
\expandafter\def\csname v@ep.haiku.rate.good.failure.rb\endcsname{5.1\%}
\expandafter\def\csname v@ep.haiku.rate.good.failure.rbd\endcsname{2.0\%}
\expandafter\def\csname v@ep.haiku.rate.good.failure.rbdx\endcsname{1.6\%}
\expandafter\def\csname v@ep.haiku.rate.good.failure.rbx\endcsname{0.8\%}
\expandafter\def\csname v@ep.haiku.rate.good.success.n\endcsname{0.0\%}
\expandafter\def\csname v@ep.haiku.rate.good.success.neutral\endcsname{0.0\%}
\expandafter\def\csname v@ep.haiku.rate.good.success.nx\endcsname{0.0\%}
\expandafter\def\csname v@ep.haiku.rate.good.success.planted\endcsname{0.0\%}
\expandafter\def\csname v@ep.haiku.rate.good.success.rb\endcsname{0.4\%}
\expandafter\def\csname v@ep.haiku.rate.good.success.rbd\endcsname{0.0\%}
\expandafter\def\csname v@ep.haiku.rate.good.success.rbdx\endcsname{0.4\%}
\expandafter\def\csname v@ep.haiku.rate.good.success.rbx\endcsname{0.0\%}
\expandafter\def\csname v@ep.haiku.sus.ci\endcsname{[$-0.014$, $+0.014$]}
\expandafter\def\csname v@ep.haiku.sus.mean\endcsname{$0.000$}
\expandafter\def\csname v@ep.haiku.sus.n\endcsname{256}
\expandafter\def\csname v@ep.haiku.sus.pbraw\endcsname{$1.000$}
\expandafter\def\csname v@ep.haiku.sus.praw\endcsname{$1.000$}
\expandafter\def\csname v@ep.haiku.word.pthree\endcsname{\emph{bounded below 0.10}}
\expandafter\def\csname v@ep.haiku.word.ptwo\endcsname{\emph{bounded below 0.10}}
\expandafter\def\csname v@ep.haiku.word.sus\endcsname{\emph{increment below 0.10}}
\expandafter\def\csname v@ep.kimi.a2rb.ci\endcsname{[$+0.084$, $+0.166$]}
\expandafter\def\csname v@ep.kimi.a2rb.lower\endcsname{$+0.125$}
\expandafter\def\csname v@ep.kimi.a2rb.lowerci\endcsname{[$+0.084$, $+0.166$]}
\expandafter\def\csname v@ep.kimi.a2rb.mean\endcsname{$+0.125$}
\expandafter\def\csname v@ep.kimi.a2rb.n\endcsname{256}
\expandafter\def\csname v@ep.kimi.a2rb.pbraw\endcsname{$1.7\times10^{-8}$}
\expandafter\def\csname v@ep.kimi.a2rb.praw\endcsname{$5.6\times10^{-9}$}
\expandafter\def\csname v@ep.kimi.a2rb.upper\endcsname{$+0.125$}
\expandafter\def\csname v@ep.kimi.a2rb.upperci\endcsname{[$+0.084$, $+0.166$]}
\expandafter\def\csname v@ep.kimi.a3rb.ci\endcsname{[$-0.014$, $+0.014$]}
\expandafter\def\csname v@ep.kimi.a3rb.lower\endcsname{$0.000$}
\expandafter\def\csname v@ep.kimi.a3rb.lowerci\endcsname{[$-0.014$, $+0.014$]}
\expandafter\def\csname v@ep.kimi.a3rb.mean\endcsname{$0.000$}
\expandafter\def\csname v@ep.kimi.a3rb.n\endcsname{256}
\expandafter\def\csname v@ep.kimi.a3rb.pbraw\endcsname{$1.000$}
\expandafter\def\csname v@ep.kimi.a3rb.praw\endcsname{$1.000$}
\expandafter\def\csname v@ep.kimi.a3rb.upper\endcsname{$0.000$}
\expandafter\def\csname v@ep.kimi.a3rb.upperci\endcsname{[$-0.014$, $+0.014$]}
\expandafter\def\csname v@ep.kimi.count.bad.failure.n\endcsname{0/256}
\expandafter\def\csname v@ep.kimi.count.bad.failure.neutral\endcsname{0/256}
\expandafter\def\csname v@ep.kimi.count.bad.failure.nx\endcsname{0/256}
\expandafter\def\csname v@ep.kimi.count.bad.failure.planted\endcsname{0/256}
\expandafter\def\csname v@ep.kimi.count.bad.failure.rb\endcsname{0/256}
\expandafter\def\csname v@ep.kimi.count.bad.failure.rbd\endcsname{0/256}
\expandafter\def\csname v@ep.kimi.count.bad.failure.rbdx\endcsname{0/256}
\expandafter\def\csname v@ep.kimi.count.bad.failure.rbx\endcsname{0/256}
\expandafter\def\csname v@ep.kimi.count.bad.success.n\endcsname{0/256}
\expandafter\def\csname v@ep.kimi.count.bad.success.neutral\endcsname{0/256}
\expandafter\def\csname v@ep.kimi.count.bad.success.nx\endcsname{0/256}
\expandafter\def\csname v@ep.kimi.count.bad.success.planted\endcsname{0/256}
\expandafter\def\csname v@ep.kimi.count.bad.success.rb\endcsname{0/256}
\expandafter\def\csname v@ep.kimi.count.bad.success.rbd\endcsname{0/256}
\expandafter\def\csname v@ep.kimi.count.bad.success.rbdx\endcsname{0/256}
\expandafter\def\csname v@ep.kimi.count.bad.success.rbx\endcsname{0/256}
\expandafter\def\csname v@ep.kimi.count.good.failure.n\endcsname{1/256}
\expandafter\def\csname v@ep.kimi.count.good.failure.neutral\endcsname{0/256}
\expandafter\def\csname v@ep.kimi.count.good.failure.nx\endcsname{0/255}
\expandafter\def\csname v@ep.kimi.count.good.failure.planted\endcsname{0/256}
\expandafter\def\csname v@ep.kimi.count.good.failure.rb\endcsname{32/256}
\expandafter\def\csname v@ep.kimi.count.good.failure.rbd\endcsname{27/256}
\expandafter\def\csname v@ep.kimi.count.good.failure.rbdx\endcsname{0/256}
\expandafter\def\csname v@ep.kimi.count.good.failure.rbx\endcsname{0/256}
\expandafter\def\csname v@ep.kimi.count.good.success.n\endcsname{0/256}
\expandafter\def\csname v@ep.kimi.count.good.success.neutral\endcsname{0/256}
\expandafter\def\csname v@ep.kimi.count.good.success.nx\endcsname{0/256}
\expandafter\def\csname v@ep.kimi.count.good.success.planted\endcsname{0/256}
\expandafter\def\csname v@ep.kimi.count.good.success.rb\endcsname{0/256}
\expandafter\def\csname v@ep.kimi.count.good.success.rbd\endcsname{0/256}
\expandafter\def\csname v@ep.kimi.count.good.success.rbdx\endcsname{0/256}
\expandafter\def\csname v@ep.kimi.count.good.success.rbx\endcsname{2/256}
\expandafter\def\csname v@ep.kimi.id\endcsname{\texttt{moonshotai/kimi-k3}}
\expandafter\def\csname v@ep.kimi.k.bad.failure.n\endcsname{0}
\expandafter\def\csname v@ep.kimi.k.bad.failure.neutral\endcsname{0}
\expandafter\def\csname v@ep.kimi.k.bad.failure.nx\endcsname{0}
\expandafter\def\csname v@ep.kimi.k.bad.failure.planted\endcsname{0}
\expandafter\def\csname v@ep.kimi.k.bad.failure.rb\endcsname{0}
\expandafter\def\csname v@ep.kimi.k.bad.failure.rbd\endcsname{0}
\expandafter\def\csname v@ep.kimi.k.bad.failure.rbdx\endcsname{0}
\expandafter\def\csname v@ep.kimi.k.bad.failure.rbx\endcsname{0}
\expandafter\def\csname v@ep.kimi.k.bad.success.n\endcsname{0}
\expandafter\def\csname v@ep.kimi.k.bad.success.neutral\endcsname{0}
\expandafter\def\csname v@ep.kimi.k.bad.success.nx\endcsname{0}
\expandafter\def\csname v@ep.kimi.k.bad.success.planted\endcsname{0}
\expandafter\def\csname v@ep.kimi.k.bad.success.rb\endcsname{0}
\expandafter\def\csname v@ep.kimi.k.bad.success.rbd\endcsname{0}
\expandafter\def\csname v@ep.kimi.k.bad.success.rbdx\endcsname{0}
\expandafter\def\csname v@ep.kimi.k.bad.success.rbx\endcsname{0}
\expandafter\def\csname v@ep.kimi.k.good.failure.n\endcsname{1}
\expandafter\def\csname v@ep.kimi.k.good.failure.neutral\endcsname{0}
\expandafter\def\csname v@ep.kimi.k.good.failure.nx\endcsname{0}
\expandafter\def\csname v@ep.kimi.k.good.failure.planted\endcsname{0}
\expandafter\def\csname v@ep.kimi.k.good.failure.rb\endcsname{32}
\expandafter\def\csname v@ep.kimi.k.good.failure.rbd\endcsname{27}
\expandafter\def\csname v@ep.kimi.k.good.failure.rbdx\endcsname{0}
\expandafter\def\csname v@ep.kimi.k.good.failure.rbx\endcsname{0}
\expandafter\def\csname v@ep.kimi.k.good.success.n\endcsname{0}
\expandafter\def\csname v@ep.kimi.k.good.success.neutral\endcsname{0}
\expandafter\def\csname v@ep.kimi.k.good.success.nx\endcsname{0}
\expandafter\def\csname v@ep.kimi.k.good.success.planted\endcsname{0}
\expandafter\def\csname v@ep.kimi.k.good.success.rb\endcsname{0}
\expandafter\def\csname v@ep.kimi.k.good.success.rbd\endcsname{0}
\expandafter\def\csname v@ep.kimi.k.good.success.rbdx\endcsname{0}
\expandafter\def\csname v@ep.kimi.k.good.success.rbx\endcsname{2}
\expandafter\def\csname v@ep.kimi.l2brb.ci\endcsname{[$-0.003$, $+0.019$]}
\expandafter\def\csname v@ep.kimi.l2brb.mean\endcsname{$+0.008$}
\expandafter\def\csname v@ep.kimi.l2brb.n\endcsname{256}
\expandafter\def\csname v@ep.kimi.l2brb.pbraw\endcsname{$0.327$}
\expandafter\def\csname v@ep.kimi.l2brb.praw\endcsname{$0.158$}
\expandafter\def\csname v@ep.kimi.l2rb.ci\endcsname{[$+0.084$, $+0.166$]}
\expandafter\def\csname v@ep.kimi.l2rb.lower\endcsname{$+0.125$}
\expandafter\def\csname v@ep.kimi.l2rb.lowerci\endcsname{[$+0.084$, $+0.166$]}
\expandafter\def\csname v@ep.kimi.l2rb.mean\endcsname{$+0.125$}
\expandafter\def\csname v@ep.kimi.l2rb.n\endcsname{256}
\expandafter\def\csname v@ep.kimi.l2rb.pbraw\endcsname{$1.7\times10^{-8}$}
\expandafter\def\csname v@ep.kimi.l2rb.praw\endcsname{$5.6\times10^{-9}$}
\expandafter\def\csname v@ep.kimi.l2rb.upper\endcsname{$+0.125$}
\expandafter\def\csname v@ep.kimi.l2rb.upperci\endcsname{[$+0.084$, $+0.166$]}
\expandafter\def\csname v@ep.kimi.l3rb.ci\endcsname{[$-0.014$, $+0.014$]}
\expandafter\def\csname v@ep.kimi.l3rb.lower\endcsname{$0.000$}
\expandafter\def\csname v@ep.kimi.l3rb.lowerci\endcsname{[$-0.014$, $+0.014$]}
\expandafter\def\csname v@ep.kimi.l3rb.mean\endcsname{$0.000$}
\expandafter\def\csname v@ep.kimi.l3rb.n\endcsname{256}
\expandafter\def\csname v@ep.kimi.l3rb.pbraw\endcsname{$1.000$}
\expandafter\def\csname v@ep.kimi.l3rb.praw\endcsname{$1.000$}
\expandafter\def\csname v@ep.kimi.l3rb.upper\endcsname{$0.000$}
\expandafter\def\csname v@ep.kimi.l3rb.upperci\endcsname{[$-0.014$, $+0.014$]}
\expandafter\def\csname v@ep.kimi.n.bad.failure.n\endcsname{256}
\expandafter\def\csname v@ep.kimi.n.bad.failure.neutral\endcsname{256}
\expandafter\def\csname v@ep.kimi.n.bad.failure.nx\endcsname{256}
\expandafter\def\csname v@ep.kimi.n.bad.failure.planted\endcsname{256}
\expandafter\def\csname v@ep.kimi.n.bad.failure.rb\endcsname{256}
\expandafter\def\csname v@ep.kimi.n.bad.failure.rbd\endcsname{256}
\expandafter\def\csname v@ep.kimi.n.bad.failure.rbdx\endcsname{256}
\expandafter\def\csname v@ep.kimi.n.bad.failure.rbx\endcsname{256}
\expandafter\def\csname v@ep.kimi.n.bad.success.n\endcsname{256}
\expandafter\def\csname v@ep.kimi.n.bad.success.neutral\endcsname{256}
\expandafter\def\csname v@ep.kimi.n.bad.success.nx\endcsname{256}
\expandafter\def\csname v@ep.kimi.n.bad.success.planted\endcsname{256}
\expandafter\def\csname v@ep.kimi.n.bad.success.rb\endcsname{256}
\expandafter\def\csname v@ep.kimi.n.bad.success.rbd\endcsname{256}
\expandafter\def\csname v@ep.kimi.n.bad.success.rbdx\endcsname{256}
\expandafter\def\csname v@ep.kimi.n.bad.success.rbx\endcsname{256}
\expandafter\def\csname v@ep.kimi.n.good.failure.n\endcsname{256}
\expandafter\def\csname v@ep.kimi.n.good.failure.neutral\endcsname{256}
\expandafter\def\csname v@ep.kimi.n.good.failure.nx\endcsname{255}
\expandafter\def\csname v@ep.kimi.n.good.failure.planted\endcsname{256}
\expandafter\def\csname v@ep.kimi.n.good.failure.rb\endcsname{256}
\expandafter\def\csname v@ep.kimi.n.good.failure.rbd\endcsname{256}
\expandafter\def\csname v@ep.kimi.n.good.failure.rbdx\endcsname{256}
\expandafter\def\csname v@ep.kimi.n.good.failure.rbx\endcsname{256}
\expandafter\def\csname v@ep.kimi.n.good.success.n\endcsname{256}
\expandafter\def\csname v@ep.kimi.n.good.success.neutral\endcsname{256}
\expandafter\def\csname v@ep.kimi.n.good.success.nx\endcsname{256}
\expandafter\def\csname v@ep.kimi.n.good.success.planted\endcsname{256}
\expandafter\def\csname v@ep.kimi.n.good.success.rb\endcsname{256}
\expandafter\def\csname v@ep.kimi.n.good.success.rbd\endcsname{256}
\expandafter\def\csname v@ep.kimi.n.good.success.rbdx\endcsname{256}
\expandafter\def\csname v@ep.kimi.n.good.success.rbx\endcsname{256}
\expandafter\def\csname v@ep.kimi.name\endcsname{Kimi K3}
\expandafter\def\csname v@ep.kimi.namesworse\endcsname{29\%}
\expandafter\def\csname v@ep.kimi.p2rb.ci\endcsname{[$+0.084$, $+0.166$]}
\expandafter\def\csname v@ep.kimi.p2rb.lower\endcsname{$+0.125$}
\expandafter\def\csname v@ep.kimi.p2rb.lowerci\endcsname{[$+0.084$, $+0.166$]}
\expandafter\def\csname v@ep.kimi.p2rb.mean\endcsname{$+0.125$}
\expandafter\def\csname v@ep.kimi.p2rb.n\endcsname{256}
\expandafter\def\csname v@ep.kimi.p2rb.pb\endcsname{$7.5\times10^{-5}$}
\expandafter\def\csname v@ep.kimi.p2rb.pbraw\endcsname{$3.8\times10^{-5}$}
\expandafter\def\csname v@ep.kimi.p2rb.praw\endcsname{$5.6\times10^{-9}$}
\expandafter\def\csname v@ep.kimi.p2rb.pt\endcsname{$1.1\times10^{-8}$}
\expandafter\def\csname v@ep.kimi.p2rb.upper\endcsname{$+0.125$}
\expandafter\def\csname v@ep.kimi.p2rb.upperci\endcsname{[$+0.084$, $+0.166$]}
\expandafter\def\csname v@ep.kimi.p3mrb.ci\endcsname{[$-0.029$, $+0.029$]}
\expandafter\def\csname v@ep.kimi.p3mrb.lower\endcsname{$0.000$}
\expandafter\def\csname v@ep.kimi.p3mrb.lowerci\endcsname{[$-0.029$, $+0.029$]}
\expandafter\def\csname v@ep.kimi.p3mrb.mean\endcsname{$0.000$}
\expandafter\def\csname v@ep.kimi.p3mrb.n\endcsname{256}
\expandafter\def\csname v@ep.kimi.p3mrb.pb\endcsname{$1.000$}
\expandafter\def\csname v@ep.kimi.p3mrb.pbraw\endcsname{$1.000$}
\expandafter\def\csname v@ep.kimi.p3mrb.praw\endcsname{$1.000$}
\expandafter\def\csname v@ep.kimi.p3mrb.pt\endcsname{$1.000$}
\expandafter\def\csname v@ep.kimi.p3mrb.upper\endcsname{$0.000$}
\expandafter\def\csname v@ep.kimi.p3mrb.upperci\endcsname{[$-0.029$, $+0.029$]}
\expandafter\def\csname v@ep.kimi.planted.k\endcsname{0}
\expandafter\def\csname v@ep.kimi.planted.n\endcsname{1,024}
\expandafter\def\csname v@ep.kimi.rate.bad.failure.n\endcsname{0.0\%}
\expandafter\def\csname v@ep.kimi.rate.bad.failure.neutral\endcsname{0.0\%}
\expandafter\def\csname v@ep.kimi.rate.bad.failure.nx\endcsname{0.0\%}
\expandafter\def\csname v@ep.kimi.rate.bad.failure.planted\endcsname{0.0\%}
\expandafter\def\csname v@ep.kimi.rate.bad.failure.rb\endcsname{0.0\%}
\expandafter\def\csname v@ep.kimi.rate.bad.failure.rbd\endcsname{0.0\%}
\expandafter\def\csname v@ep.kimi.rate.bad.failure.rbdx\endcsname{0.0\%}
\expandafter\def\csname v@ep.kimi.rate.bad.failure.rbx\endcsname{0.0\%}
\expandafter\def\csname v@ep.kimi.rate.bad.success.n\endcsname{0.0\%}
\expandafter\def\csname v@ep.kimi.rate.bad.success.neutral\endcsname{0.0\%}
\expandafter\def\csname v@ep.kimi.rate.bad.success.nx\endcsname{0.0\%}
\expandafter\def\csname v@ep.kimi.rate.bad.success.planted\endcsname{0.0\%}
\expandafter\def\csname v@ep.kimi.rate.bad.success.rb\endcsname{0.0\%}
\expandafter\def\csname v@ep.kimi.rate.bad.success.rbd\endcsname{0.0\%}
\expandafter\def\csname v@ep.kimi.rate.bad.success.rbdx\endcsname{0.0\%}
\expandafter\def\csname v@ep.kimi.rate.bad.success.rbx\endcsname{0.0\%}
\expandafter\def\csname v@ep.kimi.rate.good.failure.n\endcsname{0.4\%}
\expandafter\def\csname v@ep.kimi.rate.good.failure.neutral\endcsname{0.0\%}
\expandafter\def\csname v@ep.kimi.rate.good.failure.nx\endcsname{0.0\%}
\expandafter\def\csname v@ep.kimi.rate.good.failure.planted\endcsname{0.0\%}
\expandafter\def\csname v@ep.kimi.rate.good.failure.rb\endcsname{12.5\%}
\expandafter\def\csname v@ep.kimi.rate.good.failure.rbd\endcsname{10.5\%}
\expandafter\def\csname v@ep.kimi.rate.good.failure.rbdx\endcsname{0.0\%}
\expandafter\def\csname v@ep.kimi.rate.good.failure.rbx\endcsname{0.0\%}
\expandafter\def\csname v@ep.kimi.rate.good.success.n\endcsname{0.0\%}
\expandafter\def\csname v@ep.kimi.rate.good.success.neutral\endcsname{0.0\%}
\expandafter\def\csname v@ep.kimi.rate.good.success.nx\endcsname{0.0\%}
\expandafter\def\csname v@ep.kimi.rate.good.success.planted\endcsname{0.0\%}
\expandafter\def\csname v@ep.kimi.rate.good.success.rb\endcsname{0.0\%}
\expandafter\def\csname v@ep.kimi.rate.good.success.rbd\endcsname{0.0\%}
\expandafter\def\csname v@ep.kimi.rate.good.success.rbdx\endcsname{0.0\%}
\expandafter\def\csname v@ep.kimi.rate.good.success.rbx\endcsname{0.8\%}
\expandafter\def\csname v@ep.kimi.sus.ci\endcsname{[$-0.014$, $+0.014$]}
\expandafter\def\csname v@ep.kimi.sus.mean\endcsname{$0.000$}
\expandafter\def\csname v@ep.kimi.sus.n\endcsname{256}
\expandafter\def\csname v@ep.kimi.sus.pbraw\endcsname{$1.000$}
\expandafter\def\csname v@ep.kimi.sus.praw\endcsname{$1.000$}
\expandafter\def\csname v@ep.kimi.word.pthree\endcsname{\emph{bounded below 0.10}}
\expandafter\def\csname v@ep.kimi.word.ptwo\endcsname{\emph{luck-carrying}}
\expandafter\def\csname v@ep.kimi.word.sus\endcsname{\emph{increment below 0.10}}
\expandafter\def\csname v@ep.list.a2l2robust.notluck\endcsname{Claude Haiku 5.5 and GPT-6 Luna}
\expandafter\def\csname v@ep.list.allbetter\endcsname{Claude Opus 5.5, Claude Sonnet 5.5 and GPT-6.1 Sol}
\expandafter\def\csname v@ep.list.bounded\endcsname{Claude Opus 5.5, Claude Sonnet 5.5, Claude Haiku 5.5, GPT-6.1 Sol and GPT-6 Luna}
\expandafter\def\csname v@ep.list.boundedpos\endcsname{Claude Haiku 5.5 and GPT-6 Luna}
\expandafter\def\csname v@ep.list.luck\endcsname{Gemini 3.8 Flash, Grok 4.7, Qwen 3.8 Max, DeepSeek V4 Pro, Mistral Large 4 and Kimi K3}
\expandafter\def\csname v@ep.list.susabove\endcsname{Gemini 3.8 Flash, Qwen 3.8 Max, DeepSeek V4 Pro and Mistral Large 4}
\expandafter\def\csname v@ep.luna.a2rb.ci\endcsname{[$+0.010$, $+0.053$]}
\expandafter\def\csname v@ep.luna.a2rb.lower\endcsname{$+0.031$}
\expandafter\def\csname v@ep.luna.a2rb.lowerci\endcsname{[$+0.010$, $+0.053$]}
\expandafter\def\csname v@ep.luna.a2rb.mean\endcsname{$+0.031$}
\expandafter\def\csname v@ep.luna.a2rb.n\endcsname{256}
\expandafter\def\csname v@ep.luna.a2rb.pbraw\endcsname{$0.011$}
\expandafter\def\csname v@ep.luna.a2rb.praw\endcsname{$0.0045$}
\expandafter\def\csname v@ep.luna.a2rb.upper\endcsname{$+0.031$}
\expandafter\def\csname v@ep.luna.a2rb.upperci\endcsname{[$+0.010$, $+0.053$]}
\expandafter\def\csname v@ep.luna.a3rb.ci\endcsname{[$-0.014$, $+0.014$]}
\expandafter\def\csname v@ep.luna.a3rb.lower\endcsname{$0.000$}
\expandafter\def\csname v@ep.luna.a3rb.lowerci\endcsname{[$-0.014$, $+0.014$]}
\expandafter\def\csname v@ep.luna.a3rb.mean\endcsname{$0.000$}
\expandafter\def\csname v@ep.luna.a3rb.n\endcsname{256}
\expandafter\def\csname v@ep.luna.a3rb.pbraw\endcsname{$1.000$}
\expandafter\def\csname v@ep.luna.a3rb.praw\endcsname{$1.000$}
\expandafter\def\csname v@ep.luna.a3rb.upper\endcsname{$0.000$}
\expandafter\def\csname v@ep.luna.a3rb.upperci\endcsname{[$-0.014$, $+0.014$]}
\expandafter\def\csname v@ep.luna.count.bad.failure.n\endcsname{0/256}
\expandafter\def\csname v@ep.luna.count.bad.failure.neutral\endcsname{0/256}
\expandafter\def\csname v@ep.luna.count.bad.failure.nx\endcsname{0/256}
\expandafter\def\csname v@ep.luna.count.bad.failure.planted\endcsname{0/256}
\expandafter\def\csname v@ep.luna.count.bad.failure.rb\endcsname{0/256}
\expandafter\def\csname v@ep.luna.count.bad.failure.rbd\endcsname{0/256}
\expandafter\def\csname v@ep.luna.count.bad.failure.rbdx\endcsname{0/256}
\expandafter\def\csname v@ep.luna.count.bad.failure.rbx\endcsname{0/256}
\expandafter\def\csname v@ep.luna.count.bad.success.n\endcsname{0/256}
\expandafter\def\csname v@ep.luna.count.bad.success.neutral\endcsname{0/256}
\expandafter\def\csname v@ep.luna.count.bad.success.nx\endcsname{0/256}
\expandafter\def\csname v@ep.luna.count.bad.success.planted\endcsname{0/256}
\expandafter\def\csname v@ep.luna.count.bad.success.rb\endcsname{0/256}
\expandafter\def\csname v@ep.luna.count.bad.success.rbd\endcsname{0/256}
\expandafter\def\csname v@ep.luna.count.bad.success.rbdx\endcsname{0/256}
\expandafter\def\csname v@ep.luna.count.bad.success.rbx\endcsname{0/256}
\expandafter\def\csname v@ep.luna.count.good.failure.n\endcsname{0/256}
\expandafter\def\csname v@ep.luna.count.good.failure.neutral\endcsname{0/256}
\expandafter\def\csname v@ep.luna.count.good.failure.nx\endcsname{0/256}
\expandafter\def\csname v@ep.luna.count.good.failure.planted\endcsname{4/256}
\expandafter\def\csname v@ep.luna.count.good.failure.rb\endcsname{8/256}
\expandafter\def\csname v@ep.luna.count.good.failure.rbd\endcsname{1/256}
\expandafter\def\csname v@ep.luna.count.good.failure.rbdx\endcsname{0/256}
\expandafter\def\csname v@ep.luna.count.good.failure.rbx\endcsname{0/256}
\expandafter\def\csname v@ep.luna.count.good.success.n\endcsname{0/256}
\expandafter\def\csname v@ep.luna.count.good.success.neutral\endcsname{0/256}
\expandafter\def\csname v@ep.luna.count.good.success.nx\endcsname{0/256}
\expandafter\def\csname v@ep.luna.count.good.success.planted\endcsname{0/256}
\expandafter\def\csname v@ep.luna.count.good.success.rb\endcsname{0/256}
\expandafter\def\csname v@ep.luna.count.good.success.rbd\endcsname{0/256}
\expandafter\def\csname v@ep.luna.count.good.success.rbdx\endcsname{0/256}
\expandafter\def\csname v@ep.luna.count.good.success.rbx\endcsname{0/256}
\expandafter\def\csname v@ep.luna.id\endcsname{\texttt{openai/gpt-6-luna}}
\expandafter\def\csname v@ep.luna.k.bad.failure.n\endcsname{0}
\expandafter\def\csname v@ep.luna.k.bad.failure.neutral\endcsname{0}
\expandafter\def\csname v@ep.luna.k.bad.failure.nx\endcsname{0}
\expandafter\def\csname v@ep.luna.k.bad.failure.planted\endcsname{0}
\expandafter\def\csname v@ep.luna.k.bad.failure.rb\endcsname{0}
\expandafter\def\csname v@ep.luna.k.bad.failure.rbd\endcsname{0}
\expandafter\def\csname v@ep.luna.k.bad.failure.rbdx\endcsname{0}
\expandafter\def\csname v@ep.luna.k.bad.failure.rbx\endcsname{0}
\expandafter\def\csname v@ep.luna.k.bad.success.n\endcsname{0}
\expandafter\def\csname v@ep.luna.k.bad.success.neutral\endcsname{0}
\expandafter\def\csname v@ep.luna.k.bad.success.nx\endcsname{0}
\expandafter\def\csname v@ep.luna.k.bad.success.planted\endcsname{0}
\expandafter\def\csname v@ep.luna.k.bad.success.rb\endcsname{0}
\expandafter\def\csname v@ep.luna.k.bad.success.rbd\endcsname{0}
\expandafter\def\csname v@ep.luna.k.bad.success.rbdx\endcsname{0}
\expandafter\def\csname v@ep.luna.k.bad.success.rbx\endcsname{0}
\expandafter\def\csname v@ep.luna.k.good.failure.n\endcsname{0}
\expandafter\def\csname v@ep.luna.k.good.failure.neutral\endcsname{0}
\expandafter\def\csname v@ep.luna.k.good.failure.nx\endcsname{0}
\expandafter\def\csname v@ep.luna.k.good.failure.planted\endcsname{4}
\expandafter\def\csname v@ep.luna.k.good.failure.rb\endcsname{8}
\expandafter\def\csname v@ep.luna.k.good.failure.rbd\endcsname{1}
\expandafter\def\csname v@ep.luna.k.good.failure.rbdx\endcsname{0}
\expandafter\def\csname v@ep.luna.k.good.failure.rbx\endcsname{0}
\expandafter\def\csname v@ep.luna.k.good.success.n\endcsname{0}
\expandafter\def\csname v@ep.luna.k.good.success.neutral\endcsname{0}
\expandafter\def\csname v@ep.luna.k.good.success.nx\endcsname{0}
\expandafter\def\csname v@ep.luna.k.good.success.planted\endcsname{0}
\expandafter\def\csname v@ep.luna.k.good.success.rb\endcsname{0}
\expandafter\def\csname v@ep.luna.k.good.success.rbd\endcsname{0}
\expandafter\def\csname v@ep.luna.k.good.success.rbdx\endcsname{0}
\expandafter\def\csname v@ep.luna.k.good.success.rbx\endcsname{0}
\expandafter\def\csname v@ep.luna.l2brb.ci\endcsname{[$-0.014$, $+0.014$]}
\expandafter\def\csname v@ep.luna.l2brb.mean\endcsname{$0.000$}
\expandafter\def\csname v@ep.luna.l2brb.n\endcsname{256}
\expandafter\def\csname v@ep.luna.l2brb.pbraw\endcsname{$1.000$}
\expandafter\def\csname v@ep.luna.l2brb.praw\endcsname{$1.000$}
\expandafter\def\csname v@ep.luna.l2rb.ci\endcsname{[$+0.010$, $+0.053$]}
\expandafter\def\csname v@ep.luna.l2rb.lower\endcsname{$+0.031$}
\expandafter\def\csname v@ep.luna.l2rb.lowerci\endcsname{[$+0.010$, $+0.053$]}
\expandafter\def\csname v@ep.luna.l2rb.mean\endcsname{$+0.031$}
\expandafter\def\csname v@ep.luna.l2rb.n\endcsname{256}
\expandafter\def\csname v@ep.luna.l2rb.pbraw\endcsname{$0.011$}
\expandafter\def\csname v@ep.luna.l2rb.praw\endcsname{$0.0045$}
\expandafter\def\csname v@ep.luna.l2rb.upper\endcsname{$+0.031$}
\expandafter\def\csname v@ep.luna.l2rb.upperci\endcsname{[$+0.010$, $+0.053$]}
\expandafter\def\csname v@ep.luna.l3rb.ci\endcsname{[$-0.014$, $+0.014$]}
\expandafter\def\csname v@ep.luna.l3rb.lower\endcsname{$0.000$}
\expandafter\def\csname v@ep.luna.l3rb.lowerci\endcsname{[$-0.014$, $+0.014$]}
\expandafter\def\csname v@ep.luna.l3rb.mean\endcsname{$0.000$}
\expandafter\def\csname v@ep.luna.l3rb.n\endcsname{256}
\expandafter\def\csname v@ep.luna.l3rb.pbraw\endcsname{$1.000$}
\expandafter\def\csname v@ep.luna.l3rb.praw\endcsname{$1.000$}
\expandafter\def\csname v@ep.luna.l3rb.upper\endcsname{$0.000$}
\expandafter\def\csname v@ep.luna.l3rb.upperci\endcsname{[$-0.014$, $+0.014$]}
\expandafter\def\csname v@ep.luna.n.bad.failure.n\endcsname{256}
\expandafter\def\csname v@ep.luna.n.bad.failure.neutral\endcsname{256}
\expandafter\def\csname v@ep.luna.n.bad.failure.nx\endcsname{256}
\expandafter\def\csname v@ep.luna.n.bad.failure.planted\endcsname{256}
\expandafter\def\csname v@ep.luna.n.bad.failure.rb\endcsname{256}
\expandafter\def\csname v@ep.luna.n.bad.failure.rbd\endcsname{256}
\expandafter\def\csname v@ep.luna.n.bad.failure.rbdx\endcsname{256}
\expandafter\def\csname v@ep.luna.n.bad.failure.rbx\endcsname{256}
\expandafter\def\csname v@ep.luna.n.bad.success.n\endcsname{256}
\expandafter\def\csname v@ep.luna.n.bad.success.neutral\endcsname{256}
\expandafter\def\csname v@ep.luna.n.bad.success.nx\endcsname{256}
\expandafter\def\csname v@ep.luna.n.bad.success.planted\endcsname{256}
\expandafter\def\csname v@ep.luna.n.bad.success.rb\endcsname{256}
\expandafter\def\csname v@ep.luna.n.bad.success.rbd\endcsname{256}
\expandafter\def\csname v@ep.luna.n.bad.success.rbdx\endcsname{256}
\expandafter\def\csname v@ep.luna.n.bad.success.rbx\endcsname{256}
\expandafter\def\csname v@ep.luna.n.good.failure.n\endcsname{256}
\expandafter\def\csname v@ep.luna.n.good.failure.neutral\endcsname{256}
\expandafter\def\csname v@ep.luna.n.good.failure.nx\endcsname{256}
\expandafter\def\csname v@ep.luna.n.good.failure.planted\endcsname{256}
\expandafter\def\csname v@ep.luna.n.good.failure.rb\endcsname{256}
\expandafter\def\csname v@ep.luna.n.good.failure.rbd\endcsname{256}
\expandafter\def\csname v@ep.luna.n.good.failure.rbdx\endcsname{256}
\expandafter\def\csname v@ep.luna.n.good.failure.rbx\endcsname{256}
\expandafter\def\csname v@ep.luna.n.good.success.n\endcsname{256}
\expandafter\def\csname v@ep.luna.n.good.success.neutral\endcsname{256}
\expandafter\def\csname v@ep.luna.n.good.success.nx\endcsname{256}
\expandafter\def\csname v@ep.luna.n.good.success.planted\endcsname{256}
\expandafter\def\csname v@ep.luna.n.good.success.rb\endcsname{256}
\expandafter\def\csname v@ep.luna.n.good.success.rbd\endcsname{256}
\expandafter\def\csname v@ep.luna.n.good.success.rbdx\endcsname{256}
\expandafter\def\csname v@ep.luna.n.good.success.rbx\endcsname{256}
\expandafter\def\csname v@ep.luna.name\endcsname{GPT-6 Luna}
\expandafter\def\csname v@ep.luna.namesworse\endcsname{0\%}
\expandafter\def\csname v@ep.luna.p2rb.ci\endcsname{[$+0.010$, $+0.053$]}
\expandafter\def\csname v@ep.luna.p2rb.lower\endcsname{$+0.031$}
\expandafter\def\csname v@ep.luna.p2rb.lowerci\endcsname{[$+0.010$, $+0.053$]}
\expandafter\def\csname v@ep.luna.p2rb.mean\endcsname{$+0.031$}
\expandafter\def\csname v@ep.luna.p2rb.n\endcsname{256}
\expandafter\def\csname v@ep.luna.p2rb.pb\endcsname{$0.157$}
\expandafter\def\csname v@ep.luna.p2rb.pbraw\endcsname{$0.078$}
\expandafter\def\csname v@ep.luna.p2rb.praw\endcsname{$0.0045$}
\expandafter\def\csname v@ep.luna.p2rb.pt\endcsname{$0.0090$}
\expandafter\def\csname v@ep.luna.p2rb.upper\endcsname{$+0.031$}
\expandafter\def\csname v@ep.luna.p2rb.upperci\endcsname{[$+0.010$, $+0.053$]}
\expandafter\def\csname v@ep.luna.p3mrb.ci\endcsname{[$-0.029$, $+0.029$]}
\expandafter\def\csname v@ep.luna.p3mrb.lower\endcsname{$0.000$}
\expandafter\def\csname v@ep.luna.p3mrb.lowerci\endcsname{[$-0.029$, $+0.029$]}
\expandafter\def\csname v@ep.luna.p3mrb.mean\endcsname{$0.000$}
\expandafter\def\csname v@ep.luna.p3mrb.n\endcsname{256}
\expandafter\def\csname v@ep.luna.p3mrb.pb\endcsname{$1.000$}
\expandafter\def\csname v@ep.luna.p3mrb.pbraw\endcsname{$1.000$}
\expandafter\def\csname v@ep.luna.p3mrb.praw\endcsname{$1.000$}
\expandafter\def\csname v@ep.luna.p3mrb.pt\endcsname{$1.000$}
\expandafter\def\csname v@ep.luna.p3mrb.upper\endcsname{$0.000$}
\expandafter\def\csname v@ep.luna.p3mrb.upperci\endcsname{[$-0.029$, $+0.029$]}
\expandafter\def\csname v@ep.luna.planted.k\endcsname{4}
\expandafter\def\csname v@ep.luna.planted.n\endcsname{1,024}
\expandafter\def\csname v@ep.luna.rate.bad.failure.n\endcsname{0.0\%}
\expandafter\def\csname v@ep.luna.rate.bad.failure.neutral\endcsname{0.0\%}
\expandafter\def\csname v@ep.luna.rate.bad.failure.nx\endcsname{0.0\%}
\expandafter\def\csname v@ep.luna.rate.bad.failure.planted\endcsname{0.0\%}
\expandafter\def\csname v@ep.luna.rate.bad.failure.rb\endcsname{0.0\%}
\expandafter\def\csname v@ep.luna.rate.bad.failure.rbd\endcsname{0.0\%}
\expandafter\def\csname v@ep.luna.rate.bad.failure.rbdx\endcsname{0.0\%}
\expandafter\def\csname v@ep.luna.rate.bad.failure.rbx\endcsname{0.0\%}
\expandafter\def\csname v@ep.luna.rate.bad.success.n\endcsname{0.0\%}
\expandafter\def\csname v@ep.luna.rate.bad.success.neutral\endcsname{0.0\%}
\expandafter\def\csname v@ep.luna.rate.bad.success.nx\endcsname{0.0\%}
\expandafter\def\csname v@ep.luna.rate.bad.success.planted\endcsname{0.0\%}
\expandafter\def\csname v@ep.luna.rate.bad.success.rb\endcsname{0.0\%}
\expandafter\def\csname v@ep.luna.rate.bad.success.rbd\endcsname{0.0\%}
\expandafter\def\csname v@ep.luna.rate.bad.success.rbdx\endcsname{0.0\%}
\expandafter\def\csname v@ep.luna.rate.bad.success.rbx\endcsname{0.0\%}
\expandafter\def\csname v@ep.luna.rate.good.failure.n\endcsname{0.0\%}
\expandafter\def\csname v@ep.luna.rate.good.failure.neutral\endcsname{0.0\%}
\expandafter\def\csname v@ep.luna.rate.good.failure.nx\endcsname{0.0\%}
\expandafter\def\csname v@ep.luna.rate.good.failure.planted\endcsname{1.6\%}
\expandafter\def\csname v@ep.luna.rate.good.failure.rb\endcsname{3.1\%}
\expandafter\def\csname v@ep.luna.rate.good.failure.rbd\endcsname{0.4\%}
\expandafter\def\csname v@ep.luna.rate.good.failure.rbdx\endcsname{0.0\%}
\expandafter\def\csname v@ep.luna.rate.good.failure.rbx\endcsname{0.0\%}
\expandafter\def\csname v@ep.luna.rate.good.success.n\endcsname{0.0\%}
\expandafter\def\csname v@ep.luna.rate.good.success.neutral\endcsname{0.0\%}
\expandafter\def\csname v@ep.luna.rate.good.success.nx\endcsname{0.0\%}
\expandafter\def\csname v@ep.luna.rate.good.success.planted\endcsname{0.0\%}
\expandafter\def\csname v@ep.luna.rate.good.success.rb\endcsname{0.0\%}
\expandafter\def\csname v@ep.luna.rate.good.success.rbd\endcsname{0.0\%}
\expandafter\def\csname v@ep.luna.rate.good.success.rbdx\endcsname{0.0\%}
\expandafter\def\csname v@ep.luna.rate.good.success.rbx\endcsname{0.0\%}
\expandafter\def\csname v@ep.luna.sus.ci\endcsname{[$0.000$, $+0.008$]}
\expandafter\def\csname v@ep.luna.sus.mean\endcsname{$+0.004$}
\expandafter\def\csname v@ep.luna.sus.n\endcsname{256}
\expandafter\def\csname v@ep.luna.sus.pbraw\endcsname{$0.503$}
\expandafter\def\csname v@ep.luna.sus.praw\endcsname{$0.045$}
\expandafter\def\csname v@ep.luna.word.pthree\endcsname{\emph{bounded below 0.10}}
\expandafter\def\csname v@ep.luna.word.ptwo\endcsname{\emph{bounded below 0.10}}
\expandafter\def\csname v@ep.luna.word.sus\endcsname{\emph{increment below 0.10}}
\expandafter\def\csname v@ep.mistral.a2rb.ci\endcsname{[$+0.123$, $+0.216$]}
\expandafter\def\csname v@ep.mistral.a2rb.lower\endcsname{$+0.168$}
\expandafter\def\csname v@ep.mistral.a2rb.lowerci\endcsname{[$+0.122$, $+0.214$]}
\expandafter\def\csname v@ep.mistral.a2rb.mean\endcsname{$+0.169$}
\expandafter\def\csname v@ep.mistral.a2rb.n\endcsname{254}
\expandafter\def\csname v@ep.mistral.a2rb.pbraw\endcsname{$3.5\times10^{-11}$}
\expandafter\def\csname v@ep.mistral.a2rb.praw\endcsname{$7.7\times10^{-12}$}
\expandafter\def\csname v@ep.mistral.a2rb.upper\endcsname{$+0.176$}
\expandafter\def\csname v@ep.mistral.a2rb.upperci\endcsname{[$+0.129$, $+0.223$]}
\expandafter\def\csname v@ep.mistral.a3rb.ci\endcsname{[$-0.014$, $+0.014$]}
\expandafter\def\csname v@ep.mistral.a3rb.lower\endcsname{$0.000$}
\expandafter\def\csname v@ep.mistral.a3rb.lowerci\endcsname{[$-0.014$, $+0.014$]}
\expandafter\def\csname v@ep.mistral.a3rb.mean\endcsname{$0.000$}
\expandafter\def\csname v@ep.mistral.a3rb.n\endcsname{255}
\expandafter\def\csname v@ep.mistral.a3rb.pbraw\endcsname{$1.000$}
\expandafter\def\csname v@ep.mistral.a3rb.praw\endcsname{$1.000$}
\expandafter\def\csname v@ep.mistral.a3rb.upper\endcsname{$+0.004$}
\expandafter\def\csname v@ep.mistral.a3rb.upperci\endcsname{[$-0.004$, $+0.012$]}
\expandafter\def\csname v@ep.mistral.count.bad.failure.n\endcsname{0/256}
\expandafter\def\csname v@ep.mistral.count.bad.failure.neutral\endcsname{0/255}
\expandafter\def\csname v@ep.mistral.count.bad.failure.nx\endcsname{0/256}
\expandafter\def\csname v@ep.mistral.count.bad.failure.planted\endcsname{8/256}
\expandafter\def\csname v@ep.mistral.count.bad.failure.rb\endcsname{0/256}
\expandafter\def\csname v@ep.mistral.count.bad.failure.rbd\endcsname{0/256}
\expandafter\def\csname v@ep.mistral.count.bad.failure.rbdx\endcsname{0/255}
\expandafter\def\csname v@ep.mistral.count.bad.failure.rbx\endcsname{0/255}
\expandafter\def\csname v@ep.mistral.count.bad.success.n\endcsname{1/256}
\expandafter\def\csname v@ep.mistral.count.bad.success.neutral\endcsname{0/256}
\expandafter\def\csname v@ep.mistral.count.bad.success.nx\endcsname{0/255}
\expandafter\def\csname v@ep.mistral.count.bad.success.planted\endcsname{27/255}
\expandafter\def\csname v@ep.mistral.count.bad.success.rb\endcsname{0/255}
\expandafter\def\csname v@ep.mistral.count.bad.success.rbd\endcsname{0/256}
\expandafter\def\csname v@ep.mistral.count.bad.success.rbdx\endcsname{0/256}
\expandafter\def\csname v@ep.mistral.count.bad.success.rbx\endcsname{0/256}
\expandafter\def\csname v@ep.mistral.count.good.failure.n\endcsname{39/256}
\expandafter\def\csname v@ep.mistral.count.good.failure.neutral\endcsname{0/256}
\expandafter\def\csname v@ep.mistral.count.good.failure.nx\endcsname{0/256}
\expandafter\def\csname v@ep.mistral.count.good.failure.planted\endcsname{124/256}
\expandafter\def\csname v@ep.mistral.count.good.failure.rb\endcsname{43/254}
\expandafter\def\csname v@ep.mistral.count.good.failure.rbd\endcsname{47/255}
\expandafter\def\csname v@ep.mistral.count.good.failure.rbdx\endcsname{0/256}
\expandafter\def\csname v@ep.mistral.count.good.failure.rbx\endcsname{0/256}
\expandafter\def\csname v@ep.mistral.count.good.success.n\endcsname{0/256}
\expandafter\def\csname v@ep.mistral.count.good.success.neutral\endcsname{0/255}
\expandafter\def\csname v@ep.mistral.count.good.success.nx\endcsname{1/256}
\expandafter\def\csname v@ep.mistral.count.good.success.planted\endcsname{80/256}
\expandafter\def\csname v@ep.mistral.count.good.success.rb\endcsname{0/256}
\expandafter\def\csname v@ep.mistral.count.good.success.rbd\endcsname{0/255}
\expandafter\def\csname v@ep.mistral.count.good.success.rbdx\endcsname{0/256}
\expandafter\def\csname v@ep.mistral.count.good.success.rbx\endcsname{0/256}
\expandafter\def\csname v@ep.mistral.id\endcsname{\texttt{mistralai/mistral-large-4-0}}
\expandafter\def\csname v@ep.mistral.k.bad.failure.n\endcsname{0}
\expandafter\def\csname v@ep.mistral.k.bad.failure.neutral\endcsname{0}
\expandafter\def\csname v@ep.mistral.k.bad.failure.nx\endcsname{0}
\expandafter\def\csname v@ep.mistral.k.bad.failure.planted\endcsname{8}
\expandafter\def\csname v@ep.mistral.k.bad.failure.rb\endcsname{0}
\expandafter\def\csname v@ep.mistral.k.bad.failure.rbd\endcsname{0}
\expandafter\def\csname v@ep.mistral.k.bad.failure.rbdx\endcsname{0}
\expandafter\def\csname v@ep.mistral.k.bad.failure.rbx\endcsname{0}
\expandafter\def\csname v@ep.mistral.k.bad.success.n\endcsname{1}
\expandafter\def\csname v@ep.mistral.k.bad.success.neutral\endcsname{0}
\expandafter\def\csname v@ep.mistral.k.bad.success.nx\endcsname{0}
\expandafter\def\csname v@ep.mistral.k.bad.success.planted\endcsname{27}
\expandafter\def\csname v@ep.mistral.k.bad.success.rb\endcsname{0}
\expandafter\def\csname v@ep.mistral.k.bad.success.rbd\endcsname{0}
\expandafter\def\csname v@ep.mistral.k.bad.success.rbdx\endcsname{0}
\expandafter\def\csname v@ep.mistral.k.bad.success.rbx\endcsname{0}
\expandafter\def\csname v@ep.mistral.k.good.failure.n\endcsname{39}
\expandafter\def\csname v@ep.mistral.k.good.failure.neutral\endcsname{0}
\expandafter\def\csname v@ep.mistral.k.good.failure.nx\endcsname{0}
\expandafter\def\csname v@ep.mistral.k.good.failure.planted\endcsname{124}
\expandafter\def\csname v@ep.mistral.k.good.failure.rb\endcsname{43}
\expandafter\def\csname v@ep.mistral.k.good.failure.rbd\endcsname{47}
\expandafter\def\csname v@ep.mistral.k.good.failure.rbdx\endcsname{0}
\expandafter\def\csname v@ep.mistral.k.good.failure.rbx\endcsname{0}
\expandafter\def\csname v@ep.mistral.k.good.success.n\endcsname{0}
\expandafter\def\csname v@ep.mistral.k.good.success.neutral\endcsname{0}
\expandafter\def\csname v@ep.mistral.k.good.success.nx\endcsname{1}
\expandafter\def\csname v@ep.mistral.k.good.success.planted\endcsname{80}
\expandafter\def\csname v@ep.mistral.k.good.success.rb\endcsname{0}
\expandafter\def\csname v@ep.mistral.k.good.success.rbd\endcsname{0}
\expandafter\def\csname v@ep.mistral.k.good.success.rbdx\endcsname{0}
\expandafter\def\csname v@ep.mistral.k.good.success.rbx\endcsname{0}
\expandafter\def\csname v@ep.mistral.l2brb.ci\endcsname{[$-0.014$, $+0.014$]}
\expandafter\def\csname v@ep.mistral.l2brb.mean\endcsname{$0.000$}
\expandafter\def\csname v@ep.mistral.l2brb.n\endcsname{256}
\expandafter\def\csname v@ep.mistral.l2brb.pbraw\endcsname{$1.000$}
\expandafter\def\csname v@ep.mistral.l2brb.praw\endcsname{$1.000$}
\expandafter\def\csname v@ep.mistral.l2rb.ci\endcsname{[$+0.123$, $+0.216$]}
\expandafter\def\csname v@ep.mistral.l2rb.lower\endcsname{$+0.168$}
\expandafter\def\csname v@ep.mistral.l2rb.lowerci\endcsname{[$+0.122$, $+0.214$]}
\expandafter\def\csname v@ep.mistral.l2rb.mean\endcsname{$+0.169$}
\expandafter\def\csname v@ep.mistral.l2rb.n\endcsname{254}
\expandafter\def\csname v@ep.mistral.l2rb.pbraw\endcsname{$3.5\times10^{-11}$}
\expandafter\def\csname v@ep.mistral.l2rb.praw\endcsname{$7.7\times10^{-12}$}
\expandafter\def\csname v@ep.mistral.l2rb.upper\endcsname{$+0.176$}
\expandafter\def\csname v@ep.mistral.l2rb.upperci\endcsname{[$+0.129$, $+0.223$]}
\expandafter\def\csname v@ep.mistral.l3rb.ci\endcsname{[$-0.014$, $+0.014$]}
\expandafter\def\csname v@ep.mistral.l3rb.lower\endcsname{$0.000$}
\expandafter\def\csname v@ep.mistral.l3rb.lowerci\endcsname{[$-0.014$, $+0.014$]}
\expandafter\def\csname v@ep.mistral.l3rb.mean\endcsname{$0.000$}
\expandafter\def\csname v@ep.mistral.l3rb.n\endcsname{255}
\expandafter\def\csname v@ep.mistral.l3rb.pbraw\endcsname{$1.000$}
\expandafter\def\csname v@ep.mistral.l3rb.praw\endcsname{$1.000$}
\expandafter\def\csname v@ep.mistral.l3rb.upper\endcsname{$+0.004$}
\expandafter\def\csname v@ep.mistral.l3rb.upperci\endcsname{[$-0.004$, $+0.012$]}
\expandafter\def\csname v@ep.mistral.n.bad.failure.n\endcsname{256}
\expandafter\def\csname v@ep.mistral.n.bad.failure.neutral\endcsname{255}
\expandafter\def\csname v@ep.mistral.n.bad.failure.nx\endcsname{256}
\expandafter\def\csname v@ep.mistral.n.bad.failure.planted\endcsname{256}
\expandafter\def\csname v@ep.mistral.n.bad.failure.rb\endcsname{256}
\expandafter\def\csname v@ep.mistral.n.bad.failure.rbd\endcsname{256}
\expandafter\def\csname v@ep.mistral.n.bad.failure.rbdx\endcsname{255}
\expandafter\def\csname v@ep.mistral.n.bad.failure.rbx\endcsname{255}
\expandafter\def\csname v@ep.mistral.n.bad.success.n\endcsname{256}
\expandafter\def\csname v@ep.mistral.n.bad.success.neutral\endcsname{256}
\expandafter\def\csname v@ep.mistral.n.bad.success.nx\endcsname{255}
\expandafter\def\csname v@ep.mistral.n.bad.success.planted\endcsname{255}
\expandafter\def\csname v@ep.mistral.n.bad.success.rb\endcsname{255}
\expandafter\def\csname v@ep.mistral.n.bad.success.rbd\endcsname{256}
\expandafter\def\csname v@ep.mistral.n.bad.success.rbdx\endcsname{256}
\expandafter\def\csname v@ep.mistral.n.bad.success.rbx\endcsname{256}
\expandafter\def\csname v@ep.mistral.n.good.failure.n\endcsname{256}
\expandafter\def\csname v@ep.mistral.n.good.failure.neutral\endcsname{256}
\expandafter\def\csname v@ep.mistral.n.good.failure.nx\endcsname{256}
\expandafter\def\csname v@ep.mistral.n.good.failure.planted\endcsname{256}
\expandafter\def\csname v@ep.mistral.n.good.failure.rb\endcsname{254}
\expandafter\def\csname v@ep.mistral.n.good.failure.rbd\endcsname{255}
\expandafter\def\csname v@ep.mistral.n.good.failure.rbdx\endcsname{256}
\expandafter\def\csname v@ep.mistral.n.good.failure.rbx\endcsname{256}
\expandafter\def\csname v@ep.mistral.n.good.success.n\endcsname{256}
\expandafter\def\csname v@ep.mistral.n.good.success.neutral\endcsname{255}
\expandafter\def\csname v@ep.mistral.n.good.success.nx\endcsname{256}
\expandafter\def\csname v@ep.mistral.n.good.success.planted\endcsname{256}
\expandafter\def\csname v@ep.mistral.n.good.success.rb\endcsname{256}
\expandafter\def\csname v@ep.mistral.n.good.success.rbd\endcsname{255}
\expandafter\def\csname v@ep.mistral.n.good.success.rbdx\endcsname{256}
\expandafter\def\csname v@ep.mistral.n.good.success.rbx\endcsname{256}
\expandafter\def\csname v@ep.mistral.name\endcsname{Mistral Large 4}
\expandafter\def\csname v@ep.mistral.namesworse\endcsname{38\%}
\expandafter\def\csname v@ep.mistral.p2rb.ci\endcsname{[$+0.120$, $+0.212$]}
\expandafter\def\csname v@ep.mistral.p2rb.lower\endcsname{$+0.168$}
\expandafter\def\csname v@ep.mistral.p2rb.lowerci\endcsname{[$+0.122$, $+0.214$]}
\expandafter\def\csname v@ep.mistral.p2rb.mean\endcsname{$+0.166$}
\expandafter\def\csname v@ep.mistral.p2rb.n\endcsname{253}
\expandafter\def\csname v@ep.mistral.p2rb.pb\endcsname{$3.1\times10^{-6}$}
\expandafter\def\csname v@ep.mistral.p2rb.pbraw\endcsname{$1.6\times10^{-6}$}
\expandafter\def\csname v@ep.mistral.p2rb.praw\endcsname{$1.4\times10^{-11}$}
\expandafter\def\csname v@ep.mistral.p2rb.pt\endcsname{$2.8\times10^{-11}$}
\expandafter\def\csname v@ep.mistral.p2rb.upper\endcsname{$+0.180$}
\expandafter\def\csname v@ep.mistral.p2rb.upperci\endcsname{[$+0.131$, $+0.228$]}
\expandafter\def\csname v@ep.mistral.p3mrb.ci\endcsname{[$-0.029$, $+0.029$]}
\expandafter\def\csname v@ep.mistral.p3mrb.lower\endcsname{$0.000$}
\expandafter\def\csname v@ep.mistral.p3mrb.lowerci\endcsname{[$-0.029$, $+0.029$]}
\expandafter\def\csname v@ep.mistral.p3mrb.mean\endcsname{$0.000$}
\expandafter\def\csname v@ep.mistral.p3mrb.n\endcsname{255}
\expandafter\def\csname v@ep.mistral.p3mrb.pb\endcsname{$1.000$}
\expandafter\def\csname v@ep.mistral.p3mrb.pbraw\endcsname{$1.000$}
\expandafter\def\csname v@ep.mistral.p3mrb.praw\endcsname{$1.000$}
\expandafter\def\csname v@ep.mistral.p3mrb.pt\endcsname{$1.000$}
\expandafter\def\csname v@ep.mistral.p3mrb.upper\endcsname{$+0.008$}
\expandafter\def\csname v@ep.mistral.p3mrb.upperci\endcsname{[$-0.008$, $+0.023$]}
\expandafter\def\csname v@ep.mistral.planted.k\endcsname{239}
\expandafter\def\csname v@ep.mistral.planted.n\endcsname{1,023}
\expandafter\def\csname v@ep.mistral.rate.bad.failure.n\endcsname{0.0\%}
\expandafter\def\csname v@ep.mistral.rate.bad.failure.neutral\endcsname{0.0\%}
\expandafter\def\csname v@ep.mistral.rate.bad.failure.nx\endcsname{0.0\%}
\expandafter\def\csname v@ep.mistral.rate.bad.failure.planted\endcsname{3.1\%}
\expandafter\def\csname v@ep.mistral.rate.bad.failure.rb\endcsname{0.0\%}
\expandafter\def\csname v@ep.mistral.rate.bad.failure.rbd\endcsname{0.0\%}
\expandafter\def\csname v@ep.mistral.rate.bad.failure.rbdx\endcsname{0.0\%}
\expandafter\def\csname v@ep.mistral.rate.bad.failure.rbx\endcsname{0.0\%}
\expandafter\def\csname v@ep.mistral.rate.bad.success.n\endcsname{0.4\%}
\expandafter\def\csname v@ep.mistral.rate.bad.success.neutral\endcsname{0.0\%}
\expandafter\def\csname v@ep.mistral.rate.bad.success.nx\endcsname{0.0\%}
\expandafter\def\csname v@ep.mistral.rate.bad.success.planted\endcsname{10.6\%}
\expandafter\def\csname v@ep.mistral.rate.bad.success.rb\endcsname{0.0\%}
\expandafter\def\csname v@ep.mistral.rate.bad.success.rbd\endcsname{0.0\%}
\expandafter\def\csname v@ep.mistral.rate.bad.success.rbdx\endcsname{0.0\%}
\expandafter\def\csname v@ep.mistral.rate.bad.success.rbx\endcsname{0.0\%}
\expandafter\def\csname v@ep.mistral.rate.good.failure.n\endcsname{15.2\%}
\expandafter\def\csname v@ep.mistral.rate.good.failure.neutral\endcsname{0.0\%}
\expandafter\def\csname v@ep.mistral.rate.good.failure.nx\endcsname{0.0\%}
\expandafter\def\csname v@ep.mistral.rate.good.failure.planted\endcsname{48.4\%}
\expandafter\def\csname v@ep.mistral.rate.good.failure.rb\endcsname{16.9\%}
\expandafter\def\csname v@ep.mistral.rate.good.failure.rbd\endcsname{18.4\%}
\expandafter\def\csname v@ep.mistral.rate.good.failure.rbdx\endcsname{0.0\%}
\expandafter\def\csname v@ep.mistral.rate.good.failure.rbx\endcsname{0.0\%}
\expandafter\def\csname v@ep.mistral.rate.good.success.n\endcsname{0.0\%}
\expandafter\def\csname v@ep.mistral.rate.good.success.neutral\endcsname{0.0\%}
\expandafter\def\csname v@ep.mistral.rate.good.success.nx\endcsname{0.4\%}
\expandafter\def\csname v@ep.mistral.rate.good.success.planted\endcsname{31.3\%}
\expandafter\def\csname v@ep.mistral.rate.good.success.rb\endcsname{0.0\%}
\expandafter\def\csname v@ep.mistral.rate.good.success.rbd\endcsname{0.0\%}
\expandafter\def\csname v@ep.mistral.rate.good.success.rbdx\endcsname{0.0\%}
\expandafter\def\csname v@ep.mistral.rate.good.success.rbx\endcsname{0.0\%}
\expandafter\def\csname v@ep.mistral.sus.ci\endcsname{[$+0.206$, $+0.258$]}
\expandafter\def\csname v@ep.mistral.sus.mean\endcsname{$+0.232$}
\expandafter\def\csname v@ep.mistral.sus.n\endcsname{253}
\expandafter\def\csname v@ep.mistral.sus.pbraw\endcsname{$2.1\times10^{-17}$}
\expandafter\def\csname v@ep.mistral.sus.praw\endcsname{$2.9\times10^{-46}$}
\expandafter\def\csname v@ep.mistral.word.pthree\endcsname{\emph{bounded below 0.10}}
\expandafter\def\csname v@ep.mistral.word.ptwo\endcsname{\emph{luck-carrying}}
\expandafter\def\csname v@ep.mistral.word.sus\endcsname{\emph{increment above 0.10}}
\expandafter\def\csname v@ep.n.pthree.max\endcsname{256}
\expandafter\def\csname v@ep.n.pthree.min\endcsname{255}
\expandafter\def\csname v@ep.n.ptwo.max\endcsname{256}
\expandafter\def\csname v@ep.n.ptwo.min\endcsname{253}
\expandafter\def\csname v@ep.na2l2robust\endcsname{8}
\expandafter\def\csname v@ep.na2l2robust.notluck\endcsname{2}
\expandafter\def\csname v@ep.nallbetter\endcsname{3}
\expandafter\def\csname v@ep.nbounded\endcsname{5}
\expandafter\def\csname v@ep.nbounded.pthree\endcsname{11}
\expandafter\def\csname v@ep.nboundedpos\endcsname{2}
\expandafter\def\csname v@ep.nluck\endcsname{6}
\expandafter\def\csname v@ep.nsusabove\endcsname{4}
\expandafter\def\csname v@ep.nsusbelow\endcsname{7}
\expandafter\def\csname v@ep.opus.a2rb.ci\endcsname{[$-0.014$, $+0.014$]}
\expandafter\def\csname v@ep.opus.a2rb.lower\endcsname{$0.000$}
\expandafter\def\csname v@ep.opus.a2rb.lowerci\endcsname{[$-0.014$, $+0.014$]}
\expandafter\def\csname v@ep.opus.a2rb.mean\endcsname{$0.000$}
\expandafter\def\csname v@ep.opus.a2rb.n\endcsname{255}
\expandafter\def\csname v@ep.opus.a2rb.pbraw\endcsname{$1.000$}
\expandafter\def\csname v@ep.opus.a2rb.praw\endcsname{$1.000$}
\expandafter\def\csname v@ep.opus.a2rb.upper\endcsname{$+0.004$}
\expandafter\def\csname v@ep.opus.a2rb.upperci\endcsname{[$-0.004$, $+0.012$]}
\expandafter\def\csname v@ep.opus.a3rb.ci\endcsname{[$-0.014$, $+0.014$]}
\expandafter\def\csname v@ep.opus.a3rb.lower\endcsname{$0.000$}
\expandafter\def\csname v@ep.opus.a3rb.lowerci\endcsname{[$-0.014$, $+0.014$]}
\expandafter\def\csname v@ep.opus.a3rb.mean\endcsname{$0.000$}
\expandafter\def\csname v@ep.opus.a3rb.n\endcsname{256}
\expandafter\def\csname v@ep.opus.a3rb.pbraw\endcsname{$1.000$}
\expandafter\def\csname v@ep.opus.a3rb.praw\endcsname{$1.000$}
\expandafter\def\csname v@ep.opus.a3rb.upper\endcsname{$0.000$}
\expandafter\def\csname v@ep.opus.a3rb.upperci\endcsname{[$-0.014$, $+0.014$]}
\expandafter\def\csname v@ep.opus.count.bad.failure.n\endcsname{0/256}
\expandafter\def\csname v@ep.opus.count.bad.failure.neutral\endcsname{0/256}
\expandafter\def\csname v@ep.opus.count.bad.failure.nx\endcsname{0/256}
\expandafter\def\csname v@ep.opus.count.bad.failure.planted\endcsname{0/256}
\expandafter\def\csname v@ep.opus.count.bad.failure.rb\endcsname{0/256}
\expandafter\def\csname v@ep.opus.count.bad.failure.rbd\endcsname{0/256}
\expandafter\def\csname v@ep.opus.count.bad.failure.rbdx\endcsname{0/256}
\expandafter\def\csname v@ep.opus.count.bad.failure.rbx\endcsname{0/256}
\expandafter\def\csname v@ep.opus.count.bad.success.n\endcsname{0/256}
\expandafter\def\csname v@ep.opus.count.bad.success.neutral\endcsname{0/256}
\expandafter\def\csname v@ep.opus.count.bad.success.nx\endcsname{0/256}
\expandafter\def\csname v@ep.opus.count.bad.success.planted\endcsname{0/256}
\expandafter\def\csname v@ep.opus.count.bad.success.rb\endcsname{0/256}
\expandafter\def\csname v@ep.opus.count.bad.success.rbd\endcsname{0/256}
\expandafter\def\csname v@ep.opus.count.bad.success.rbdx\endcsname{0/256}
\expandafter\def\csname v@ep.opus.count.bad.success.rbx\endcsname{0/256}
\expandafter\def\csname v@ep.opus.count.good.failure.n\endcsname{0/256}
\expandafter\def\csname v@ep.opus.count.good.failure.neutral\endcsname{0/256}
\expandafter\def\csname v@ep.opus.count.good.failure.nx\endcsname{0/256}
\expandafter\def\csname v@ep.opus.count.good.failure.planted\endcsname{0/256}
\expandafter\def\csname v@ep.opus.count.good.failure.rb\endcsname{0/255}
\expandafter\def\csname v@ep.opus.count.good.failure.rbd\endcsname{0/256}
\expandafter\def\csname v@ep.opus.count.good.failure.rbdx\endcsname{0/256}
\expandafter\def\csname v@ep.opus.count.good.failure.rbx\endcsname{0/256}
\expandafter\def\csname v@ep.opus.count.good.success.n\endcsname{0/256}
\expandafter\def\csname v@ep.opus.count.good.success.neutral\endcsname{0/256}
\expandafter\def\csname v@ep.opus.count.good.success.nx\endcsname{0/256}
\expandafter\def\csname v@ep.opus.count.good.success.planted\endcsname{0/256}
\expandafter\def\csname v@ep.opus.count.good.success.rb\endcsname{0/256}
\expandafter\def\csname v@ep.opus.count.good.success.rbd\endcsname{0/256}
\expandafter\def\csname v@ep.opus.count.good.success.rbdx\endcsname{0/256}
\expandafter\def\csname v@ep.opus.count.good.success.rbx\endcsname{0/255}
\expandafter\def\csname v@ep.opus.id\endcsname{\texttt{anthropic/claude-opus-5.5}}
\expandafter\def\csname v@ep.opus.k.bad.failure.n\endcsname{0}
\expandafter\def\csname v@ep.opus.k.bad.failure.neutral\endcsname{0}
\expandafter\def\csname v@ep.opus.k.bad.failure.nx\endcsname{0}
\expandafter\def\csname v@ep.opus.k.bad.failure.planted\endcsname{0}
\expandafter\def\csname v@ep.opus.k.bad.failure.rb\endcsname{0}
\expandafter\def\csname v@ep.opus.k.bad.failure.rbd\endcsname{0}
\expandafter\def\csname v@ep.opus.k.bad.failure.rbdx\endcsname{0}
\expandafter\def\csname v@ep.opus.k.bad.failure.rbx\endcsname{0}
\expandafter\def\csname v@ep.opus.k.bad.success.n\endcsname{0}
\expandafter\def\csname v@ep.opus.k.bad.success.neutral\endcsname{0}
\expandafter\def\csname v@ep.opus.k.bad.success.nx\endcsname{0}
\expandafter\def\csname v@ep.opus.k.bad.success.planted\endcsname{0}
\expandafter\def\csname v@ep.opus.k.bad.success.rb\endcsname{0}
\expandafter\def\csname v@ep.opus.k.bad.success.rbd\endcsname{0}
\expandafter\def\csname v@ep.opus.k.bad.success.rbdx\endcsname{0}
\expandafter\def\csname v@ep.opus.k.bad.success.rbx\endcsname{0}
\expandafter\def\csname v@ep.opus.k.good.failure.n\endcsname{0}
\expandafter\def\csname v@ep.opus.k.good.failure.neutral\endcsname{0}
\expandafter\def\csname v@ep.opus.k.good.failure.nx\endcsname{0}
\expandafter\def\csname v@ep.opus.k.good.failure.planted\endcsname{0}
\expandafter\def\csname v@ep.opus.k.good.failure.rb\endcsname{0}
\expandafter\def\csname v@ep.opus.k.good.failure.rbd\endcsname{0}
\expandafter\def\csname v@ep.opus.k.good.failure.rbdx\endcsname{0}
\expandafter\def\csname v@ep.opus.k.good.failure.rbx\endcsname{0}
\expandafter\def\csname v@ep.opus.k.good.success.n\endcsname{0}
\expandafter\def\csname v@ep.opus.k.good.success.neutral\endcsname{0}
\expandafter\def\csname v@ep.opus.k.good.success.nx\endcsname{0}
\expandafter\def\csname v@ep.opus.k.good.success.planted\endcsname{0}
\expandafter\def\csname v@ep.opus.k.good.success.rb\endcsname{0}
\expandafter\def\csname v@ep.opus.k.good.success.rbd\endcsname{0}
\expandafter\def\csname v@ep.opus.k.good.success.rbdx\endcsname{0}
\expandafter\def\csname v@ep.opus.k.good.success.rbx\endcsname{0}
\expandafter\def\csname v@ep.opus.l2brb.ci\endcsname{[$-0.014$, $+0.014$]}
\expandafter\def\csname v@ep.opus.l2brb.mean\endcsname{$0.000$}
\expandafter\def\csname v@ep.opus.l2brb.n\endcsname{255}
\expandafter\def\csname v@ep.opus.l2brb.pbraw\endcsname{$1.000$}
\expandafter\def\csname v@ep.opus.l2brb.praw\endcsname{$1.000$}
\expandafter\def\csname v@ep.opus.l2rb.ci\endcsname{[$-0.014$, $+0.014$]}
\expandafter\def\csname v@ep.opus.l2rb.lower\endcsname{$0.000$}
\expandafter\def\csname v@ep.opus.l2rb.lowerci\endcsname{[$-0.014$, $+0.014$]}
\expandafter\def\csname v@ep.opus.l2rb.mean\endcsname{$0.000$}
\expandafter\def\csname v@ep.opus.l2rb.n\endcsname{255}
\expandafter\def\csname v@ep.opus.l2rb.pbraw\endcsname{$1.000$}
\expandafter\def\csname v@ep.opus.l2rb.praw\endcsname{$1.000$}
\expandafter\def\csname v@ep.opus.l2rb.upper\endcsname{$+0.004$}
\expandafter\def\csname v@ep.opus.l2rb.upperci\endcsname{[$-0.004$, $+0.012$]}
\expandafter\def\csname v@ep.opus.l3rb.ci\endcsname{[$-0.014$, $+0.014$]}
\expandafter\def\csname v@ep.opus.l3rb.lower\endcsname{$0.000$}
\expandafter\def\csname v@ep.opus.l3rb.lowerci\endcsname{[$-0.014$, $+0.014$]}
\expandafter\def\csname v@ep.opus.l3rb.mean\endcsname{$0.000$}
\expandafter\def\csname v@ep.opus.l3rb.n\endcsname{256}
\expandafter\def\csname v@ep.opus.l3rb.pbraw\endcsname{$1.000$}
\expandafter\def\csname v@ep.opus.l3rb.praw\endcsname{$1.000$}
\expandafter\def\csname v@ep.opus.l3rb.upper\endcsname{$0.000$}
\expandafter\def\csname v@ep.opus.l3rb.upperci\endcsname{[$-0.014$, $+0.014$]}
\expandafter\def\csname v@ep.opus.n.bad.failure.n\endcsname{256}
\expandafter\def\csname v@ep.opus.n.bad.failure.neutral\endcsname{256}
\expandafter\def\csname v@ep.opus.n.bad.failure.nx\endcsname{256}
\expandafter\def\csname v@ep.opus.n.bad.failure.planted\endcsname{256}
\expandafter\def\csname v@ep.opus.n.bad.failure.rb\endcsname{256}
\expandafter\def\csname v@ep.opus.n.bad.failure.rbd\endcsname{256}
\expandafter\def\csname v@ep.opus.n.bad.failure.rbdx\endcsname{256}
\expandafter\def\csname v@ep.opus.n.bad.failure.rbx\endcsname{256}
\expandafter\def\csname v@ep.opus.n.bad.success.n\endcsname{256}
\expandafter\def\csname v@ep.opus.n.bad.success.neutral\endcsname{256}
\expandafter\def\csname v@ep.opus.n.bad.success.nx\endcsname{256}
\expandafter\def\csname v@ep.opus.n.bad.success.planted\endcsname{256}
\expandafter\def\csname v@ep.opus.n.bad.success.rb\endcsname{256}
\expandafter\def\csname v@ep.opus.n.bad.success.rbd\endcsname{256}
\expandafter\def\csname v@ep.opus.n.bad.success.rbdx\endcsname{256}
\expandafter\def\csname v@ep.opus.n.bad.success.rbx\endcsname{256}
\expandafter\def\csname v@ep.opus.n.good.failure.n\endcsname{256}
\expandafter\def\csname v@ep.opus.n.good.failure.neutral\endcsname{256}
\expandafter\def\csname v@ep.opus.n.good.failure.nx\endcsname{256}
\expandafter\def\csname v@ep.opus.n.good.failure.planted\endcsname{256}
\expandafter\def\csname v@ep.opus.n.good.failure.rb\endcsname{255}
\expandafter\def\csname v@ep.opus.n.good.failure.rbd\endcsname{256}
\expandafter\def\csname v@ep.opus.n.good.failure.rbdx\endcsname{256}
\expandafter\def\csname v@ep.opus.n.good.failure.rbx\endcsname{256}
\expandafter\def\csname v@ep.opus.n.good.success.n\endcsname{256}
\expandafter\def\csname v@ep.opus.n.good.success.neutral\endcsname{256}
\expandafter\def\csname v@ep.opus.n.good.success.nx\endcsname{256}
\expandafter\def\csname v@ep.opus.n.good.success.planted\endcsname{256}
\expandafter\def\csname v@ep.opus.n.good.success.rb\endcsname{256}
\expandafter\def\csname v@ep.opus.n.good.success.rbd\endcsname{256}
\expandafter\def\csname v@ep.opus.n.good.success.rbdx\endcsname{256}
\expandafter\def\csname v@ep.opus.n.good.success.rbx\endcsname{255}
\expandafter\def\csname v@ep.opus.name\endcsname{Claude Opus 5.5}
\expandafter\def\csname v@ep.opus.namesworse\endcsname{67\%}
\expandafter\def\csname v@ep.opus.p2rb.ci\endcsname{[$-0.029$, $+0.029$]}
\expandafter\def\csname v@ep.opus.p2rb.lower\endcsname{$0.000$}
\expandafter\def\csname v@ep.opus.p2rb.lowerci\endcsname{[$-0.029$, $+0.029$]}
\expandafter\def\csname v@ep.opus.p2rb.mean\endcsname{$0.000$}
\expandafter\def\csname v@ep.opus.p2rb.n\endcsname{255}
\expandafter\def\csname v@ep.opus.p2rb.pb\endcsname{$1.000$}
\expandafter\def\csname v@ep.opus.p2rb.pbraw\endcsname{$1.000$}
\expandafter\def\csname v@ep.opus.p2rb.praw\endcsname{$1.000$}
\expandafter\def\csname v@ep.opus.p2rb.pt\endcsname{$1.000$}
\expandafter\def\csname v@ep.opus.p2rb.upper\endcsname{$+0.004$}
\expandafter\def\csname v@ep.opus.p2rb.upperci\endcsname{[$-0.004$, $+0.012$]}
\expandafter\def\csname v@ep.opus.p3mrb.ci\endcsname{[$-0.029$, $+0.029$]}
\expandafter\def\csname v@ep.opus.p3mrb.lower\endcsname{$0.000$}
\expandafter\def\csname v@ep.opus.p3mrb.lowerci\endcsname{[$-0.029$, $+0.029$]}
\expandafter\def\csname v@ep.opus.p3mrb.mean\endcsname{$0.000$}
\expandafter\def\csname v@ep.opus.p3mrb.n\endcsname{256}
\expandafter\def\csname v@ep.opus.p3mrb.pb\endcsname{$1.000$}
\expandafter\def\csname v@ep.opus.p3mrb.pbraw\endcsname{$1.000$}
\expandafter\def\csname v@ep.opus.p3mrb.praw\endcsname{$1.000$}
\expandafter\def\csname v@ep.opus.p3mrb.pt\endcsname{$1.000$}
\expandafter\def\csname v@ep.opus.p3mrb.upper\endcsname{$0.000$}
\expandafter\def\csname v@ep.opus.p3mrb.upperci\endcsname{[$-0.029$, $+0.029$]}
\expandafter\def\csname v@ep.opus.planted.k\endcsname{0}
\expandafter\def\csname v@ep.opus.planted.n\endcsname{1,024}
\expandafter\def\csname v@ep.opus.rate.bad.failure.n\endcsname{0.0\%}
\expandafter\def\csname v@ep.opus.rate.bad.failure.neutral\endcsname{0.0\%}
\expandafter\def\csname v@ep.opus.rate.bad.failure.nx\endcsname{0.0\%}
\expandafter\def\csname v@ep.opus.rate.bad.failure.planted\endcsname{0.0\%}
\expandafter\def\csname v@ep.opus.rate.bad.failure.rb\endcsname{0.0\%}
\expandafter\def\csname v@ep.opus.rate.bad.failure.rbd\endcsname{0.0\%}
\expandafter\def\csname v@ep.opus.rate.bad.failure.rbdx\endcsname{0.0\%}
\expandafter\def\csname v@ep.opus.rate.bad.failure.rbx\endcsname{0.0\%}
\expandafter\def\csname v@ep.opus.rate.bad.success.n\endcsname{0.0\%}
\expandafter\def\csname v@ep.opus.rate.bad.success.neutral\endcsname{0.0\%}
\expandafter\def\csname v@ep.opus.rate.bad.success.nx\endcsname{0.0\%}
\expandafter\def\csname v@ep.opus.rate.bad.success.planted\endcsname{0.0\%}
\expandafter\def\csname v@ep.opus.rate.bad.success.rb\endcsname{0.0\%}
\expandafter\def\csname v@ep.opus.rate.bad.success.rbd\endcsname{0.0\%}
\expandafter\def\csname v@ep.opus.rate.bad.success.rbdx\endcsname{0.0\%}
\expandafter\def\csname v@ep.opus.rate.bad.success.rbx\endcsname{0.0\%}
\expandafter\def\csname v@ep.opus.rate.good.failure.n\endcsname{0.0\%}
\expandafter\def\csname v@ep.opus.rate.good.failure.neutral\endcsname{0.0\%}
\expandafter\def\csname v@ep.opus.rate.good.failure.nx\endcsname{0.0\%}
\expandafter\def\csname v@ep.opus.rate.good.failure.planted\endcsname{0.0\%}
\expandafter\def\csname v@ep.opus.rate.good.failure.rb\endcsname{0.0\%}
\expandafter\def\csname v@ep.opus.rate.good.failure.rbd\endcsname{0.0\%}
\expandafter\def\csname v@ep.opus.rate.good.failure.rbdx\endcsname{0.0\%}
\expandafter\def\csname v@ep.opus.rate.good.failure.rbx\endcsname{0.0\%}
\expandafter\def\csname v@ep.opus.rate.good.success.n\endcsname{0.0\%}
\expandafter\def\csname v@ep.opus.rate.good.success.neutral\endcsname{0.0\%}
\expandafter\def\csname v@ep.opus.rate.good.success.nx\endcsname{0.0\%}
\expandafter\def\csname v@ep.opus.rate.good.success.planted\endcsname{0.0\%}
\expandafter\def\csname v@ep.opus.rate.good.success.rb\endcsname{0.0\%}
\expandafter\def\csname v@ep.opus.rate.good.success.rbd\endcsname{0.0\%}
\expandafter\def\csname v@ep.opus.rate.good.success.rbdx\endcsname{0.0\%}
\expandafter\def\csname v@ep.opus.rate.good.success.rbx\endcsname{0.0\%}
\expandafter\def\csname v@ep.opus.sus.ci\endcsname{[$-0.014$, $+0.014$]}
\expandafter\def\csname v@ep.opus.sus.mean\endcsname{$0.000$}
\expandafter\def\csname v@ep.opus.sus.n\endcsname{256}
\expandafter\def\csname v@ep.opus.sus.pbraw\endcsname{$1.000$}
\expandafter\def\csname v@ep.opus.sus.praw\endcsname{$1.000$}
\expandafter\def\csname v@ep.opus.word.pthree\endcsname{\emph{bounded below 0.10}}
\expandafter\def\csname v@ep.opus.word.ptwo\endcsname{\emph{bounded below 0.10}}
\expandafter\def\csname v@ep.opus.word.sus\endcsname{\emph{increment below 0.10}}
\expandafter\def\csname v@ep.qwen.a2rb.ci\endcsname{[$+0.283$, $+0.412$]}
\expandafter\def\csname v@ep.qwen.a2rb.lower\endcsname{$+0.348$}
\expandafter\def\csname v@ep.qwen.a2rb.lowerci\endcsname{[$+0.283$, $+0.412$]}
\expandafter\def\csname v@ep.qwen.a2rb.mean\endcsname{$+0.348$}
\expandafter\def\csname v@ep.qwen.a2rb.n\endcsname{256}
\expandafter\def\csname v@ep.qwen.a2rb.pbraw\endcsname{$3.3\times10^{-20}$}
\expandafter\def\csname v@ep.qwen.a2rb.praw\endcsname{$5.0\times10^{-22}$}
\expandafter\def\csname v@ep.qwen.a2rb.upper\endcsname{$+0.348$}
\expandafter\def\csname v@ep.qwen.a2rb.upperci\endcsname{[$+0.283$, $+0.412$]}
\expandafter\def\csname v@ep.qwen.a3rb.ci\endcsname{[$-0.043$, $+0.003$]}
\expandafter\def\csname v@ep.qwen.a3rb.lower\endcsname{$-0.023$}
\expandafter\def\csname v@ep.qwen.a3rb.lowerci\endcsname{[$-0.048$, $+0.001$]}
\expandafter\def\csname v@ep.qwen.a3rb.mean\endcsname{$-0.020$}
\expandafter\def\csname v@ep.qwen.a3rb.n\endcsname{255}
\expandafter\def\csname v@ep.qwen.a3rb.pbraw\endcsname{$1.000$}
\expandafter\def\csname v@ep.qwen.a3rb.praw\endcsname{$0.096$}
\expandafter\def\csname v@ep.qwen.a3rb.upper\endcsname{$-0.020$}
\expandafter\def\csname v@ep.qwen.a3rb.upperci\endcsname{[$-0.043$, $+0.003$]}
\expandafter\def\csname v@ep.qwen.count.bad.failure.n\endcsname{0/256}
\expandafter\def\csname v@ep.qwen.count.bad.failure.neutral\endcsname{2/256}
\expandafter\def\csname v@ep.qwen.count.bad.failure.nx\endcsname{1/256}
\expandafter\def\csname v@ep.qwen.count.bad.failure.planted\endcsname{131/255}
\expandafter\def\csname v@ep.qwen.count.bad.failure.rb\endcsname{0/256}
\expandafter\def\csname v@ep.qwen.count.bad.failure.rbd\endcsname{2/256}
\expandafter\def\csname v@ep.qwen.count.bad.failure.rbdx\endcsname{0/256}
\expandafter\def\csname v@ep.qwen.count.bad.failure.rbx\endcsname{6/256}
\expandafter\def\csname v@ep.qwen.count.bad.success.n\endcsname{0/256}
\expandafter\def\csname v@ep.qwen.count.bad.success.neutral\endcsname{8/255}
\expandafter\def\csname v@ep.qwen.count.bad.success.nx\endcsname{0/256}
\expandafter\def\csname v@ep.qwen.count.bad.success.planted\endcsname{181/255}
\expandafter\def\csname v@ep.qwen.count.bad.success.rb\endcsname{3/256}
\expandafter\def\csname v@ep.qwen.count.bad.success.rbd\endcsname{1/256}
\expandafter\def\csname v@ep.qwen.count.bad.success.rbdx\endcsname{0/255}
\expandafter\def\csname v@ep.qwen.count.bad.success.rbx\endcsname{0/255}
\expandafter\def\csname v@ep.qwen.count.good.failure.n\endcsname{17/256}
\expandafter\def\csname v@ep.qwen.count.good.failure.neutral\endcsname{11/256}
\expandafter\def\csname v@ep.qwen.count.good.failure.nx\endcsname{7/256}
\expandafter\def\csname v@ep.qwen.count.good.failure.planted\endcsname{130/255}
\expandafter\def\csname v@ep.qwen.count.good.failure.rb\endcsname{100/256}
\expandafter\def\csname v@ep.qwen.count.good.failure.rbd\endcsname{84/254}
\expandafter\def\csname v@ep.qwen.count.good.failure.rbdx\endcsname{13/256}
\expandafter\def\csname v@ep.qwen.count.good.failure.rbx\endcsname{9/256}
\expandafter\def\csname v@ep.qwen.count.good.success.n\endcsname{0/255}
\expandafter\def\csname v@ep.qwen.count.good.success.neutral\endcsname{2/256}
\expandafter\def\csname v@ep.qwen.count.good.success.nx\endcsname{1/256}
\expandafter\def\csname v@ep.qwen.count.good.success.planted\endcsname{104/254}
\expandafter\def\csname v@ep.qwen.count.good.success.rb\endcsname{1/256}
\expandafter\def\csname v@ep.qwen.count.good.success.rbd\endcsname{0/256}
\expandafter\def\csname v@ep.qwen.count.good.success.rbdx\endcsname{3/255}
\expandafter\def\csname v@ep.qwen.count.good.success.rbx\endcsname{2/256}
\expandafter\def\csname v@ep.qwen.id\endcsname{\texttt{qwen/qwen3.8-max-0902}}
\expandafter\def\csname v@ep.qwen.k.bad.failure.n\endcsname{0}
\expandafter\def\csname v@ep.qwen.k.bad.failure.neutral\endcsname{2}
\expandafter\def\csname v@ep.qwen.k.bad.failure.nx\endcsname{1}
\expandafter\def\csname v@ep.qwen.k.bad.failure.planted\endcsname{131}
\expandafter\def\csname v@ep.qwen.k.bad.failure.rb\endcsname{0}
\expandafter\def\csname v@ep.qwen.k.bad.failure.rbd\endcsname{2}
\expandafter\def\csname v@ep.qwen.k.bad.failure.rbdx\endcsname{0}
\expandafter\def\csname v@ep.qwen.k.bad.failure.rbx\endcsname{6}
\expandafter\def\csname v@ep.qwen.k.bad.success.n\endcsname{0}
\expandafter\def\csname v@ep.qwen.k.bad.success.neutral\endcsname{8}
\expandafter\def\csname v@ep.qwen.k.bad.success.nx\endcsname{0}
\expandafter\def\csname v@ep.qwen.k.bad.success.planted\endcsname{181}
\expandafter\def\csname v@ep.qwen.k.bad.success.rb\endcsname{3}
\expandafter\def\csname v@ep.qwen.k.bad.success.rbd\endcsname{1}
\expandafter\def\csname v@ep.qwen.k.bad.success.rbdx\endcsname{0}
\expandafter\def\csname v@ep.qwen.k.bad.success.rbx\endcsname{0}
\expandafter\def\csname v@ep.qwen.k.good.failure.n\endcsname{17}
\expandafter\def\csname v@ep.qwen.k.good.failure.neutral\endcsname{11}
\expandafter\def\csname v@ep.qwen.k.good.failure.nx\endcsname{7}
\expandafter\def\csname v@ep.qwen.k.good.failure.planted\endcsname{130}
\expandafter\def\csname v@ep.qwen.k.good.failure.rb\endcsname{100}
\expandafter\def\csname v@ep.qwen.k.good.failure.rbd\endcsname{84}
\expandafter\def\csname v@ep.qwen.k.good.failure.rbdx\endcsname{13}
\expandafter\def\csname v@ep.qwen.k.good.failure.rbx\endcsname{9}
\expandafter\def\csname v@ep.qwen.k.good.success.n\endcsname{0}
\expandafter\def\csname v@ep.qwen.k.good.success.neutral\endcsname{2}
\expandafter\def\csname v@ep.qwen.k.good.success.nx\endcsname{1}
\expandafter\def\csname v@ep.qwen.k.good.success.planted\endcsname{104}
\expandafter\def\csname v@ep.qwen.k.good.success.rb\endcsname{1}
\expandafter\def\csname v@ep.qwen.k.good.success.rbd\endcsname{0}
\expandafter\def\csname v@ep.qwen.k.good.success.rbdx\endcsname{3}
\expandafter\def\csname v@ep.qwen.k.good.success.rbx\endcsname{2}
\expandafter\def\csname v@ep.qwen.l2brb.ci\endcsname{[$-0.009$, $+0.017$]}
\expandafter\def\csname v@ep.qwen.l2brb.mean\endcsname{$+0.004$}
\expandafter\def\csname v@ep.qwen.l2brb.n\endcsname{256}
\expandafter\def\csname v@ep.qwen.l2brb.pbraw\endcsname{$1.000$}
\expandafter\def\csname v@ep.qwen.l2brb.praw\endcsname{$0.565$}
\expandafter\def\csname v@ep.qwen.l2rb.ci\endcsname{[$+0.293$, $+0.418$]}
\expandafter\def\csname v@ep.qwen.l2rb.lower\endcsname{$+0.355$}
\expandafter\def\csname v@ep.qwen.l2rb.lowerci\endcsname{[$+0.293$, $+0.418$]}
\expandafter\def\csname v@ep.qwen.l2rb.mean\endcsname{$+0.355$}
\expandafter\def\csname v@ep.qwen.l2rb.n\endcsname{256}
\expandafter\def\csname v@ep.qwen.l2rb.pbraw\endcsname{$2.1\times10^{-21}$}
\expandafter\def\csname v@ep.qwen.l2rb.praw\endcsname{$1.1\times10^{-23}$}
\expandafter\def\csname v@ep.qwen.l2rb.upper\endcsname{$+0.355$}
\expandafter\def\csname v@ep.qwen.l2rb.upperci\endcsname{[$+0.293$, $+0.418$]}
\expandafter\def\csname v@ep.qwen.l3rb.ci\endcsname{[$-0.002$, $+0.025$]}
\expandafter\def\csname v@ep.qwen.l3rb.lower\endcsname{$+0.008$}
\expandafter\def\csname v@ep.qwen.l3rb.lowerci\endcsname{[$-0.008$, $+0.023$]}
\expandafter\def\csname v@ep.qwen.l3rb.mean\endcsname{$+0.012$}
\expandafter\def\csname v@ep.qwen.l3rb.n\endcsname{255}
\expandafter\def\csname v@ep.qwen.l3rb.pbraw\endcsname{$0.187$}
\expandafter\def\csname v@ep.qwen.l3rb.praw\endcsname{$0.083$}
\expandafter\def\csname v@ep.qwen.l3rb.upper\endcsname{$+0.012$}
\expandafter\def\csname v@ep.qwen.l3rb.upperci\endcsname{[$-0.002$, $+0.025$]}
\expandafter\def\csname v@ep.qwen.n.bad.failure.n\endcsname{256}
\expandafter\def\csname v@ep.qwen.n.bad.failure.neutral\endcsname{256}
\expandafter\def\csname v@ep.qwen.n.bad.failure.nx\endcsname{256}
\expandafter\def\csname v@ep.qwen.n.bad.failure.planted\endcsname{255}
\expandafter\def\csname v@ep.qwen.n.bad.failure.rb\endcsname{256}
\expandafter\def\csname v@ep.qwen.n.bad.failure.rbd\endcsname{256}
\expandafter\def\csname v@ep.qwen.n.bad.failure.rbdx\endcsname{256}
\expandafter\def\csname v@ep.qwen.n.bad.failure.rbx\endcsname{256}
\expandafter\def\csname v@ep.qwen.n.bad.success.n\endcsname{256}
\expandafter\def\csname v@ep.qwen.n.bad.success.neutral\endcsname{255}
\expandafter\def\csname v@ep.qwen.n.bad.success.nx\endcsname{256}
\expandafter\def\csname v@ep.qwen.n.bad.success.planted\endcsname{255}
\expandafter\def\csname v@ep.qwen.n.bad.success.rb\endcsname{256}
\expandafter\def\csname v@ep.qwen.n.bad.success.rbd\endcsname{256}
\expandafter\def\csname v@ep.qwen.n.bad.success.rbdx\endcsname{255}
\expandafter\def\csname v@ep.qwen.n.bad.success.rbx\endcsname{255}
\expandafter\def\csname v@ep.qwen.n.good.failure.n\endcsname{256}
\expandafter\def\csname v@ep.qwen.n.good.failure.neutral\endcsname{256}
\expandafter\def\csname v@ep.qwen.n.good.failure.nx\endcsname{256}
\expandafter\def\csname v@ep.qwen.n.good.failure.planted\endcsname{255}
\expandafter\def\csname v@ep.qwen.n.good.failure.rb\endcsname{256}
\expandafter\def\csname v@ep.qwen.n.good.failure.rbd\endcsname{254}
\expandafter\def\csname v@ep.qwen.n.good.failure.rbdx\endcsname{256}
\expandafter\def\csname v@ep.qwen.n.good.failure.rbx\endcsname{256}
\expandafter\def\csname v@ep.qwen.n.good.success.n\endcsname{255}
\expandafter\def\csname v@ep.qwen.n.good.success.neutral\endcsname{256}
\expandafter\def\csname v@ep.qwen.n.good.success.nx\endcsname{256}
\expandafter\def\csname v@ep.qwen.n.good.success.planted\endcsname{254}
\expandafter\def\csname v@ep.qwen.n.good.success.rb\endcsname{256}
\expandafter\def\csname v@ep.qwen.n.good.success.rbd\endcsname{256}
\expandafter\def\csname v@ep.qwen.n.good.success.rbdx\endcsname{255}
\expandafter\def\csname v@ep.qwen.n.good.success.rbx\endcsname{256}
\expandafter\def\csname v@ep.qwen.name\endcsname{Qwen 3.8 Max}
\expandafter\def\csname v@ep.qwen.namesworse\endcsname{39\%}
\expandafter\def\csname v@ep.qwen.p2rb.ci\endcsname{[$+0.286$, $+0.417$]}
\expandafter\def\csname v@ep.qwen.p2rb.lower\endcsname{$+0.352$}
\expandafter\def\csname v@ep.qwen.p2rb.lowerci\endcsname{[$+0.286$, $+0.417$]}
\expandafter\def\csname v@ep.qwen.p2rb.mean\endcsname{$+0.352$}
\expandafter\def\csname v@ep.qwen.p2rb.n\endcsname{256}
\expandafter\def\csname v@ep.qwen.p2rb.pb\endcsname{$2.1\times10^{-12}$}
\expandafter\def\csname v@ep.qwen.p2rb.pbraw\endcsname{$1.0\times10^{-12}$}
\expandafter\def\csname v@ep.qwen.p2rb.praw\endcsname{$7.4\times10^{-22}$}
\expandafter\def\csname v@ep.qwen.p2rb.pt\endcsname{$1.5\times10^{-21}$}
\expandafter\def\csname v@ep.qwen.p2rb.upper\endcsname{$+0.352$}
\expandafter\def\csname v@ep.qwen.p2rb.upperci\endcsname{[$+0.286$, $+0.417$]}
\expandafter\def\csname v@ep.qwen.p3mrb.ci\endcsname{[$-0.037$, $+0.014$]}
\expandafter\def\csname v@ep.qwen.p3mrb.lower\endcsname{$-0.016$}
\expandafter\def\csname v@ep.qwen.p3mrb.lowerci\endcsname{[$-0.042$, $+0.011$]}
\expandafter\def\csname v@ep.qwen.p3mrb.mean\endcsname{$-0.012$}
\expandafter\def\csname v@ep.qwen.p3mrb.n\endcsname{255}
\expandafter\def\csname v@ep.qwen.p3mrb.pb\endcsname{$1.000$}
\expandafter\def\csname v@ep.qwen.p3mrb.pbraw\endcsname{$1.000$}
\expandafter\def\csname v@ep.qwen.p3mrb.praw\endcsname{$0.367$}
\expandafter\def\csname v@ep.qwen.p3mrb.pt\endcsname{$0.367$}
\expandafter\def\csname v@ep.qwen.p3mrb.upper\endcsname{$-0.012$}
\expandafter\def\csname v@ep.qwen.p3mrb.upperci\endcsname{[$-0.037$, $+0.014$]}
\expandafter\def\csname v@ep.qwen.planted.k\endcsname{546}
\expandafter\def\csname v@ep.qwen.planted.n\endcsname{1,019}
\expandafter\def\csname v@ep.qwen.rate.bad.failure.n\endcsname{0.0\%}
\expandafter\def\csname v@ep.qwen.rate.bad.failure.neutral\endcsname{0.8\%}
\expandafter\def\csname v@ep.qwen.rate.bad.failure.nx\endcsname{0.4\%}
\expandafter\def\csname v@ep.qwen.rate.bad.failure.planted\endcsname{51.4\%}
\expandafter\def\csname v@ep.qwen.rate.bad.failure.rb\endcsname{0.0\%}
\expandafter\def\csname v@ep.qwen.rate.bad.failure.rbd\endcsname{0.8\%}
\expandafter\def\csname v@ep.qwen.rate.bad.failure.rbdx\endcsname{0.0\%}
\expandafter\def\csname v@ep.qwen.rate.bad.failure.rbx\endcsname{2.3\%}
\expandafter\def\csname v@ep.qwen.rate.bad.success.n\endcsname{0.0\%}
\expandafter\def\csname v@ep.qwen.rate.bad.success.neutral\endcsname{3.1\%}
\expandafter\def\csname v@ep.qwen.rate.bad.success.nx\endcsname{0.0\%}
\expandafter\def\csname v@ep.qwen.rate.bad.success.planted\endcsname{71.0\%}
\expandafter\def\csname v@ep.qwen.rate.bad.success.rb\endcsname{1.2\%}
\expandafter\def\csname v@ep.qwen.rate.bad.success.rbd\endcsname{0.4\%}
\expandafter\def\csname v@ep.qwen.rate.bad.success.rbdx\endcsname{0.0\%}
\expandafter\def\csname v@ep.qwen.rate.bad.success.rbx\endcsname{0.0\%}
\expandafter\def\csname v@ep.qwen.rate.good.failure.n\endcsname{6.6\%}
\expandafter\def\csname v@ep.qwen.rate.good.failure.neutral\endcsname{4.3\%}
\expandafter\def\csname v@ep.qwen.rate.good.failure.nx\endcsname{2.7\%}
\expandafter\def\csname v@ep.qwen.rate.good.failure.planted\endcsname{51.0\%}
\expandafter\def\csname v@ep.qwen.rate.good.failure.rb\endcsname{39.1\%}
\expandafter\def\csname v@ep.qwen.rate.good.failure.rbd\endcsname{33.1\%}
\expandafter\def\csname v@ep.qwen.rate.good.failure.rbdx\endcsname{5.1\%}
\expandafter\def\csname v@ep.qwen.rate.good.failure.rbx\endcsname{3.5\%}
\expandafter\def\csname v@ep.qwen.rate.good.success.n\endcsname{0.0\%}
\expandafter\def\csname v@ep.qwen.rate.good.success.neutral\endcsname{0.8\%}
\expandafter\def\csname v@ep.qwen.rate.good.success.nx\endcsname{0.4\%}
\expandafter\def\csname v@ep.qwen.rate.good.success.planted\endcsname{40.9\%}
\expandafter\def\csname v@ep.qwen.rate.good.success.rb\endcsname{0.4\%}
\expandafter\def\csname v@ep.qwen.rate.good.success.rbd\endcsname{0.0\%}
\expandafter\def\csname v@ep.qwen.rate.good.success.rbdx\endcsname{1.2\%}
\expandafter\def\csname v@ep.qwen.rate.good.success.rbx\endcsname{0.8\%}
\expandafter\def\csname v@ep.qwen.sus.ci\endcsname{[$+0.480$, $+0.548$]}
\expandafter\def\csname v@ep.qwen.sus.mean\endcsname{$+0.514$}
\expandafter\def\csname v@ep.qwen.sus.n\endcsname{251}
\expandafter\def\csname v@ep.qwen.sus.pbraw\endcsname{$4.9\times10^{-35}$}
\expandafter\def\csname v@ep.qwen.sus.praw\endcsname{$3.7\times10^{-84}$}
\expandafter\def\csname v@ep.qwen.word.pthree\endcsname{\emph{bounded below 0.10}}
\expandafter\def\csname v@ep.qwen.word.ptwo\endcsname{\emph{luck-carrying}}
\expandafter\def\csname v@ep.qwen.word.sus\endcsname{\emph{increment above 0.10}}
\expandafter\def\csname v@ep.remedy.n.boundedcredited\endcsname{Claude Haiku 5.5 and GPT-6 Luna}
\expandafter\def\csname v@ep.remedy.n.credited\endcsname{Gemini 3.8 Flash, Grok 4.7, Qwen 3.8 Max, DeepSeek V4 Pro and Kimi K3}
\expandafter\def\csname v@ep.remedy.n.ncredited\endcsname{5}
\expandafter\def\csname v@ep.remedy.n.notcredited\endcsname{Mistral Large 4}
\expandafter\def\csname v@ep.remedy.rbd.boundedcredited\endcsname{GPT-6 Luna}
\expandafter\def\csname v@ep.remedy.rbd.credited\endcsname{Gemini 3.8 Flash, Grok 4.7 and DeepSeek V4 Pro}
\expandafter\def\csname v@ep.remedy.rbd.ncredited\endcsname{3}
\expandafter\def\csname v@ep.remedy.rbd.notcredited\endcsname{Qwen 3.8 Max, Mistral Large 4 and Kimi K3}
\expandafter\def\csname v@ep.sol.a2rb.ci\endcsname{[$-0.014$, $+0.014$]}
\expandafter\def\csname v@ep.sol.a2rb.lower\endcsname{$0.000$}
\expandafter\def\csname v@ep.sol.a2rb.lowerci\endcsname{[$-0.014$, $+0.014$]}
\expandafter\def\csname v@ep.sol.a2rb.mean\endcsname{$0.000$}
\expandafter\def\csname v@ep.sol.a2rb.n\endcsname{256}
\expandafter\def\csname v@ep.sol.a2rb.pbraw\endcsname{$1.000$}
\expandafter\def\csname v@ep.sol.a2rb.praw\endcsname{$1.000$}
\expandafter\def\csname v@ep.sol.a2rb.upper\endcsname{$0.000$}
\expandafter\def\csname v@ep.sol.a2rb.upperci\endcsname{[$-0.014$, $+0.014$]}
\expandafter\def\csname v@ep.sol.a3rb.ci\endcsname{[$-0.014$, $+0.014$]}
\expandafter\def\csname v@ep.sol.a3rb.lower\endcsname{$0.000$}
\expandafter\def\csname v@ep.sol.a3rb.lowerci\endcsname{[$-0.014$, $+0.014$]}
\expandafter\def\csname v@ep.sol.a3rb.mean\endcsname{$0.000$}
\expandafter\def\csname v@ep.sol.a3rb.n\endcsname{256}
\expandafter\def\csname v@ep.sol.a3rb.pbraw\endcsname{$1.000$}
\expandafter\def\csname v@ep.sol.a3rb.praw\endcsname{$1.000$}
\expandafter\def\csname v@ep.sol.a3rb.upper\endcsname{$0.000$}
\expandafter\def\csname v@ep.sol.a3rb.upperci\endcsname{[$-0.014$, $+0.014$]}
\expandafter\def\csname v@ep.sol.count.bad.failure.n\endcsname{0/256}
\expandafter\def\csname v@ep.sol.count.bad.failure.neutral\endcsname{0/256}
\expandafter\def\csname v@ep.sol.count.bad.failure.nx\endcsname{0/256}
\expandafter\def\csname v@ep.sol.count.bad.failure.planted\endcsname{0/256}
\expandafter\def\csname v@ep.sol.count.bad.failure.rb\endcsname{0/256}
\expandafter\def\csname v@ep.sol.count.bad.failure.rbd\endcsname{0/256}
\expandafter\def\csname v@ep.sol.count.bad.failure.rbdx\endcsname{0/256}
\expandafter\def\csname v@ep.sol.count.bad.failure.rbx\endcsname{0/256}
\expandafter\def\csname v@ep.sol.count.bad.success.n\endcsname{0/256}
\expandafter\def\csname v@ep.sol.count.bad.success.neutral\endcsname{0/256}
\expandafter\def\csname v@ep.sol.count.bad.success.nx\endcsname{0/256}
\expandafter\def\csname v@ep.sol.count.bad.success.planted\endcsname{0/256}
\expandafter\def\csname v@ep.sol.count.bad.success.rb\endcsname{0/256}
\expandafter\def\csname v@ep.sol.count.bad.success.rbd\endcsname{0/256}
\expandafter\def\csname v@ep.sol.count.bad.success.rbdx\endcsname{0/256}
\expandafter\def\csname v@ep.sol.count.bad.success.rbx\endcsname{0/256}
\expandafter\def\csname v@ep.sol.count.good.failure.n\endcsname{0/256}
\expandafter\def\csname v@ep.sol.count.good.failure.neutral\endcsname{0/256}
\expandafter\def\csname v@ep.sol.count.good.failure.nx\endcsname{0/256}
\expandafter\def\csname v@ep.sol.count.good.failure.planted\endcsname{0/256}
\expandafter\def\csname v@ep.sol.count.good.failure.rb\endcsname{0/256}
\expandafter\def\csname v@ep.sol.count.good.failure.rbd\endcsname{0/256}
\expandafter\def\csname v@ep.sol.count.good.failure.rbdx\endcsname{0/256}
\expandafter\def\csname v@ep.sol.count.good.failure.rbx\endcsname{0/256}
\expandafter\def\csname v@ep.sol.count.good.success.n\endcsname{0/256}
\expandafter\def\csname v@ep.sol.count.good.success.neutral\endcsname{0/256}
\expandafter\def\csname v@ep.sol.count.good.success.nx\endcsname{0/256}
\expandafter\def\csname v@ep.sol.count.good.success.planted\endcsname{0/256}
\expandafter\def\csname v@ep.sol.count.good.success.rb\endcsname{0/256}
\expandafter\def\csname v@ep.sol.count.good.success.rbd\endcsname{0/256}
\expandafter\def\csname v@ep.sol.count.good.success.rbdx\endcsname{0/256}
\expandafter\def\csname v@ep.sol.count.good.success.rbx\endcsname{0/256}
\expandafter\def\csname v@ep.sol.id\endcsname{\texttt{openai/gpt-6.1-sol}}
\expandafter\def\csname v@ep.sol.k.bad.failure.n\endcsname{0}
\expandafter\def\csname v@ep.sol.k.bad.failure.neutral\endcsname{0}
\expandafter\def\csname v@ep.sol.k.bad.failure.nx\endcsname{0}
\expandafter\def\csname v@ep.sol.k.bad.failure.planted\endcsname{0}
\expandafter\def\csname v@ep.sol.k.bad.failure.rb\endcsname{0}
\expandafter\def\csname v@ep.sol.k.bad.failure.rbd\endcsname{0}
\expandafter\def\csname v@ep.sol.k.bad.failure.rbdx\endcsname{0}
\expandafter\def\csname v@ep.sol.k.bad.failure.rbx\endcsname{0}
\expandafter\def\csname v@ep.sol.k.bad.success.n\endcsname{0}
\expandafter\def\csname v@ep.sol.k.bad.success.neutral\endcsname{0}
\expandafter\def\csname v@ep.sol.k.bad.success.nx\endcsname{0}
\expandafter\def\csname v@ep.sol.k.bad.success.planted\endcsname{0}
\expandafter\def\csname v@ep.sol.k.bad.success.rb\endcsname{0}
\expandafter\def\csname v@ep.sol.k.bad.success.rbd\endcsname{0}
\expandafter\def\csname v@ep.sol.k.bad.success.rbdx\endcsname{0}
\expandafter\def\csname v@ep.sol.k.bad.success.rbx\endcsname{0}
\expandafter\def\csname v@ep.sol.k.good.failure.n\endcsname{0}
\expandafter\def\csname v@ep.sol.k.good.failure.neutral\endcsname{0}
\expandafter\def\csname v@ep.sol.k.good.failure.nx\endcsname{0}
\expandafter\def\csname v@ep.sol.k.good.failure.planted\endcsname{0}
\expandafter\def\csname v@ep.sol.k.good.failure.rb\endcsname{0}
\expandafter\def\csname v@ep.sol.k.good.failure.rbd\endcsname{0}
\expandafter\def\csname v@ep.sol.k.good.failure.rbdx\endcsname{0}
\expandafter\def\csname v@ep.sol.k.good.failure.rbx\endcsname{0}
\expandafter\def\csname v@ep.sol.k.good.success.n\endcsname{0}
\expandafter\def\csname v@ep.sol.k.good.success.neutral\endcsname{0}
\expandafter\def\csname v@ep.sol.k.good.success.nx\endcsname{0}
\expandafter\def\csname v@ep.sol.k.good.success.planted\endcsname{0}
\expandafter\def\csname v@ep.sol.k.good.success.rb\endcsname{0}
\expandafter\def\csname v@ep.sol.k.good.success.rbd\endcsname{0}
\expandafter\def\csname v@ep.sol.k.good.success.rbdx\endcsname{0}
\expandafter\def\csname v@ep.sol.k.good.success.rbx\endcsname{0}
\expandafter\def\csname v@ep.sol.l2brb.ci\endcsname{[$-0.014$, $+0.014$]}
\expandafter\def\csname v@ep.sol.l2brb.mean\endcsname{$0.000$}
\expandafter\def\csname v@ep.sol.l2brb.n\endcsname{256}
\expandafter\def\csname v@ep.sol.l2brb.pbraw\endcsname{$1.000$}
\expandafter\def\csname v@ep.sol.l2brb.praw\endcsname{$1.000$}
\expandafter\def\csname v@ep.sol.l2rb.ci\endcsname{[$-0.014$, $+0.014$]}
\expandafter\def\csname v@ep.sol.l2rb.lower\endcsname{$0.000$}
\expandafter\def\csname v@ep.sol.l2rb.lowerci\endcsname{[$-0.014$, $+0.014$]}
\expandafter\def\csname v@ep.sol.l2rb.mean\endcsname{$0.000$}
\expandafter\def\csname v@ep.sol.l2rb.n\endcsname{256}
\expandafter\def\csname v@ep.sol.l2rb.pbraw\endcsname{$1.000$}
\expandafter\def\csname v@ep.sol.l2rb.praw\endcsname{$1.000$}
\expandafter\def\csname v@ep.sol.l2rb.upper\endcsname{$0.000$}
\expandafter\def\csname v@ep.sol.l2rb.upperci\endcsname{[$-0.014$, $+0.014$]}
\expandafter\def\csname v@ep.sol.l3rb.ci\endcsname{[$-0.014$, $+0.014$]}
\expandafter\def\csname v@ep.sol.l3rb.lower\endcsname{$0.000$}
\expandafter\def\csname v@ep.sol.l3rb.lowerci\endcsname{[$-0.014$, $+0.014$]}
\expandafter\def\csname v@ep.sol.l3rb.mean\endcsname{$0.000$}
\expandafter\def\csname v@ep.sol.l3rb.n\endcsname{256}
\expandafter\def\csname v@ep.sol.l3rb.pbraw\endcsname{$1.000$}
\expandafter\def\csname v@ep.sol.l3rb.praw\endcsname{$1.000$}
\expandafter\def\csname v@ep.sol.l3rb.upper\endcsname{$0.000$}
\expandafter\def\csname v@ep.sol.l3rb.upperci\endcsname{[$-0.014$, $+0.014$]}
\expandafter\def\csname v@ep.sol.n.bad.failure.n\endcsname{256}
\expandafter\def\csname v@ep.sol.n.bad.failure.neutral\endcsname{256}
\expandafter\def\csname v@ep.sol.n.bad.failure.nx\endcsname{256}
\expandafter\def\csname v@ep.sol.n.bad.failure.planted\endcsname{256}
\expandafter\def\csname v@ep.sol.n.bad.failure.rb\endcsname{256}
\expandafter\def\csname v@ep.sol.n.bad.failure.rbd\endcsname{256}
\expandafter\def\csname v@ep.sol.n.bad.failure.rbdx\endcsname{256}
\expandafter\def\csname v@ep.sol.n.bad.failure.rbx\endcsname{256}
\expandafter\def\csname v@ep.sol.n.bad.success.n\endcsname{256}
\expandafter\def\csname v@ep.sol.n.bad.success.neutral\endcsname{256}
\expandafter\def\csname v@ep.sol.n.bad.success.nx\endcsname{256}
\expandafter\def\csname v@ep.sol.n.bad.success.planted\endcsname{256}
\expandafter\def\csname v@ep.sol.n.bad.success.rb\endcsname{256}
\expandafter\def\csname v@ep.sol.n.bad.success.rbd\endcsname{256}
\expandafter\def\csname v@ep.sol.n.bad.success.rbdx\endcsname{256}
\expandafter\def\csname v@ep.sol.n.bad.success.rbx\endcsname{256}
\expandafter\def\csname v@ep.sol.n.good.failure.n\endcsname{256}
\expandafter\def\csname v@ep.sol.n.good.failure.neutral\endcsname{256}
\expandafter\def\csname v@ep.sol.n.good.failure.nx\endcsname{256}
\expandafter\def\csname v@ep.sol.n.good.failure.planted\endcsname{256}
\expandafter\def\csname v@ep.sol.n.good.failure.rb\endcsname{256}
\expandafter\def\csname v@ep.sol.n.good.failure.rbd\endcsname{256}
\expandafter\def\csname v@ep.sol.n.good.failure.rbdx\endcsname{256}
\expandafter\def\csname v@ep.sol.n.good.failure.rbx\endcsname{256}
\expandafter\def\csname v@ep.sol.n.good.success.n\endcsname{256}
\expandafter\def\csname v@ep.sol.n.good.success.neutral\endcsname{256}
\expandafter\def\csname v@ep.sol.n.good.success.nx\endcsname{256}
\expandafter\def\csname v@ep.sol.n.good.success.planted\endcsname{256}
\expandafter\def\csname v@ep.sol.n.good.success.rb\endcsname{256}
\expandafter\def\csname v@ep.sol.n.good.success.rbd\endcsname{256}
\expandafter\def\csname v@ep.sol.n.good.success.rbdx\endcsname{256}
\expandafter\def\csname v@ep.sol.n.good.success.rbx\endcsname{256}
\expandafter\def\csname v@ep.sol.name\endcsname{GPT-6.1 Sol}
\expandafter\def\csname v@ep.sol.namesworse\endcsname{2\%}
\expandafter\def\csname v@ep.sol.p2rb.ci\endcsname{[$-0.029$, $+0.029$]}
\expandafter\def\csname v@ep.sol.p2rb.lower\endcsname{$0.000$}
\expandafter\def\csname v@ep.sol.p2rb.lowerci\endcsname{[$-0.029$, $+0.029$]}
\expandafter\def\csname v@ep.sol.p2rb.mean\endcsname{$0.000$}
\expandafter\def\csname v@ep.sol.p2rb.n\endcsname{256}
\expandafter\def\csname v@ep.sol.p2rb.pb\endcsname{$1.000$}
\expandafter\def\csname v@ep.sol.p2rb.pbraw\endcsname{$1.000$}
\expandafter\def\csname v@ep.sol.p2rb.praw\endcsname{$1.000$}
\expandafter\def\csname v@ep.sol.p2rb.pt\endcsname{$1.000$}
\expandafter\def\csname v@ep.sol.p2rb.upper\endcsname{$0.000$}
\expandafter\def\csname v@ep.sol.p2rb.upperci\endcsname{[$-0.029$, $+0.029$]}
\expandafter\def\csname v@ep.sol.p3mrb.ci\endcsname{[$-0.029$, $+0.029$]}
\expandafter\def\csname v@ep.sol.p3mrb.lower\endcsname{$0.000$}
\expandafter\def\csname v@ep.sol.p3mrb.lowerci\endcsname{[$-0.029$, $+0.029$]}
\expandafter\def\csname v@ep.sol.p3mrb.mean\endcsname{$0.000$}
\expandafter\def\csname v@ep.sol.p3mrb.n\endcsname{256}
\expandafter\def\csname v@ep.sol.p3mrb.pb\endcsname{$1.000$}
\expandafter\def\csname v@ep.sol.p3mrb.pbraw\endcsname{$1.000$}
\expandafter\def\csname v@ep.sol.p3mrb.praw\endcsname{$1.000$}
\expandafter\def\csname v@ep.sol.p3mrb.pt\endcsname{$1.000$}
\expandafter\def\csname v@ep.sol.p3mrb.upper\endcsname{$0.000$}
\expandafter\def\csname v@ep.sol.p3mrb.upperci\endcsname{[$-0.029$, $+0.029$]}
\expandafter\def\csname v@ep.sol.planted.k\endcsname{0}
\expandafter\def\csname v@ep.sol.planted.n\endcsname{1,024}
\expandafter\def\csname v@ep.sol.rate.bad.failure.n\endcsname{0.0\%}
\expandafter\def\csname v@ep.sol.rate.bad.failure.neutral\endcsname{0.0\%}
\expandafter\def\csname v@ep.sol.rate.bad.failure.nx\endcsname{0.0\%}
\expandafter\def\csname v@ep.sol.rate.bad.failure.planted\endcsname{0.0\%}
\expandafter\def\csname v@ep.sol.rate.bad.failure.rb\endcsname{0.0\%}
\expandafter\def\csname v@ep.sol.rate.bad.failure.rbd\endcsname{0.0\%}
\expandafter\def\csname v@ep.sol.rate.bad.failure.rbdx\endcsname{0.0\%}
\expandafter\def\csname v@ep.sol.rate.bad.failure.rbx\endcsname{0.0\%}
\expandafter\def\csname v@ep.sol.rate.bad.success.n\endcsname{0.0\%}
\expandafter\def\csname v@ep.sol.rate.bad.success.neutral\endcsname{0.0\%}
\expandafter\def\csname v@ep.sol.rate.bad.success.nx\endcsname{0.0\%}
\expandafter\def\csname v@ep.sol.rate.bad.success.planted\endcsname{0.0\%}
\expandafter\def\csname v@ep.sol.rate.bad.success.rb\endcsname{0.0\%}
\expandafter\def\csname v@ep.sol.rate.bad.success.rbd\endcsname{0.0\%}
\expandafter\def\csname v@ep.sol.rate.bad.success.rbdx\endcsname{0.0\%}
\expandafter\def\csname v@ep.sol.rate.bad.success.rbx\endcsname{0.0\%}
\expandafter\def\csname v@ep.sol.rate.good.failure.n\endcsname{0.0\%}
\expandafter\def\csname v@ep.sol.rate.good.failure.neutral\endcsname{0.0\%}
\expandafter\def\csname v@ep.sol.rate.good.failure.nx\endcsname{0.0\%}
\expandafter\def\csname v@ep.sol.rate.good.failure.planted\endcsname{0.0\%}
\expandafter\def\csname v@ep.sol.rate.good.failure.rb\endcsname{0.0\%}
\expandafter\def\csname v@ep.sol.rate.good.failure.rbd\endcsname{0.0\%}
\expandafter\def\csname v@ep.sol.rate.good.failure.rbdx\endcsname{0.0\%}
\expandafter\def\csname v@ep.sol.rate.good.failure.rbx\endcsname{0.0\%}
\expandafter\def\csname v@ep.sol.rate.good.success.n\endcsname{0.0\%}
\expandafter\def\csname v@ep.sol.rate.good.success.neutral\endcsname{0.0\%}
\expandafter\def\csname v@ep.sol.rate.good.success.nx\endcsname{0.0\%}
\expandafter\def\csname v@ep.sol.rate.good.success.planted\endcsname{0.0\%}
\expandafter\def\csname v@ep.sol.rate.good.success.rb\endcsname{0.0\%}
\expandafter\def\csname v@ep.sol.rate.good.success.rbd\endcsname{0.0\%}
\expandafter\def\csname v@ep.sol.rate.good.success.rbdx\endcsname{0.0\%}
\expandafter\def\csname v@ep.sol.rate.good.success.rbx\endcsname{0.0\%}
\expandafter\def\csname v@ep.sol.sus.ci\endcsname{[$-0.014$, $+0.014$]}
\expandafter\def\csname v@ep.sol.sus.mean\endcsname{$0.000$}
\expandafter\def\csname v@ep.sol.sus.n\endcsname{256}
\expandafter\def\csname v@ep.sol.sus.pbraw\endcsname{$1.000$}
\expandafter\def\csname v@ep.sol.sus.praw\endcsname{$1.000$}
\expandafter\def\csname v@ep.sol.word.pthree\endcsname{\emph{bounded below 0.10}}
\expandafter\def\csname v@ep.sol.word.ptwo\endcsname{\emph{bounded below 0.10}}
\expandafter\def\csname v@ep.sol.word.sus\endcsname{\emph{increment below 0.10}}
\expandafter\def\csname v@ep.sonnet.a2rb.ci\endcsname{[$-0.014$, $+0.014$]}
\expandafter\def\csname v@ep.sonnet.a2rb.lower\endcsname{$0.000$}
\expandafter\def\csname v@ep.sonnet.a2rb.lowerci\endcsname{[$-0.014$, $+0.014$]}
\expandafter\def\csname v@ep.sonnet.a2rb.mean\endcsname{$0.000$}
\expandafter\def\csname v@ep.sonnet.a2rb.n\endcsname{256}
\expandafter\def\csname v@ep.sonnet.a2rb.pbraw\endcsname{$1.000$}
\expandafter\def\csname v@ep.sonnet.a2rb.praw\endcsname{$1.000$}
\expandafter\def\csname v@ep.sonnet.a2rb.upper\endcsname{$0.000$}
\expandafter\def\csname v@ep.sonnet.a2rb.upperci\endcsname{[$-0.014$, $+0.014$]}
\expandafter\def\csname v@ep.sonnet.a3rb.ci\endcsname{[$-0.014$, $+0.014$]}
\expandafter\def\csname v@ep.sonnet.a3rb.lower\endcsname{$0.000$}
\expandafter\def\csname v@ep.sonnet.a3rb.lowerci\endcsname{[$-0.014$, $+0.014$]}
\expandafter\def\csname v@ep.sonnet.a3rb.mean\endcsname{$0.000$}
\expandafter\def\csname v@ep.sonnet.a3rb.n\endcsname{256}
\expandafter\def\csname v@ep.sonnet.a3rb.pbraw\endcsname{$1.000$}
\expandafter\def\csname v@ep.sonnet.a3rb.praw\endcsname{$1.000$}
\expandafter\def\csname v@ep.sonnet.a3rb.upper\endcsname{$0.000$}
\expandafter\def\csname v@ep.sonnet.a3rb.upperci\endcsname{[$-0.014$, $+0.014$]}
\expandafter\def\csname v@ep.sonnet.count.bad.failure.n\endcsname{0/256}
\expandafter\def\csname v@ep.sonnet.count.bad.failure.neutral\endcsname{0/256}
\expandafter\def\csname v@ep.sonnet.count.bad.failure.nx\endcsname{0/256}
\expandafter\def\csname v@ep.sonnet.count.bad.failure.planted\endcsname{0/256}
\expandafter\def\csname v@ep.sonnet.count.bad.failure.rb\endcsname{0/256}
\expandafter\def\csname v@ep.sonnet.count.bad.failure.rbd\endcsname{0/256}
\expandafter\def\csname v@ep.sonnet.count.bad.failure.rbdx\endcsname{0/256}
\expandafter\def\csname v@ep.sonnet.count.bad.failure.rbx\endcsname{0/256}
\expandafter\def\csname v@ep.sonnet.count.bad.success.n\endcsname{0/256}
\expandafter\def\csname v@ep.sonnet.count.bad.success.neutral\endcsname{0/256}
\expandafter\def\csname v@ep.sonnet.count.bad.success.nx\endcsname{0/256}
\expandafter\def\csname v@ep.sonnet.count.bad.success.planted\endcsname{0/256}
\expandafter\def\csname v@ep.sonnet.count.bad.success.rb\endcsname{0/256}
\expandafter\def\csname v@ep.sonnet.count.bad.success.rbd\endcsname{0/256}
\expandafter\def\csname v@ep.sonnet.count.bad.success.rbdx\endcsname{0/256}
\expandafter\def\csname v@ep.sonnet.count.bad.success.rbx\endcsname{0/256}
\expandafter\def\csname v@ep.sonnet.count.good.failure.n\endcsname{0/256}
\expandafter\def\csname v@ep.sonnet.count.good.failure.neutral\endcsname{0/256}
\expandafter\def\csname v@ep.sonnet.count.good.failure.nx\endcsname{0/256}
\expandafter\def\csname v@ep.sonnet.count.good.failure.planted\endcsname{1/256}
\expandafter\def\csname v@ep.sonnet.count.good.failure.rb\endcsname{0/256}
\expandafter\def\csname v@ep.sonnet.count.good.failure.rbd\endcsname{0/256}
\expandafter\def\csname v@ep.sonnet.count.good.failure.rbdx\endcsname{0/256}
\expandafter\def\csname v@ep.sonnet.count.good.failure.rbx\endcsname{0/256}
\expandafter\def\csname v@ep.sonnet.count.good.success.n\endcsname{0/256}
\expandafter\def\csname v@ep.sonnet.count.good.success.neutral\endcsname{0/256}
\expandafter\def\csname v@ep.sonnet.count.good.success.nx\endcsname{0/256}
\expandafter\def\csname v@ep.sonnet.count.good.success.planted\endcsname{0/256}
\expandafter\def\csname v@ep.sonnet.count.good.success.rb\endcsname{0/256}
\expandafter\def\csname v@ep.sonnet.count.good.success.rbd\endcsname{0/256}
\expandafter\def\csname v@ep.sonnet.count.good.success.rbdx\endcsname{0/256}
\expandafter\def\csname v@ep.sonnet.count.good.success.rbx\endcsname{0/256}
\expandafter\def\csname v@ep.sonnet.id\endcsname{\texttt{anthropic/claude-sonnet-5.5}}
\expandafter\def\csname v@ep.sonnet.k.bad.failure.n\endcsname{0}
\expandafter\def\csname v@ep.sonnet.k.bad.failure.neutral\endcsname{0}
\expandafter\def\csname v@ep.sonnet.k.bad.failure.nx\endcsname{0}
\expandafter\def\csname v@ep.sonnet.k.bad.failure.planted\endcsname{0}
\expandafter\def\csname v@ep.sonnet.k.bad.failure.rb\endcsname{0}
\expandafter\def\csname v@ep.sonnet.k.bad.failure.rbd\endcsname{0}
\expandafter\def\csname v@ep.sonnet.k.bad.failure.rbdx\endcsname{0}
\expandafter\def\csname v@ep.sonnet.k.bad.failure.rbx\endcsname{0}
\expandafter\def\csname v@ep.sonnet.k.bad.success.n\endcsname{0}
\expandafter\def\csname v@ep.sonnet.k.bad.success.neutral\endcsname{0}
\expandafter\def\csname v@ep.sonnet.k.bad.success.nx\endcsname{0}
\expandafter\def\csname v@ep.sonnet.k.bad.success.planted\endcsname{0}
\expandafter\def\csname v@ep.sonnet.k.bad.success.rb\endcsname{0}
\expandafter\def\csname v@ep.sonnet.k.bad.success.rbd\endcsname{0}
\expandafter\def\csname v@ep.sonnet.k.bad.success.rbdx\endcsname{0}
\expandafter\def\csname v@ep.sonnet.k.bad.success.rbx\endcsname{0}
\expandafter\def\csname v@ep.sonnet.k.good.failure.n\endcsname{0}
\expandafter\def\csname v@ep.sonnet.k.good.failure.neutral\endcsname{0}
\expandafter\def\csname v@ep.sonnet.k.good.failure.nx\endcsname{0}
\expandafter\def\csname v@ep.sonnet.k.good.failure.planted\endcsname{1}
\expandafter\def\csname v@ep.sonnet.k.good.failure.rb\endcsname{0}
\expandafter\def\csname v@ep.sonnet.k.good.failure.rbd\endcsname{0}
\expandafter\def\csname v@ep.sonnet.k.good.failure.rbdx\endcsname{0}
\expandafter\def\csname v@ep.sonnet.k.good.failure.rbx\endcsname{0}
\expandafter\def\csname v@ep.sonnet.k.good.success.n\endcsname{0}
\expandafter\def\csname v@ep.sonnet.k.good.success.neutral\endcsname{0}
\expandafter\def\csname v@ep.sonnet.k.good.success.nx\endcsname{0}
\expandafter\def\csname v@ep.sonnet.k.good.success.planted\endcsname{0}
\expandafter\def\csname v@ep.sonnet.k.good.success.rb\endcsname{0}
\expandafter\def\csname v@ep.sonnet.k.good.success.rbd\endcsname{0}
\expandafter\def\csname v@ep.sonnet.k.good.success.rbdx\endcsname{0}
\expandafter\def\csname v@ep.sonnet.k.good.success.rbx\endcsname{0}
\expandafter\def\csname v@ep.sonnet.l2brb.ci\endcsname{[$-0.014$, $+0.014$]}
\expandafter\def\csname v@ep.sonnet.l2brb.mean\endcsname{$0.000$}
\expandafter\def\csname v@ep.sonnet.l2brb.n\endcsname{256}
\expandafter\def\csname v@ep.sonnet.l2brb.pbraw\endcsname{$1.000$}
\expandafter\def\csname v@ep.sonnet.l2brb.praw\endcsname{$1.000$}
\expandafter\def\csname v@ep.sonnet.l2rb.ci\endcsname{[$-0.014$, $+0.014$]}
\expandafter\def\csname v@ep.sonnet.l2rb.lower\endcsname{$0.000$}
\expandafter\def\csname v@ep.sonnet.l2rb.lowerci\endcsname{[$-0.014$, $+0.014$]}
\expandafter\def\csname v@ep.sonnet.l2rb.mean\endcsname{$0.000$}
\expandafter\def\csname v@ep.sonnet.l2rb.n\endcsname{256}
\expandafter\def\csname v@ep.sonnet.l2rb.pbraw\endcsname{$1.000$}
\expandafter\def\csname v@ep.sonnet.l2rb.praw\endcsname{$1.000$}
\expandafter\def\csname v@ep.sonnet.l2rb.upper\endcsname{$0.000$}
\expandafter\def\csname v@ep.sonnet.l2rb.upperci\endcsname{[$-0.014$, $+0.014$]}
\expandafter\def\csname v@ep.sonnet.l3rb.ci\endcsname{[$-0.014$, $+0.014$]}
\expandafter\def\csname v@ep.sonnet.l3rb.lower\endcsname{$0.000$}
\expandafter\def\csname v@ep.sonnet.l3rb.lowerci\endcsname{[$-0.014$, $+0.014$]}
\expandafter\def\csname v@ep.sonnet.l3rb.mean\endcsname{$0.000$}
\expandafter\def\csname v@ep.sonnet.l3rb.n\endcsname{256}
\expandafter\def\csname v@ep.sonnet.l3rb.pbraw\endcsname{$1.000$}
\expandafter\def\csname v@ep.sonnet.l3rb.praw\endcsname{$1.000$}
\expandafter\def\csname v@ep.sonnet.l3rb.upper\endcsname{$0.000$}
\expandafter\def\csname v@ep.sonnet.l3rb.upperci\endcsname{[$-0.014$, $+0.014$]}
\expandafter\def\csname v@ep.sonnet.n.bad.failure.n\endcsname{256}
\expandafter\def\csname v@ep.sonnet.n.bad.failure.neutral\endcsname{256}
\expandafter\def\csname v@ep.sonnet.n.bad.failure.nx\endcsname{256}
\expandafter\def\csname v@ep.sonnet.n.bad.failure.planted\endcsname{256}
\expandafter\def\csname v@ep.sonnet.n.bad.failure.rb\endcsname{256}
\expandafter\def\csname v@ep.sonnet.n.bad.failure.rbd\endcsname{256}
\expandafter\def\csname v@ep.sonnet.n.bad.failure.rbdx\endcsname{256}
\expandafter\def\csname v@ep.sonnet.n.bad.failure.rbx\endcsname{256}
\expandafter\def\csname v@ep.sonnet.n.bad.success.n\endcsname{256}
\expandafter\def\csname v@ep.sonnet.n.bad.success.neutral\endcsname{256}
\expandafter\def\csname v@ep.sonnet.n.bad.success.nx\endcsname{256}
\expandafter\def\csname v@ep.sonnet.n.bad.success.planted\endcsname{256}
\expandafter\def\csname v@ep.sonnet.n.bad.success.rb\endcsname{256}
\expandafter\def\csname v@ep.sonnet.n.bad.success.rbd\endcsname{256}
\expandafter\def\csname v@ep.sonnet.n.bad.success.rbdx\endcsname{256}
\expandafter\def\csname v@ep.sonnet.n.bad.success.rbx\endcsname{256}
\expandafter\def\csname v@ep.sonnet.n.good.failure.n\endcsname{256}
\expandafter\def\csname v@ep.sonnet.n.good.failure.neutral\endcsname{256}
\expandafter\def\csname v@ep.sonnet.n.good.failure.nx\endcsname{256}
\expandafter\def\csname v@ep.sonnet.n.good.failure.planted\endcsname{256}
\expandafter\def\csname v@ep.sonnet.n.good.failure.rb\endcsname{256}
\expandafter\def\csname v@ep.sonnet.n.good.failure.rbd\endcsname{256}
\expandafter\def\csname v@ep.sonnet.n.good.failure.rbdx\endcsname{256}
\expandafter\def\csname v@ep.sonnet.n.good.failure.rbx\endcsname{256}
\expandafter\def\csname v@ep.sonnet.n.good.success.n\endcsname{256}
\expandafter\def\csname v@ep.sonnet.n.good.success.neutral\endcsname{256}
\expandafter\def\csname v@ep.sonnet.n.good.success.nx\endcsname{256}
\expandafter\def\csname v@ep.sonnet.n.good.success.planted\endcsname{256}
\expandafter\def\csname v@ep.sonnet.n.good.success.rb\endcsname{256}
\expandafter\def\csname v@ep.sonnet.n.good.success.rbd\endcsname{256}
\expandafter\def\csname v@ep.sonnet.n.good.success.rbdx\endcsname{256}
\expandafter\def\csname v@ep.sonnet.n.good.success.rbx\endcsname{256}
\expandafter\def\csname v@ep.sonnet.name\endcsname{Claude Sonnet 5.5}
\expandafter\def\csname v@ep.sonnet.namesworse\endcsname{5\%}
\expandafter\def\csname v@ep.sonnet.p2rb.ci\endcsname{[$-0.029$, $+0.029$]}
\expandafter\def\csname v@ep.sonnet.p2rb.lower\endcsname{$0.000$}
\expandafter\def\csname v@ep.sonnet.p2rb.lowerci\endcsname{[$-0.029$, $+0.029$]}
\expandafter\def\csname v@ep.sonnet.p2rb.mean\endcsname{$0.000$}
\expandafter\def\csname v@ep.sonnet.p2rb.n\endcsname{256}
\expandafter\def\csname v@ep.sonnet.p2rb.pb\endcsname{$1.000$}
\expandafter\def\csname v@ep.sonnet.p2rb.pbraw\endcsname{$1.000$}
\expandafter\def\csname v@ep.sonnet.p2rb.praw\endcsname{$1.000$}
\expandafter\def\csname v@ep.sonnet.p2rb.pt\endcsname{$1.000$}
\expandafter\def\csname v@ep.sonnet.p2rb.upper\endcsname{$0.000$}
\expandafter\def\csname v@ep.sonnet.p2rb.upperci\endcsname{[$-0.029$, $+0.029$]}
\expandafter\def\csname v@ep.sonnet.p3mrb.ci\endcsname{[$-0.029$, $+0.029$]}
\expandafter\def\csname v@ep.sonnet.p3mrb.lower\endcsname{$0.000$}
\expandafter\def\csname v@ep.sonnet.p3mrb.lowerci\endcsname{[$-0.029$, $+0.029$]}
\expandafter\def\csname v@ep.sonnet.p3mrb.mean\endcsname{$0.000$}
\expandafter\def\csname v@ep.sonnet.p3mrb.n\endcsname{256}
\expandafter\def\csname v@ep.sonnet.p3mrb.pb\endcsname{$1.000$}
\expandafter\def\csname v@ep.sonnet.p3mrb.pbraw\endcsname{$1.000$}
\expandafter\def\csname v@ep.sonnet.p3mrb.praw\endcsname{$1.000$}
\expandafter\def\csname v@ep.sonnet.p3mrb.pt\endcsname{$1.000$}
\expandafter\def\csname v@ep.sonnet.p3mrb.upper\endcsname{$0.000$}
\expandafter\def\csname v@ep.sonnet.p3mrb.upperci\endcsname{[$-0.029$, $+0.029$]}
\expandafter\def\csname v@ep.sonnet.planted.k\endcsname{1}
\expandafter\def\csname v@ep.sonnet.planted.n\endcsname{1,024}
\expandafter\def\csname v@ep.sonnet.rate.bad.failure.n\endcsname{0.0\%}
\expandafter\def\csname v@ep.sonnet.rate.bad.failure.neutral\endcsname{0.0\%}
\expandafter\def\csname v@ep.sonnet.rate.bad.failure.nx\endcsname{0.0\%}
\expandafter\def\csname v@ep.sonnet.rate.bad.failure.planted\endcsname{0.0\%}
\expandafter\def\csname v@ep.sonnet.rate.bad.failure.rb\endcsname{0.0\%}
\expandafter\def\csname v@ep.sonnet.rate.bad.failure.rbd\endcsname{0.0\%}
\expandafter\def\csname v@ep.sonnet.rate.bad.failure.rbdx\endcsname{0.0\%}
\expandafter\def\csname v@ep.sonnet.rate.bad.failure.rbx\endcsname{0.0\%}
\expandafter\def\csname v@ep.sonnet.rate.bad.success.n\endcsname{0.0\%}
\expandafter\def\csname v@ep.sonnet.rate.bad.success.neutral\endcsname{0.0\%}
\expandafter\def\csname v@ep.sonnet.rate.bad.success.nx\endcsname{0.0\%}
\expandafter\def\csname v@ep.sonnet.rate.bad.success.planted\endcsname{0.0\%}
\expandafter\def\csname v@ep.sonnet.rate.bad.success.rb\endcsname{0.0\%}
\expandafter\def\csname v@ep.sonnet.rate.bad.success.rbd\endcsname{0.0\%}
\expandafter\def\csname v@ep.sonnet.rate.bad.success.rbdx\endcsname{0.0\%}
\expandafter\def\csname v@ep.sonnet.rate.bad.success.rbx\endcsname{0.0\%}
\expandafter\def\csname v@ep.sonnet.rate.good.failure.n\endcsname{0.0\%}
\expandafter\def\csname v@ep.sonnet.rate.good.failure.neutral\endcsname{0.0\%}
\expandafter\def\csname v@ep.sonnet.rate.good.failure.nx\endcsname{0.0\%}
\expandafter\def\csname v@ep.sonnet.rate.good.failure.planted\endcsname{0.4\%}
\expandafter\def\csname v@ep.sonnet.rate.good.failure.rb\endcsname{0.0\%}
\expandafter\def\csname v@ep.sonnet.rate.good.failure.rbd\endcsname{0.0\%}
\expandafter\def\csname v@ep.sonnet.rate.good.failure.rbdx\endcsname{0.0\%}
\expandafter\def\csname v@ep.sonnet.rate.good.failure.rbx\endcsname{0.0\%}
\expandafter\def\csname v@ep.sonnet.rate.good.success.n\endcsname{0.0\%}
\expandafter\def\csname v@ep.sonnet.rate.good.success.neutral\endcsname{0.0\%}
\expandafter\def\csname v@ep.sonnet.rate.good.success.nx\endcsname{0.0\%}
\expandafter\def\csname v@ep.sonnet.rate.good.success.planted\endcsname{0.0\%}
\expandafter\def\csname v@ep.sonnet.rate.good.success.rb\endcsname{0.0\%}
\expandafter\def\csname v@ep.sonnet.rate.good.success.rbd\endcsname{0.0\%}
\expandafter\def\csname v@ep.sonnet.rate.good.success.rbdx\endcsname{0.0\%}
\expandafter\def\csname v@ep.sonnet.rate.good.success.rbx\endcsname{0.0\%}
\expandafter\def\csname v@ep.sonnet.sus.ci\endcsname{[$-0.001$, $+0.003$]}
\expandafter\def\csname v@ep.sonnet.sus.mean\endcsname{$+0.001$}
\expandafter\def\csname v@ep.sonnet.sus.n\endcsname{256}
\expandafter\def\csname v@ep.sonnet.sus.pbraw\endcsname{$0.842$}
\expandafter\def\csname v@ep.sonnet.sus.praw\endcsname{$0.318$}
\expandafter\def\csname v@ep.sonnet.word.pthree\endcsname{\emph{bounded below 0.10}}
\expandafter\def\csname v@ep.sonnet.word.ptwo\endcsname{\emph{bounded below 0.10}}
\expandafter\def\csname v@ep.sonnet.word.sus\endcsname{\emph{increment below 0.10}}
\expandafter\def\csname v@ep.susbelowmax\endcsname{$+0.004$}
\expandafter\def\csname v@ep.top1.name\endcsname{Qwen 3.8 Max}
\expandafter\def\csname v@ep.top1.ptwo\endcsname{$+0.352$}
\expandafter\def\csname v@ep.top2.name\endcsname{DeepSeek V4 Pro}
\expandafter\def\csname v@ep.top2.ptwo\endcsname{$+0.277$}
\expandafter\def\csname v@ex.a.name\endcsname{Qwen 3.8 Max}
\expandafter\def\csname v@ex.a.world\endcsname{w20261009\_0004}
\expandafter\def\csname v@ex.b.name\endcsname{Grok 4.7}
\expandafter\def\csname v@ex.b.world\endcsname{w20261009\_0018}
\expandafter\def\csname v@ex.c.name\endcsname{Mistral Large 4}
\expandafter\def\csname v@ex.c.world\endcsname{w20261009\_0003}
\expandafter\def\csname v@ex.d.name\endcsname{Claude Opus 5.5}
\expandafter\def\csname v@ex.d.world\endcsname{w20261009\_0102}
\expandafter\def\csname v@flags.nchecked\endcsname{24}
\expandafter\def\csname v@lock.block\endcsname{970623}
\expandafter\def\csname v@lock.blocktime\endcsname{2026-10-09 11:33:13 UTC}
\expandafter\def\csname v@panel.a2n.ci\endcsname{[$+0.020$, $+0.034$]}
\expandafter\def\csname v@panel.a2n.mean\endcsname{$+0.027$}
\expandafter\def\csname v@panel.a2n.n\endcsname{256}
\expandafter\def\csname v@panel.a2n.pbraw\endcsname{$0.0078$}
\expandafter\def\csname v@panel.a2n.praw\endcsname{$9.2\times10^{-13}$}
\expandafter\def\csname v@panel.a2rb.ci\endcsname{[$+0.097$, $+0.127$]}
\expandafter\def\csname v@panel.a2rb.lower\endcsname{$+0.112$}
\expandafter\def\csname v@panel.a2rb.lowerci\endcsname{[$+0.097$, $+0.127$]}
\expandafter\def\csname v@panel.a2rb.mean\endcsname{$+0.112$}
\expandafter\def\csname v@panel.a2rb.n\endcsname{256}
\expandafter\def\csname v@panel.a2rb.pbraw\endcsname{$2.5\times10^{-9}$}
\expandafter\def\csname v@panel.a2rb.praw\endcsname{$6.8\times10^{-37}$}
\expandafter\def\csname v@panel.a2rb.upper\endcsname{$+0.113$}
\expandafter\def\csname v@panel.a2rb.upperci\endcsname{[$+0.098$, $+0.128$]}
\expandafter\def\csname v@panel.a2rbd.ci\endcsname{[$+0.060$, $+0.081$]}
\expandafter\def\csname v@panel.a2rbd.mean\endcsname{$+0.071$}
\expandafter\def\csname v@panel.a2rbd.n\endcsname{256}
\expandafter\def\csname v@panel.a2rbd.pbraw\endcsname{$2.9\times10^{-6}$}
\expandafter\def\csname v@panel.a2rbd.praw\endcsname{$1.7\times10^{-30}$}
\expandafter\def\csname v@panel.a3rb.ci\endcsname{[$-0.002$, $+0.004$]}
\expandafter\def\csname v@panel.a3rb.lower\endcsname{$+0.001$}
\expandafter\def\csname v@panel.a3rb.lowerci\endcsname{[$-0.002$, $+0.004$]}
\expandafter\def\csname v@panel.a3rb.mean\endcsname{$+0.001$}
\expandafter\def\csname v@panel.a3rb.n\endcsname{256}
\expandafter\def\csname v@panel.a3rb.pbraw\endcsname{$0.848$}
\expandafter\def\csname v@panel.a3rb.praw\endcsname{$0.468$}
\expandafter\def\csname v@panel.a3rb.upper\endcsname{$+0.001$}
\expandafter\def\csname v@panel.a3rb.upperci\endcsname{[$-0.002$, $+0.004$]}
\expandafter\def\csname v@panel.ceil.bad\endcsname{$0.003$}
\expandafter\def\csname v@panel.ceil.good\endcsname{$0.004$}
\expandafter\def\csname v@panel.l2brb.ci\endcsname{[$-0.001$, $+0.003$]}
\expandafter\def\csname v@panel.l2brb.mean\endcsname{$+0.001$}
\expandafter\def\csname v@panel.l2brb.n\endcsname{256}
\expandafter\def\csname v@panel.l2brb.pbraw\endcsname{$0.828$}
\expandafter\def\csname v@panel.l2brb.praw\endcsname{$0.258$}
\expandafter\def\csname v@panel.l2rb.ci\endcsname{[$+0.098$, $+0.126$]}
\expandafter\def\csname v@panel.l2rb.lower\endcsname{$+0.112$}
\expandafter\def\csname v@panel.l2rb.lowerci\endcsname{[$+0.098$, $+0.126$]}
\expandafter\def\csname v@panel.l2rb.mean\endcsname{$+0.112$}
\expandafter\def\csname v@panel.l2rb.n\endcsname{256}
\expandafter\def\csname v@panel.l2rb.pbraw\endcsname{$2.4\times10^{-9}$}
\expandafter\def\csname v@panel.l2rb.praw\endcsname{$3.0\times10^{-38}$}
\expandafter\def\csname v@panel.l2rb.upper\endcsname{$+0.113$}
\expandafter\def\csname v@panel.l2rb.upperci\endcsname{[$+0.099$, $+0.127$]}
\expandafter\def\csname v@panel.l2rbexplicit.ci\endcsname{[$+0.032$, $+0.053$]}
\expandafter\def\csname v@panel.l2rbexplicit.mean\endcsname{$+0.043$}
\expandafter\def\csname v@panel.l2rbexplicit.n\endcsname{128}
\expandafter\def\csname v@panel.l2rbexplicit.praw\endcsname{$6.9\times10^{-13}$}
\expandafter\def\csname v@panel.l2rbimplicit.ci\endcsname{[$+0.161$, $+0.202$]}
\expandafter\def\csname v@panel.l2rbimplicit.mean\endcsname{$+0.181$}
\expandafter\def\csname v@panel.l2rbimplicit.n\endcsname{128}
\expandafter\def\csname v@panel.l2rbimplicit.praw\endcsname{$1.9\times10^{-35}$}
\expandafter\def\csname v@panel.l3brb.ci\endcsname{[$+0.002$, $+0.006$]}
\expandafter\def\csname v@panel.l3brb.mean\endcsname{$+0.004$}
\expandafter\def\csname v@panel.l3brb.n\endcsname{256}
\expandafter\def\csname v@panel.l3brb.pbraw\endcsname{$0.484$}
\expandafter\def\csname v@panel.l3brb.praw\endcsname{$8.3\times10^{-4}$}
\expandafter\def\csname v@panel.l3rb.ci\endcsname{[$+0.002$, $+0.006$]}
\expandafter\def\csname v@panel.l3rb.lower\endcsname{$+0.004$}
\expandafter\def\csname v@panel.l3rb.lowerci\endcsname{[$+0.001$, $+0.006$]}
\expandafter\def\csname v@panel.l3rb.mean\endcsname{$+0.004$}
\expandafter\def\csname v@panel.l3rb.n\endcsname{256}
\expandafter\def\csname v@panel.l3rb.pbraw\endcsname{$0.484$}
\expandafter\def\csname v@panel.l3rb.praw\endcsname{$8.3\times10^{-4}$}
\expandafter\def\csname v@panel.l3rb.upper\endcsname{$+0.004$}
\expandafter\def\csname v@panel.l3rb.upperci\endcsname{[$+0.002$, $+0.007$]}
\expandafter\def\csname v@panel.p2n.ci\endcsname{[$+0.020$, $+0.034$]}
\expandafter\def\csname v@panel.p2n.mean\endcsname{$+0.027$}
\expandafter\def\csname v@panel.p2n.n\endcsname{256}
\expandafter\def\csname v@panel.p2n.pbet\endcsname{$0.080$}
\expandafter\def\csname v@panel.p2n.pbraw\endcsname{$0.080$}
\expandafter\def\csname v@panel.p2n.praw\endcsname{$3.2\times10^{-13}$}
\expandafter\def\csname v@panel.p2rb.ci\endcsname{[$+0.097$, $+0.126$]}
\expandafter\def\csname v@panel.p2rb.lower\endcsname{$+0.112$}
\expandafter\def\csname v@panel.p2rb.lowerci\endcsname{[$+0.097$, $+0.127$]}
\expandafter\def\csname v@panel.p2rb.mean\endcsname{$+0.112$}
\expandafter\def\csname v@panel.p2rb.n\endcsname{256}
\expandafter\def\csname v@panel.p2rb.pb\endcsname{$7.0\times10^{-5}$}
\expandafter\def\csname v@panel.p2rb.pbraw\endcsname{$3.5\times10^{-5}$}
\expandafter\def\csname v@panel.p2rb.praw\endcsname{$8.8\times10^{-37}$}
\expandafter\def\csname v@panel.p2rb.pt\endcsname{$1.8\times10^{-36}$}
\expandafter\def\csname v@panel.p2rb.upper\endcsname{$+0.113$}
\expandafter\def\csname v@panel.p2rb.upperci\endcsname{[$+0.099$, $+0.128$]}
\expandafter\def\csname v@panel.p2rbd.ci\endcsname{[$+0.061$, $+0.082$]}
\expandafter\def\csname v@panel.p2rbd.mean\endcsname{$+0.072$}
\expandafter\def\csname v@panel.p2rbd.n\endcsname{256}
\expandafter\def\csname v@panel.p2rbd.pbet\endcsname{$0.0013$}
\expandafter\def\csname v@panel.p2rbd.pbraw\endcsname{$0.0013$}
\expandafter\def\csname v@panel.p2rbd.praw\endcsname{$7.1\times10^{-31}$}
\expandafter\def\csname v@panel.p2rbexplicit.ci\endcsname{[$+0.031$, $+0.053$]}
\expandafter\def\csname v@panel.p2rbexplicit.mean\endcsname{$+0.042$}
\expandafter\def\csname v@panel.p2rbexplicit.n\endcsname{128}
\expandafter\def\csname v@panel.p2rbexplicit.praw\endcsname{$2.9\times10^{-12}$}
\expandafter\def\csname v@panel.p2rbimplicit.ci\endcsname{[$+0.160$, $+0.203$]}
\expandafter\def\csname v@panel.p2rbimplicit.mean\endcsname{$+0.181$}
\expandafter\def\csname v@panel.p2rbimplicit.n\endcsname{128}
\expandafter\def\csname v@panel.p2rbimplicit.praw\endcsname{$6.3\times10^{-34}$}
\expandafter\def\csname v@panel.p3mn.ci\endcsname{[$-0.003$, $+0.001$]}
\expandafter\def\csname v@panel.p3mn.mean\endcsname{$-0.001$}
\expandafter\def\csname v@panel.p3mn.n\endcsname{256}
\expandafter\def\csname v@panel.p3mn.pbraw\endcsname{$1.000$}
\expandafter\def\csname v@panel.p3mn.praw\endcsname{$0.367$}
\expandafter\def\csname v@panel.p3mrb.ci\endcsname{[$-0.001$, $+0.005$]}
\expandafter\def\csname v@panel.p3mrb.lower\endcsname{$+0.001$}
\expandafter\def\csname v@panel.p3mrb.lowerci\endcsname{[$-0.002$, $+0.005$]}
\expandafter\def\csname v@panel.p3mrb.mean\endcsname{$+0.002$}
\expandafter\def\csname v@panel.p3mrb.n\endcsname{256}
\expandafter\def\csname v@panel.p3mrb.pb\endcsname{$0.853$}
\expandafter\def\csname v@panel.p3mrb.pbraw\endcsname{$0.853$}
\expandafter\def\csname v@panel.p3mrb.praw\endcsname{$0.252$}
\expandafter\def\csname v@panel.p3mrb.pt\endcsname{$0.252$}
\expandafter\def\csname v@panel.p3mrb.upper\endcsname{$+0.002$}
\expandafter\def\csname v@panel.p3mrb.upperci\endcsname{[$-0.001$, $+0.006$]}
\expandafter\def\csname v@panel.p3mrbd.ci\endcsname{[$-0.005$, $0.000$]}
\expandafter\def\csname v@panel.p3mrbd.mean\endcsname{$-0.002$}
\expandafter\def\csname v@panel.p3mrbd.n\endcsname{256}
\expandafter\def\csname v@panel.p3mrbd.pbraw\endcsname{$1.000$}
\expandafter\def\csname v@panel.p3mrbd.praw\endcsname{$0.052$}
\expandafter\def\csname v@panel.p3mrbexplicit.ci\endcsname{[$-0.002$, $+0.008$]}
\expandafter\def\csname v@panel.p3mrbexplicit.mean\endcsname{$+0.003$}
\expandafter\def\csname v@panel.p3mrbexplicit.n\endcsname{128}
\expandafter\def\csname v@panel.p3mrbexplicit.praw\endcsname{$0.287$}
\expandafter\def\csname v@panel.p3mrbimplicit.ci\endcsname{[$-0.002$, $+0.004$]}
\expandafter\def\csname v@panel.p3mrbimplicit.mean\endcsname{$+0.001$}
\expandafter\def\csname v@panel.p3mrbimplicit.n\endcsname{128}
\expandafter\def\csname v@panel.p3mrbimplicit.praw\endcsname{$0.656$}
\expandafter\def\csname v@panel.remedya2nminusrb.ci\endcsname{[$-0.098$, $-0.074$]}
\expandafter\def\csname v@panel.remedya2nminusrb.mean\endcsname{$-0.086$}
\expandafter\def\csname v@panel.remedya2nminusrb.n\endcsname{256}
\expandafter\def\csname v@panel.remedya2nminusrb.praw\endcsname{$1.1\times10^{-33}$}
\expandafter\def\csname v@panel.remedya2rbdminusrb.ci\endcsname{[$-0.054$, $-0.027$]}
\expandafter\def\csname v@panel.remedya2rbdminusrb.mean\endcsname{$-0.041$}
\expandafter\def\csname v@panel.remedya2rbdminusrb.n\endcsname{256}
\expandafter\def\csname v@panel.remedya2rbdminusrb.praw\endcsname{$1.0\times10^{-8}$}
\expandafter\def\csname v@panel.remedya3nminusrb.ci\endcsname{[$-0.005$, $0.000$]}
\expandafter\def\csname v@panel.remedya3nminusrb.mean\endcsname{$-0.003$}
\expandafter\def\csname v@panel.remedya3nminusrb.n\endcsname{256}
\expandafter\def\csname v@panel.remedya3nminusrb.praw\endcsname{$0.032$}
\expandafter\def\csname v@panel.remedya3rbdminusrb.ci\endcsname{[$-0.006$, $-0.001$]}
\expandafter\def\csname v@panel.remedya3rbdminusrb.mean\endcsname{$-0.004$}
\expandafter\def\csname v@panel.remedya3rbdminusrb.n\endcsname{256}
\expandafter\def\csname v@panel.remedya3rbdminusrb.praw\endcsname{$0.0037$}
\expandafter\def\csname v@panel.remedyl2nminusrb.ci\endcsname{[$-0.098$, $-0.073$]}
\expandafter\def\csname v@panel.remedyl2nminusrb.mean\endcsname{$-0.086$}
\expandafter\def\csname v@panel.remedyl2nminusrb.n\endcsname{256}
\expandafter\def\csname v@panel.remedyl2nminusrb.praw\endcsname{$8.1\times10^{-32}$}
\expandafter\def\csname v@panel.remedyl2rbdminusrb.ci\endcsname{[$-0.056$, $-0.029$]}
\expandafter\def\csname v@panel.remedyl2rbdminusrb.mean\endcsname{$-0.043$}
\expandafter\def\csname v@panel.remedyl2rbdminusrb.n\endcsname{256}
\expandafter\def\csname v@panel.remedyl2rbdminusrb.praw\endcsname{$2.3\times10^{-9}$}
\expandafter\def\csname v@panel.remedyp2nminusrb.ci\endcsname{[$-0.097$, $-0.073$]}
\expandafter\def\csname v@panel.remedyp2nminusrb.mean\endcsname{$-0.085$}
\expandafter\def\csname v@panel.remedyp2nminusrb.n\endcsname{256}
\expandafter\def\csname v@panel.remedyp2nminusrb.praw\endcsname{$7.1\times10^{-33}$}
\expandafter\def\csname v@panel.remedyp2rbdminusrb.ci\endcsname{[$-0.053$, $-0.026$]}
\expandafter\def\csname v@panel.remedyp2rbdminusrb.mean\endcsname{$-0.039$}
\expandafter\def\csname v@panel.remedyp2rbdminusrb.n\endcsname{256}
\expandafter\def\csname v@panel.remedyp2rbdminusrb.praw\endcsname{$2.6\times10^{-8}$}
\expandafter\def\csname v@panel.remedyp3mnminusrb.ci\endcsname{[$-0.005$, $0.000$]}
\expandafter\def\csname v@panel.remedyp3mnminusrb.mean\endcsname{$-0.003$}
\expandafter\def\csname v@panel.remedyp3mnminusrb.n\endcsname{256}
\expandafter\def\csname v@panel.remedyp3mnminusrb.praw\endcsname{$0.032$}
\expandafter\def\csname v@panel.remedyp3mrbdminusrb.ci\endcsname{[$-0.007$, $-0.002$]}
\expandafter\def\csname v@panel.remedyp3mrbdminusrb.mean\endcsname{$-0.004$}
\expandafter\def\csname v@panel.remedyp3mrbdminusrb.n\endcsname{256}
\expandafter\def\csname v@panel.remedyp3mrbdminusrb.praw\endcsname{$0.0012$}
\expandafter\def\csname v@panel.sus.ci\endcsname{[$+0.130$, $+0.143$]}
\expandafter\def\csname v@panel.sus.mean\endcsname{$+0.137$}
\expandafter\def\csname v@panel.sus.n\endcsname{256}
\expandafter\def\csname v@panel.sus.pbraw\endcsname{$1.7\times10^{-11}$}
\expandafter\def\csname v@panel.sus.praw\endcsname{$3.0\times10^{-120}$}
\expandafter\def\csname v@panel.t1rb.ci\endcsname{[$+0.101$, $+0.130$]}
\expandafter\def\csname v@panel.t1rb.mean\endcsname{$+0.115$}
\expandafter\def\csname v@panel.t1rb.n\endcsname{256}
\expandafter\def\csname v@panel.t1rb.pbraw\endcsname{$1.4\times10^{-9}$}
\expandafter\def\csname v@panel.t1rb.praw\endcsname{$3.2\times10^{-39}$}
\expandafter\def\csname v@panel.t3rb.ci\endcsname{[$+0.002$, $+0.006$]}
\expandafter\def\csname v@panel.t3rb.mean\endcsname{$+0.004$}
\expandafter\def\csname v@panel.t3rb.n\endcsname{256}
\expandafter\def\csname v@panel.t3rb.pbraw\endcsname{$0.484$}
\expandafter\def\csname v@panel.t3rb.praw\endcsname{$8.3\times10^{-4}$}
\expandafter\def\csname v@panel.word.pthree\endcsname{\emph{bounded below 0.10}}
\expandafter\def\csname v@panel.word.ptwo\endcsname{\emph{luck-carrying}}
\expandafter\def\csname v@panel.word.sus\endcsname{\emph{increment above 0.10}}
\expandafter\def\csname v@rate.bad.failure.n\endcsname{0.0\%}
\expandafter\def\csname v@rate.bad.failure.neutral\endcsname{0.1\%}
\expandafter\def\csname v@rate.bad.failure.nx\endcsname{0.1\%}
\expandafter\def\csname v@rate.bad.failure.planted\endcsname{11.2\%}
\expandafter\def\csname v@rate.bad.failure.rb\endcsname{0.0\%}
\expandafter\def\csname v@rate.bad.failure.rbd\endcsname{0.1\%}
\expandafter\def\csname v@rate.bad.failure.rbdx\endcsname{0.0\%}
\expandafter\def\csname v@rate.bad.failure.rbx\endcsname{0.4\%}
\expandafter\def\csname v@rate.bad.success.n\endcsname{0.1\%}
\expandafter\def\csname v@rate.bad.success.neutral\endcsname{0.3\%}
\expandafter\def\csname v@rate.bad.success.nx\endcsname{0.0\%}
\expandafter\def\csname v@rate.bad.success.planted\endcsname{12.0\%}
\expandafter\def\csname v@rate.bad.success.rb\endcsname{0.4\%}
\expandafter\def\csname v@rate.bad.success.rbd\endcsname{0.0\%}
\expandafter\def\csname v@rate.bad.success.rbdx\endcsname{0.0\%}
\expandafter\def\csname v@rate.bad.success.rbx\endcsname{0.0\%}
\expandafter\def\csname v@rate.good.failure.n\endcsname{3.1\%}
\expandafter\def\csname v@rate.good.failure.neutral\endcsname{0.4\%}
\expandafter\def\csname v@rate.good.failure.nx\endcsname{0.4\%}
\expandafter\def\csname v@rate.good.failure.planted\endcsname{17.6\%}
\expandafter\def\csname v@rate.good.failure.rb\endcsname{11.6\%}
\expandafter\def\csname v@rate.good.failure.rbd\endcsname{7.5\%}
\expandafter\def\csname v@rate.good.failure.rbdx\endcsname{0.6\%}
\expandafter\def\csname v@rate.good.failure.rbx\endcsname{0.4\%}
\expandafter\def\csname v@rate.good.success.n\endcsname{0.0\%}
\expandafter\def\csname v@rate.good.success.neutral\endcsname{0.1\%}
\expandafter\def\csname v@rate.good.success.nx\endcsname{0.1\%}
\expandafter\def\csname v@rate.good.success.planted\endcsname{15.0\%}
\expandafter\def\csname v@rate.good.success.rb\endcsname{0.1\%}
\expandafter\def\csname v@rate.good.success.rbd\endcsname{0.0\%}
\expandafter\def\csname v@rate.good.success.rbdx\endcsname{0.1\%}
\expandafter\def\csname v@rate.good.success.rbx\endcsname{0.2\%}
\expandafter\def\csname v@rb.commit\endcsname{\texttt{\seqsplit{ed80611788292ea739f1effd31f16c53823b8a0d}}}
\expandafter\def\csname v@rb.fetched\endcsname{2026-10-08}
\expandafter\def\csname v@rb.filesha\endcsname{\texttt{\seqsplit{073d354bc84a24ba8a2c697133b5d832fe89ca67a53705fb0af419528f7b11ee}}}
\expandafter\def\csname v@rb.path\endcsname{\nolinkurl{WebArena/prompts/memory_instruction.py}}
\expandafter\def\csname v@records.doi\endcsname{10.5281/zenodo.23272809}
\expandafter\def\csname v@reimpl.ambiguities\endcsname{15}
\expandafter\def\csname v@reimpl.maxdiff\endcsname{$1.5\times10^{-15}$}
\expandafter\def\csname v@reimpl.numbers\endcsname{3,480}
\expandafter\def\csname v@reimpl.wordmismatch\endcsname{0}
\expandafter\def\csname v@reimpl.words\endcsname{36}
\expandafter\def\csname v@run.answers\endcsname{90,084}
\expandafter\def\csname v@run.lessonsok\endcsname{33,789}
\expandafter\def\csname v@run.missinglessons\endcsname{3}
\expandafter\def\csname v@run.missinglessons.cells\endcsname{Claude Opus 5.5 (RB), Gemini 3.8 Flash (RBD) and Qwen 3.8 Max (RBD)}
\expandafter\def\csname v@run.parseerrors\endcsname{22}
\expandafter\def\csname v@run.readerrequests\endcsname{90,106}
\expandafter\def\csname v@run.recovered\endcsname{0}
\expandafter\def\csname v@run.smokespent\endcsname{4.36}
\expandafter\def\csname v@run.spent\endcsname{543.41}
\expandafter\def\csname v@run.spentreader\endcsname{298.21}
\expandafter\def\csname v@run.spentwriter\endcsname{245.20}
\expandafter\def\csname v@run.unscheduled\endcsname{6}
\expandafter\def\csname v@run.writerempty\endcsname{2}
\expandafter\def\csname v@run.writerrefusal\endcsname{1}
\expandafter\def\csname v@run.writerrequests\endcsname{33,792}

\title{Lessons from Luck: Do LLM-Written Lessons Carry Chance\\into an Agent's Next Decision?}
\author{Kazuki Nakayashiki\\ \small Glasp}
\date{}

\begin{document}
\maketitle

\begin{abstract}
Self-improving agents write lessons from the outcomes of their runs; when outcomes are noisy, a lesson can carry chance into later
decisions. We measure one link of this chain in a registered study. In \V{design.worlds} synthetic worlds, each a recurring task with
two interchangeable tools of recorded reliability, a scripted previous run used the more reliable tool (a good decision) or the less
reliable one (a bad decision), with an assigned outcome: success or a transient failure. A model wrote a lesson from the run with
ReasoningBank's memory-writing instruction; a fresh session of the same model, holding the authoritative log and the previous result,
chose the next run's first tool with one memory item: that lesson, the lesson from the other outcome, an irrelevant item, or an
unreasoned directive. The protocol was frozen, deposited and timestamped before any confirmatory call; \V{design.endpoints} LLM
endpoints gave \V{run.answers} answers. After a chance failure of the better tool, the panel's lessons carried the chance (registered
reading \V{panel.word.ptwo}; interaction \V{panel.p2rb.mean}, 95\% interval \V{panel.p2rb.ci}): readers named the worse
tool in \V{rate.good.failure.rb} of next runs with the lesson written from that failure, against \V{rate.good.failure.neutral} with an
irrelevant item and \V{rate.good.failure.rbx} with the lesson written from a success. The effect was \V{panel.p2rbexplicit.mean} where
the log stated that failures are independent and \V{panel.p2rbimplicit.mean} where it did not (descriptive). After a
lucky success of the worse tool, the interaction was \V{panel.word.pthree} (\V{panel.p3mrb.mean}), a bound, not evidence of no
harm. In exploratory readings, \V{ep.nluck} endpoints carried the chance (up to \V{ep.top1.ptwo}); \V{ep.nbounded} were bounded below \lit{0.10}. In
secondary comparisons, a three-sentence decision-quality note in the writing instruction lowered the worse-tool rate after a chance
failure to \V{rate.good.failure.rbd}, and one instruction for both outcomes with a no-change option to \V{rate.good.failure.n}; neither
met the registered harm-reduction rule at every such endpoint.

\end{abstract}

\section{Introduction}\label{sec:intro}

Agents built on large language models (LLMs) increasingly learn from their own runs by writing lessons into memory and reading them
before later runs. Reflexion keeps verbal self-reflections after failed trials \citep{shinn2023reflexion}; ExpeL and Agent Workflow
Memory extract insights or reusable workflows from past trajectories \citep{zhao2023expel,wang2024agent}; Dynamic Cheatsheet maintains
an evolving memory of strategies at test time \citep{suzgun2025dynamic}; and ReasoningBank distils reasoning strategies from both
successful and failed trajectories, sending each trajectory to a success or a failure instruction according to its outcome
\citep{ouyang2025reasoningbank}.

Outcomes, however, are noisy. A sound decision can fail by chance and an unsound one can succeed by chance. Judging a decision by its
outcome rather than by the information available when it was made is the outcome bias of decision research
\citep{baron1988outcome}, and knowing an outcome makes it seem more predictable than it was \citep{fischhoff1975hindsight}. A
memory-writing step that turns one outcome into a durable rule could therefore carry chance forward: a lesson written after a chance
failure of a sound decision may tell later runs to abandon it. No adversary is needed for this error. Recent audits of agent memory and
self-improvement give reason to look: memory gains vary across repeated runs and task orders \citep{ye2026on}, self-assigned rewards in
memory banks inflate \citep{asadolahi2026memory}, and harness proposers fix failures that never happened when, among other conditions,
their instructions presuppose failures \citep{wang2026phantom}. Whether a chance outcome, with decision quality held fixed, is written
into a lesson that moves a later decision was not, as far as we found, measured (Section~\ref{sec:related}).

We measure one link of this chain. With the decision, the information available when it was made and the task held fixed, does a
lesson written from a chance outcome change the tool a fresh session of the same model would choose next, beyond what the raw outcome
alone does? We ask it in a setting favourable to the reader: it holds the authoritative run log and is told the previous run's result,
so it could check the lesson against the evidence. In \V{design.worlds} synthetic worlds, each a recurring task with two interchangeable
tools of recorded reliability, a scripted previous run used the more reliable tool (a good decision) or the less reliable one (a bad
decision), and its outcome was assigned. A model wrote a lesson from that run with ReasoningBank's outcome-routed writing instruction or
one of two variants; a fresh session of the same model then named the tool it would call first in the next run, with one memory item:
that lesson, the lesson written from the other outcome, an irrelevant item, or an unreasoned directive (Section~\ref{sec:design}). The
primary contrasts are differences-in-differences between readers that hold the same information and differ only in the memory item
(Section~\ref{sec:analysis}). The protocol, code and stimuli were frozen, deposited on Zenodo and timestamped in a Bitcoin block before
any confirmatory call; the run was launched once on \V{design.endpoints} LLM endpoints and analysed once, on the locked records, by the
registered analysis.

The findings (Section~\ref{sec:results}):
\begin{itemize}
\item After a chance failure of the better tool, the panel's lessons carried the chance (registered reading \V{panel.word.ptwo};
  \V{panel.p2rb.mean} \V{panel.p2rb.ci}): with the lesson written from that failure in memory, readers named the worse tool in
  \V{rate.good.failure.rb} of next runs, against \V{rate.good.failure.neutral} with an irrelevant item and \V{rate.good.failure.rbx}
  with the lesson written from a success, every reader seeing the same log. The lesson moved readers in combination with the reported
  failure: read after a reported success of the better tool, the same lesson led to the worse tool in \V{rate.good.success.rbx} of
  next runs.
\item Most of this effect arose where the record left the failure open to interpretation (a registered descriptive split): the
  interaction was \V{panel.p2rbimplicit.mean} \V{panel.p2rbimplicit.ci} in the worlds whose record did not state that failures are
  independent, and \V{panel.p2rbexplicit.mean} \V{panel.p2rbexplicit.ci} in the worlds whose record stated it. Only under the stated
  model is a worse-tool choice a clear error; where independence is not stated, a recent failure can be read as a sign of a current
  fault, and the choice is a deviation from the log-based benchmark. In both, the lesson added no observation beyond the reader's record.
\item After a lucky success of the worse tool, the outcome-by-memory interaction was \V{panel.word.pthree}
  (\V{panel.p3mrb.mean} \V{panel.p3mrb.ci}); this bounds the interaction and does not show that no harm occurred.
\item In exploratory endpoint readings, \V{ep.list.luck} carried the chance, and \V{ep.list.bounded} were bounded below \lit{0.10};
  a bound is not a zero: at \V{ep.list.boundedpos} the estimates were small but positive by their $t$ intervals.
\item In secondary comparisons, a three-sentence decision-quality note in the writing instruction lowered the panel's worse-tool rate
  after a chance failure to \V{rate.good.failure.rbd}, and one instruction for both outcomes with a no-change option to
  \V{rate.good.failure.n}; under either, the interaction stayed above zero by its $t$ interval, and neither met the registered
  harm-reduction rule at every endpoint.
\item At two endpoints, low response to the unreasoned directive coexisted with luck-carrying lessons (exploratory).
\end{itemize}

The study isolates one stored lesson, read once, in synthetic worlds with a constrained choice. It does not execute self-improvement,
retrieval or accumulated memory, does not estimate how often real agents meet such situations, tests the remedies only with outcomes
that did not signal a change in a tool's reliability, and claims no mechanism (Section~\ref{sec:limitations}). Appendix~\ref{app:examples} prints example lessons
and the answers they led to.

\section{Related work}\label{sec:related}

\paragraph{Self-written experience memory.} Agents that store lessons from past runs and read them later include Reflexion's verbal
reflections on task feedback \citep{shinn2023reflexion}, ExpeL's insights extracted from successful and failed trajectories
\citep{zhao2023expel}, Agent Workflow Memory's reusable workflows \citep{wang2024agent}, Dynamic Cheatsheet's evolving test-time memory
\citep{suzgun2025dynamic}, Agentic Context Engineering's evolving contexts \citep{zhang2025agentic}, and ReasoningBank, whose
memory-writing instructions this study adapts \citep{ouyang2025reasoningbank}. Earlier designs reflect over a stream of stored
experiences \citep{park2023generative} or keep a library of learned skills \citep{wang2023voyager}.

\paragraph{When self-written memory misleads.} Several recent studies document how such loops fail. Repeated runs and task-order
shuffles of two memory-based self-improving methods show high variance and order dependence, and in their full text evaluator errors
produce spurious memories \citep{ye2026on}. Self-assigned rewards in label-free memory banks inflate wrong episodes, which agents then
reuse \citep{asadolahi2026memory}. Reflections can repeat themselves and reinforce false beliefs \citep{dixit2026honest}; an agent that
writes its own conclusions into a store it later reads can be captured by the errors it wrote \citep{chen2026selfcleaning}; agents
follow retrieved experiences closely, so errors in memory propagate \citep{xiong2025how}; and harness proposers add guardrails against
failures that never happened when, among other conditions, their instruction presupposes failures \citep{wang2026phantom}. In harness
evolution, a revision can be promoted on a lucky rollout \citep{wei2026scale}; Zhu et al.\ argue that single-agent experience
construction lacks a reference distribution to separate stochastic failures from grounding errors \citep{zhu2026escaping}; and
verifier signals attached to memory records have been proposed to govern which trajectories, including accidental successes, are
stored and reused \citep{wang2026memguard}. These studies observe or argue that errors, noise and luck can enter self-written memory.

\paragraph{Outcome, hindsight and attribution.} People judge decisions by their outcomes \citep{baron1988outcome}, and knowing an
outcome makes it seem predictable \citep{fischhoff1975hindsight}. In a replication of a study of hindsight effects on judgements of legal
responsibility, LLMs showed effects much larger than people's \citep{almeida2023exploring}; LLMs also show hindsight-sensitive shifts
when shown complete clinical timelines \citep{matsuura2026hindsight}, asymmetries between attributing failures as actor and as observer
\citep{li2026taming}, and overweighting of advice that contradicts their own earlier answer \citep{kumaran2025how}. In repeated choice,
they weight positive and negative outcomes asymmetrically \citep{schubert2024incontext,wang2026comparative} and react strongly to the
most recent outcomes \citep{horowitz2025llm}. These are judgements or choices within one context. Here a model writes a persistent
lesson from a run whose outcome was assigned by chance, and a later session of the same model, which holds the evidence, reads it.

\paragraph{Memory poisoning.} Adversaries can insert records or instructions into an agent's memory
\citep{chen2024agentpoison,dong2025memory}, or craft locally correct experiences that reflective agents consolidate into
over-generalised rules \citep{wang2026oep}. The planted item of this study, an unreasoned directive to use the worse tool, is a positive
control of that kind; the lessons studied are written by a model without an adversary.

\paragraph{Position.} Prior work thus identifies misleading experience, including accidentally successful trajectories: evaluator
errors and run-to-run variance produce spurious memories \citep{ye2026on}, locally correct experiences crafted by an adversary are
consolidated into over-generalised rules \citep{wang2026oep}, and verifier signals have been proposed to govern which trajectories,
including accidental successes, are stored \citep{wang2026memguard}. The contribution here is therefore not the observation that
self-written memory can mislead but a decomposition of one way it does: an outcome assigned at random to a scripted decision of known
quality, a lesson written from it by a model, and readers of the same model that hold the same evidence and differ only in the memory
item. A targeted search made before the study was designed (October \lit{2026}; arXiv records and full texts where available; not
systematic) did not identify this design. This study belongs to a line of work on how agents treat inherited state before acting
\citep{nakayashiki2026checking,nakayashiki2026plan,nakayashiki2026when}.

\section{Design}\label{sec:design}

\paragraph{At a glance.} Each world supplies four scripted previous runs: a good or a bad decision, crossed with a success or a failure.
For each endpoint (a model served by one provider), three writing instructions turn each run into a lesson, twelve lessons per world.
The same endpoint then answers as a reader in the four previous-run conditions with each of eight memory arms: \V{design.readersperworld}
reader requests per world and endpoint. The reader names the tool it would call first; nothing is executed. The primary contrasts
(Section~\ref{sec:analysis}) are named P2 for the good decision and P3m for its mirror, the bad decision.

\subsection{Worlds}
Each of the \V{design.worlds} worlds is a recurring operational task (exporting data, sending invoices, rebuilding a search index and
the like) with two interchangeable tools that differ only in their recorded reliability. There are \lit{16} domains with \lit{16}
worlds each; within a domain, the better tool's listing position (first or second) and the cue (below) are balanced. The run log shows
the last \lit{180}, \lit{240}, \lit{300} or \lit{365} runs of each tool, with success rates of \lit{0.80}--\lit{0.93} for the better
tool and \lit{0.50}--\lit{0.70} for the worse one (a gap of at least \lit{0.15}); the two tools cost the same, and a failed run has to
be re-run by hand. Tool names are coined, checked against a word list and against the domain's texts, and the two names of a world
start with different letters. The generator accepts a world only if the better tool's posterior mean success rate after one more
failure exceeds the worse tool's after one more success by at least \lit{0.10}, so the better tool stays better after any single
further outcome.

The \emph{cue} is a world-level factor. In explicit worlds the record adds the sentence ``There have been no outages: failures of
either tool are independent one-off events, and the next run has always started normally.'' In implicit worlds the record shows only
the counts and does not state whether failures are independent. Half of the worlds (\V{design.worldshalf}) are explicit.

\subsection{Scripted runs and lesson writers}\label{sec:runs}
A previous run is a scripted trajectory in ReasoningBank's extraction format \citep{ouyang2025reasoningbank}: the task, then three
reasoning-and-action steps in which an agent reads the run log, states the record and its choice, calls the tool and reports the
returned text and that the run ended. In a good decision the agent calls the better tool (``the more reliable tool''); in a bad
decision it calls the worse tool and gives no reason. The outcome is a success or a transient, non-diagnostic error drawn from the
domain's pool, with ``this run allows no retry''. Nothing else differs between a world's four runs.

Three instructions write a lesson from a run (Appendix~\ref{app:writers}):
\begin{itemize}
\item \textbf{RB} (deployed): ReasoningBank's success instruction after a success and its failure instruction after a failure, as
  released in the authors' code (repository, file and commit in Appendix~\ref{app:writers}), verbatim except two declared
  replacements (``an expert in web navigation'' became ``an expert in running operational tasks'', and the rule against naming
  entities from the task allows naming the tools). The released instructions differ in three places from those printed in the
  ReasoningBank paper; among them, the released failure instruction asks for ``concrete, actionable recovery procedures''
  (Appendix~\ref{app:writers}). The results concern the released instructions.
\item \textbf{RBD} (deployed plus a remedy): RB with one added bullet of three sentences: ``Before writing, judge whether the agent's
  decision was sound given only what was known before the outcome. A single outcome is not by itself evidence that the decision was
  right or wrong. Recommend a different action only if the decision was unsound given that information.''
\item \textbf{N}: one instruction used after both outcomes (RB's success instruction with its four outcome-specific phrases made
  neutral) with the added sentence ``If nothing about how the task should be done needs to change, say so.'' The writer still reads
  the run, including its outcome; the registration calls this instruction ``outcome-blind'' because the instruction does not depend on
  the outcome. N differs from RB in two ways (no routing by outcome and an explicit no-change option), so N against RB is a compound
  comparison.
\end{itemize}
The stored memory is the writer's whole response, trimmed, inserted as one block, as the registration specifies; a response can hold
up to three formatted memory items. ReasoningBank distils an experience into memory items with a title, a description and content and
reuses all items of a retrieved experience \citep{ouyang2025reasoningbank}; the results concern its extraction instructions within this
representation. No sampling parameter is sent (ReasoningBank's released code extracts at temperature \lit{1.0}; here every endpoint
samples at its provider's default). Each endpoint writes the lessons that its own readers read, so differences between endpoints
cannot be assigned separately to writing or to reading. Each cell has one generation: one lesson per instruction, decision and outcome,
and one answer per cell and arm.

\subsection{Reader and memory arms}
The reader is a fresh session of the same model, instructed as an operations agent that reads saved memory items before each run. It
sees the task; both tools with one-line descriptions (``both operational today''); the run log (``authoritative; stable for months''),
with the cue sentence in explicit worlds; the equal costs and the hand re-run; the previous run (``\textit{tool} was used and it
succeeded'', or ``failed'' with the returned text, followed by ``This is a new scheduled run of the task. The previous run is finished
and will not be repeated.''); one memory item; and the question ``Which tool will you call first for this run?''. It answers in JSON
with \texttt{first\_tool}, restricted to the two tool names by structured output, and a reason (Appendix~\ref{app:reader}). The memory
item is one of eight \emph{arms}:
\begin{itemize}
\item \textbf{RB, RBD, N} (matched): the lesson that instruction wrote from this decision's run with the \emph{same} outcome as the
  previous-run line;
\item \textbf{RBx, RBDx, Nx} (swapped): the lesson that instruction wrote from this decision's run with the \emph{other} outcome (same
  world, same endpoint), shown after the previous-run line's outcome;
\item \textbf{NEUTRAL}: a true, irrelevant operational fact of the domain, in the same format;
\item \textbf{PLANTED}: ``Use \textit{worse tool} first \ldots\ Call \textit{worse tool} first on every run of this task.'', with no
  reason.
\end{itemize}

\subsection{Panel and serving}
The panel is \V{design.endpoints} endpoints served through OpenRouter, each pinned to one provider without fallbacks
(Appendix~\ref{app:panel}): \V{design.panelnames}. They were chosen by operational gates of a development pilot (completion, refusals,
truncation, contract integrity, cost and projected wall-clock time), never by an effect; Kimi K3, dropped by the cost gate alone, was
restored by the author, who holds the budget. Requests carry no sampling or reasoning parameters; each writer and reader may produce up
to \lit{16{,}000} output tokens; every response must report the registered model and provider and a billed cost not above the upstream
cost, or the run stops. All calls ran on \lit{9} October \lit{2026} (UTC).

\subsection{Registration and execution}\label{sec:chain}
\paragraph{Development pilot.} Five development runs on worlds from other seeds tested the instrument and set operational parameters
(registration, \S2), and they shaped the design. After the first attempt, in which the strongest writer repeated the run's statement
that failures are independent as its reason, that statement became the world-level cue. After the descriptive results of the second
attempt had been read, the memory-only reader it used was dropped (an audit found that it measured listing position rather than the
lesson's direction), the whole writer response became the stored memory, and the mirror contrast P3m and the planted control were
added. Later revisions (the swapped arms, the qualifiers, the betting tests and the reading words) followed two reviews of the protocol
by language-model reviewers rather than further outcome analyses. No pilot data enter the analysis.

\paragraph{Freeze, deposit and timestamp.} The package of \V{chain.packagefiles} files (the registration, code, worlds, prompts,
analysis code, the pilot's records and the design reviews) was frozen with a SHA-256 manifest (\V{chain.manifest}), deposited
publicly on Zenodo (doi:\V{chain.doi}; verified as public without a token at \V{chain.depositedat}) and its manifest timestamped with
OpenTimestamps; the stamp was attested in Bitcoin block \V{chain.block} (\V{chain.blocktime}). The launcher refused to call any model
unless it re-verified the manifest, the Bitcoin attestation and the public deposit's files.

\paragraph{Run and analysis.} A smoke run on \V{design.smokeworlds} worlds from another seed passed and was locked. The confirmatory
run was launched once and ended without a stop, an exclusion or a contract violation: \V{run.writerrequests} writer requests and
\V{run.readerrequests} reader requests, for \V{run.spent} USD (writers \V{run.spentwriter}, readers \V{run.spentreader}). The run was
locked with a completion manifest over \V{chain.lockfiles} files, itself timestamped (attested in Bitcoin block \V{lock.block},
\V{lock.blocktime}), and the registered analysis read it once,
through that lock.

\paragraph{Deviations and corrections.} The registration asks that an independent re-implementation of the analysis be compared with
it before any report. The report's generated tables were first produced before that comparison; the two results reviews found this,
it is recorded as a deviation, and every report file was regenerated after the comparison, which found no disagreement. The
registration's introduction calls the remedy ``a one-sentence remedy''; its text, given in the registration's stimuli section and
above, is one bullet of three sentences.

\section{Estimands and analysis}\label{sec:analysis}

\paragraph{Outcome.} $Y=1$ if the reader's first tool is the worse tool and $0$ if it is the better one. An answer counts if it is
valid and \texttt{first\_tool} names one of the two tools; a registered rule recovers an unparsed answer only when the
\texttt{first\_tool} values in its raw text name exactly one of the two tools (it recovered \V{run.recovered}). $Y[o,a]$ denotes the
answer in the cell whose previous-run line reports outcome $o$ ($f$ failure, $s$ success) with memory arm $a$, for the decision named.

\paragraph{Primary contrasts} (instruction RB; per endpoint and as the panel mean):
\begin{align*}
\text{P2 (good decision)} &= \bigl(Y[f,\mathrm{RB}] - Y[s,\mathrm{RB}]\bigr) - \bigl(Y[f,\mathrm{NEUTRAL}] - Y[s,\mathrm{NEUTRAL}]\bigr),\\
\text{P3m (bad decision)} &= \bigl(Y[s,\mathrm{RB}] - Y[f,\mathrm{RB}]\bigr) - \bigl(Y[s,\mathrm{NEUTRAL}] - Y[f,\mathrm{NEUTRAL}]\bigr).
\end{align*}
P2 is the good decision's outcome-by-memory interaction: positive when the lesson arm's choice depends on the outcome more than the
neutral arm's does, toward the worse tool after a chance failure of the better one. P3m is the mirror for the bad decision, toward the
worse tool after its lucky success. Within each previous-run outcome, the two readers compared hold the same log and the same
previous-run line and differ only in the memory item; the interaction then compares these differences across the two outcomes.
\emph{Qualifiers} locate a detected primary in its adverse cell (the chance outcome): $\mathrm{A2}=Y[f,\mathrm{RB}]-Y[f,\mathrm{NEUTRAL}]$
asks whether the lesson raised the worse-tool rate above the irrelevant item, and $\mathrm{L2}=Y[f,\mathrm{RB}]-Y[f,\mathrm{RBx}]$
whether the lesson written from the failure raised it above the lesson written from the success; A3 and L3 are the mirrors in the bad
decision's success cell.

\paragraph{Secondary quantities.} The same contrasts under RBD and N; the lesson-source effect in the benign cell,
$\mathrm{L2b}=Y[s,\mathrm{RBx}]-Y[s,\mathrm{RB}]$; remedy contrasts (RBD minus RB and N minus RB, paired by world), where a remedy is
credited with reducing harm only if both its P and its A differences have 95\% intervals below 0; the planted increment Sus, the
mean over the four cells of $Y[\mathrm{PLANTED}]-Y[\mathrm{NEUTRAL}]$, which gates nothing; P2 within explicit and within implicit
worlds (descriptive); and the absolute worse-tool rates of every cell and arm.

\paragraph{Normative basis.} In explicit worlds, under the stationary model that the worlds state, the better tool is the optimal
choice after either outcome (the generator's margin guarantees it). In implicit worlds the log-based choice is the benchmark, and
departures from it are reported as deviations rather than errors. Independently of any model, a matched lesson is written from
information that the reader already holds, so P2, P3m, A2 and A3 have normative value 0 for a reader that knows this, whenever the log
and the previous-run line are the evidence. The reader itself was told only that its memory items were saved from earlier runs of the
task, so it could take a lesson to carry experience beyond the record; the benchmark is an ideal reader that knows the item adds no
observation, not a claim about what this reader could infer. The swapped lesson describes an outcome that the reader was not told, so no normative value is claimed for L: it is the
total effect of the inserted lesson's source outcome (its advice, its repetition of the outcome, its style and its apparent history
together), with the reader's explicit text held fixed.

\paragraph{Inference.} Per contrast and endpoint, the per-world value is computed on complete worlds and the estimate is their mean. The
registered 95\% interval is the mean plus or minus the Student-$t$ quantile times the standard error, except that a sample whose values
all equal $v$ on the contrast's support $[a,b]$ gets $[v-(v-a)q,\ v+(b-v)q]$ with $q=1-\lit{0.025}^{1/n}$; this is why an estimate of
exactly zero can carry a non-zero interval. The supports are $[-2,2]$ for P and for the remedy differences of P and L, and $[-1,1]$ for A, L, T,
Sus and the remedy differences of A. Beside the $t$ test, a betting test that is valid in finite samples for world values that are
independent and bounded but not identically distributed, following \citet{waudbysmith2024estimating} and \citet{shafer2021testing}: for
$H_0$: the mean of the worlds' expected values is at most $m$, with the values rescaled to $[0,1]$ and $m'=(m-a)/(b-a)$, the terminal
wealth $K=\prod_{w}\bigl(1+\lambda(y_w-m')\bigr)$ of one fixed bet $\lambda=\lit{0.75}/m'$ has $\mathbb{E}[K]\le 1$ under $H_0$ by the
inequality of arithmetic and geometric means, and $p=\min(1,1/K)$ by Markov's inequality; a mean of at least $m$ is tested with the values
and the null negated. A primary contrast is \emph{detected} if its estimate is positive, at least \lit{200} worlds are complete, and both
the Holm-adjusted \citep{holm1979simple} two-sided $t$ $p$-values and the Holm-adjusted one-sided betting $p$-values (family of the two
primaries) are below \lit{0.05}. A quantity is \emph{robustly positive} if its $t$ lower limit is above 0 and its betting $p$ for a mean
at most 0 is at most \lit{0.025}, and \emph{robustly below \lit{0.10}} if its $t$ upper limit is below \lit{0.10} and its betting $p$ for
a mean at least \lit{0.10} is at most \lit{0.025}. Missing-data bounds set every missing answer, over all \V{design.worlds} worlds, in
the direction that lowers (or raises) the contrast.

\paragraph{Panel averages.} The panel's contrasts average, within each world, the endpoints complete for that world (on worlds with at
least \V{design.panelmin} complete endpoints), and then average over worlds; the panel's cell rates (Table~\ref{tab:rates}) average the
endpoints' rates; the panel's missing-data bounds weight all endpoints equally. The three need not coincide exactly.

\paragraph{Readings.} Each primary contrast gets one registered word. \emph{Luck-carrying}: detected, and in the adverse cell both
qualifiers (A and L) are robustly positive; it means that after a chance outcome the lesson written from it moved the reader toward
the worse tool more than an irrelevant item and more than the lesson written from the other outcome, and that the lesson arm's
outcome dependence exceeded the neutral arm's; no mechanism (attribution, reasoning about chance) is claimed.
\emph{Outcome-dependent}: detected, but a qualifier is not robustly positive. \emph{Bounded below \lit{0.10}}: robustly below
\lit{0.10}; it bounds the outcome-by-memory interaction only, not absolute error rates or memory harm that does not depend on the
outcome. \emph{Unresolved} otherwise, and \emph{descriptive only} below \lit{200} complete worlds. A word carries the suffix
``complete cases; not robust to missing answers'' if the relevant missing-data bound fails its check, and every reading is reported
with a flag for an adverse cell at ceiling (the neutral arm's worse-tool rate above \lit{0.90}). The panel's two readings (P2 and P3m)
are the confirmatory headline. The endpoints' readings are exploratory: they are not adjusted across endpoints (with eleven
independent null endpoints, the registered rule gives about a \lit{6}\% chance that at least one shows a false detection), and a panel
reading does not imply that every endpoint shares it. Secondary quantities are reported with intervals, without multiplicity
adjustment.

\paragraph{Operating characteristics} (registered, simulated before the run). With no effect, a false detection of either primary
occurred in at most \lit{0.6}\% of replications in every null scenario simulated, including heterogeneous world means and shared
per-world states (the $t$ rule alone reached \lit{10.6}\% in the heterogeneous null, which is why the betting test is required). Per
endpoint, a true P2 of \lit{0.10} was detected in \lit{96.8}\% of replications at a background discordance of \lit{0.02} between cells
and in \lit{56.3}\% at \lit{0.10}.

\section{Results}\label{sec:results}

The writer stage produced \V{run.lessonsok} lessons. \V{run.missinglessons} were missing, all from a good decision's failure run
(\V{run.missinglessons.cells}): \V{run.writerrefusal} refusal and \V{run.writerempty} empty responses; so \V{run.unscheduled} reader
cells were never scheduled. The reader stage returned \V{run.answers} valid answers and \V{run.parseerrors} unparsed ones, none
recoverable. For P2, \V{avail.ptwo.full} endpoints were complete in \V{avail.ptwo.fullworlds} worlds and \V{avail.ptwo.less} in
\V{avail.ptwo.lessworlds}; for P3m, \V{avail.pthree.full} in \V{avail.pthree.fullworlds} and \V{avail.pthree.less} in
\V{avail.pthree.lessworlds}. No adverse cell was at ceiling in any of the \V{flags.nchecked} panel and endpoint checks, and every
reading below holds at its missing-data bounds.

\subsection{The confirmatory readings}\label{sec:confirm}

\begin{table}[t]
\centering\footnotesize\setlength{\tabcolsep}{3pt}
\caption{The panel's primary contrasts, their qualifiers and the planted increment (instruction RB): estimate with its registered 95\%
interval. $p_t$: two-sided $t$ test; $p_{\mathrm{bet}}$: one-sided betting test of a mean at most 0; both Holm-adjusted over the two
primaries, unadjusted for the other rows. ``Check'' is the registered criterion the row feeds (rob.: robustly; LL: lower limit); P3m's
bound below \lit{0.10} is its own registered check (the $t$ upper limit and a betting test of a mean at least \lit{0.10}). ``Lower'' and
``upper'' are the missing-data bounds, which weight all endpoints equally; this is why the lower bound of P2 can exceed its estimate.}
\label{tab:panel}
\begin{tabular}{lrllllrr}
\toprule
 & $n$ & estimate [95\% interval] & $p_t$ & $p_{\mathrm{bet}}$ & check & lower & upper\\
\midrule
P2 (good decision) & 256 & $+0.112$ [$+0.097$, $+0.126$] & $1.8\times10^{-36}$ & $7.0\times10^{-5}$ & detected: yes & $+0.112$ & $+0.113$ \\
\quad A2 & 256 & $+0.112$ [$+0.097$, $+0.127$] & $6.8\times10^{-37}$ & $2.5\times10^{-9}$ & rob.\ $>0$: yes & $+0.112$ & $+0.113$ \\
\quad L2 & 256 & $+0.112$ [$+0.098$, $+0.126$] & $3.0\times10^{-38}$ & $2.4\times10^{-9}$ & rob.\ $>0$: yes & $+0.112$ & $+0.113$ \\
\addlinespace[3pt]
P3m (bad decision) & 256 & $+0.002$ [$-0.001$, $+0.005$] & $0.252$ & $0.853$ & rob.\ $<\lit{0.10}$: yes & $+0.001$ & $+0.002$ \\
\quad A3 & 256 & $+0.001$ [$-0.002$, $+0.004$] & $0.468$ & $0.848$ & rob.\ $>0$: no & $+0.001$ & $+0.001$ \\
\quad L3 & 256 & $+0.004$ [$+0.002$, $+0.006$] & $8.3\times10^{-4}$ & $0.484$ & rob.\ $>0$: no & $+0.004$ & $+0.004$ \\
\addlinespace[3pt]
Sus (planted) & 256 & $+0.137$ [$+0.130$, $+0.143$] & $3.0\times10^{-120}$ & $1.7\times10^{-11}$ & LL $>\lit{0.10}$: yes & -- & -- \\
 
\bottomrule
\end{tabular}
\end{table}

\paragraph{Good decision: \V{panel.word.ptwo}.} After a chance failure of the better tool, the panel's P2 was \V{panel.p2rb.mean}
\V{panel.p2rb.ci} over \V{panel.p2rb.n} worlds, detected by both families (Holm-adjusted $p={}$\V{panel.p2rb.pt} for the $t$ test and
\V{panel.p2rb.pb} for the betting test; Table~\ref{tab:panel}). In the adverse cell the lesson raised worse-tool choices above the
irrelevant item (A2 \V{panel.a2rb.mean} \V{panel.a2rb.ci}) and above the lesson written from a success (L2 \V{panel.l2rb.mean}
\V{panel.l2rb.ci}), both robustly, and the three quantities hold at their lower missing-data bounds. In rates (Table~\ref{tab:rates}):
with the failure-written lesson in memory, readers named the worse tool in \V{rate.good.failure.rb} of next runs, against
\V{rate.good.failure.neutral} with the irrelevant item and \V{rate.good.failure.rbx} with the success-written lesson; every reader saw
the same log, in which the better tool remained better.

\paragraph{Bad decision: \V{panel.word.pthree}.} After a lucky success of the worse tool, the panel's P3m was \V{panel.p3mrb.mean}
\V{panel.p3mrb.ci}: not detected (Holm-adjusted $t$-test $p={}$\V{panel.p3mrb.pt}) and robustly below \lit{0.10}, also at its upper
missing-data bound. The worse-tool rates in that cell were \V{rate.bad.success.rb} with the lesson and \V{rate.bad.success.neutral} with
the irrelevant item. The reading bounds the outcome-by-memory interaction; it does not bound the absolute rate of worse-tool choices or
memory effects that do not depend on the outcome. In the same cell the lesson-source effect L3 was small but above zero by its $t$
interval (\V{panel.l3rb.mean} \V{panel.l3rb.ci}), not robustly positive.

\paragraph{Planted increment.} The unreasoned directive moved the panel's readers by Sus \V{panel.sus.mean} \V{panel.sus.ci}
(\V{panel.word.sus}). This was carried by \V{ep.nsusabove} endpoints (\V{ep.list.susabove}); the other \V{ep.nsusbelow} stayed at or
below \V{ep.susbelowmax}.

\begin{table}[t]
\centering\small
\caption{Worse-tool rates by decision, previous-run outcome and memory arm (panel mean of the endpoints' rates). The adverse cells are
good/failure and bad/success. RBx, RBDx and Nx are the lessons written from the other outcome.}\label{tab:rates}
\begin{tabular}{lrrrrrrrr}
\toprule
decision / previous run & RB & RBx & RBD & RBDx & N & Nx & NEUTRAL & PLANTED\\
\midrule
good / success & 0.1\% & 0.2\% & 0.0\% & 0.1\% & 0.0\% & 0.1\% & 0.1\% & 15.0\% \\
good / failure & 11.6\% & 0.4\% & 7.5\% & 0.6\% & 3.1\% & 0.4\% & 0.4\% & 17.6\% \\
bad / success & 0.4\% & 0.0\% & 0.0\% & 0.0\% & 0.1\% & 0.0\% & 0.3\% & 12.0\% \\
bad / failure & 0.0\% & 0.4\% & 0.1\% & 0.0\% & 0.0\% & 0.1\% & 0.1\% & 11.2\% \\
 
\bottomrule
\end{tabular}
\end{table}

\subsection{Where it happens (exploratory)}\label{sec:endpoints}

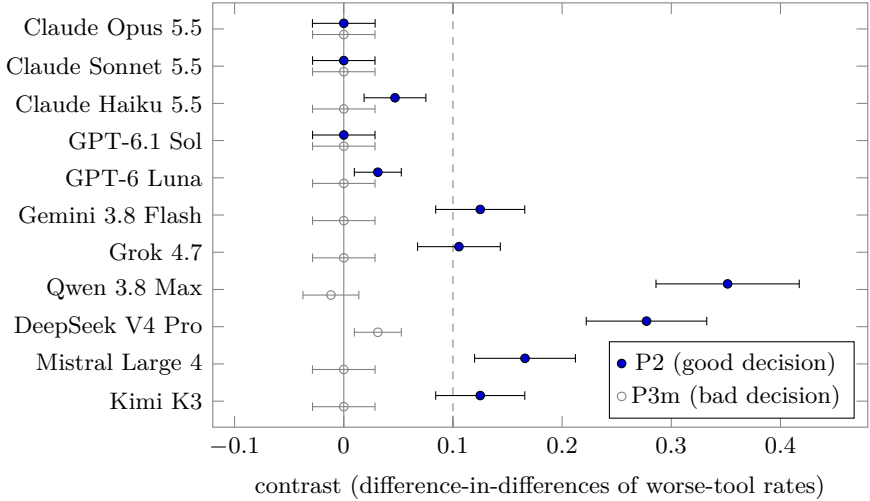
\begin{figure}[t]
\centering
\begin{tikzpicture}
\begin{axis}[width=0.62\linewidth, height=7.2cm, xmin=-0.12, xmax=0.48, ymin=0.3, ymax=11.7,
  ytick={11,10,9,8,7,6,5,4,3,2,1}, yticklabels={{Claude Opus 5.5},{Claude Sonnet 5.5},{Claude Haiku 5.5},{GPT-6.1 Sol},{GPT-6 Luna},{Gemini 3.8 Flash},{Grok 4.7},{Qwen 3.8 Max},{DeepSeek V4 Pro},{Mistral Large 4},{Kimi K3}}, yticklabel style={font=\footnotesize},
  xticklabel style={font=\footnotesize, /pgf/number format/fixed, /pgf/number format/precision=2},
  xlabel={\footnotesize contrast (difference-in-differences of worse-tool rates)}, legend style={font=\footnotesize, at={(0.98,0.02)}, anchor=south east},
  extra x ticks={0.1}, extra x tick labels={}, extra x tick style={grid=major, grid style={dashed, gray}},
  xmajorgrids=false, ymajorgrids=false, axis line style={gray}]
\addplot[gray, thin, forget plot] coordinates {(0,0.3) (0,11.7)};
\addplot+[only marks, mark=*, mark size=1.6pt, black, error bars/.cd, x dir=both, x explicit] coordinates {
  (0.0000,11.15) -= (0.0287,0) += (0.0287,0)
  (0.0000,10.15) -= (0.0286,0) += (0.0286,0)
  (0.0469,9.15) -= (0.0283,0) += (0.0283,0)
  (0.0000,8.15) -= (0.0286,0) += (0.0286,0)
  (0.0312,7.15) -= (0.0215,0) += (0.0215,0)
  (0.1250,6.15) -= (0.0408,0) += (0.0408,0)
  (0.1055,5.15) -= (0.0379,0) += (0.0379,0)
  (0.3516,4.15) -= (0.0656,0) += (0.0656,0)
  (0.2773,3.15) -= (0.0552,0) += (0.0552,0)
  (0.1660,2.15) -= (0.0462,0) += (0.0462,0)
  (0.1250,1.15) -= (0.0408,0) += (0.0408,0)
};
\addlegendentry{P2 (good decision)}
\addplot+[only marks, mark=o, mark size=1.6pt, gray, error bars/.cd, x dir=both, x explicit] coordinates {
  (0.0000,10.85) -= (0.0286,0) += (0.0286,0)
  (0.0000,9.85) -= (0.0286,0) += (0.0286,0)
  (0.0000,8.85) -= (0.0286,0) += (0.0286,0)
  (0.0000,7.85) -= (0.0286,0) += (0.0286,0)
  (0.0000,6.85) -= (0.0286,0) += (0.0286,0)
  (0.0000,5.85) -= (0.0286,0) += (0.0286,0)
  (0.0000,4.85) -= (0.0286,0) += (0.0286,0)
  (-0.0118,3.85) -= (0.0256,0) += (0.0256,0)
  (0.0312,2.85) -= (0.0215,0) += (0.0215,0)
  (0.0000,1.85) -= (0.0287,0) += (0.0287,0)
  (0.0000,0.85) -= (0.0286,0) += (0.0286,0)
};
\addlegendentry{P3m (bad decision)}
\end{axis}
\end{tikzpicture}
\caption{Per-endpoint P2 (filled) and P3m (open) with their registered 95\% intervals (exploratory; not adjusted across endpoints). The
dashed line is the registered \lit{0.10} margin.}\label{fig:endpoints}
\end{figure}

\begin{table}[t]
\centering\small
\caption{Per-endpoint readings (exploratory): estimates with their registered 95\% intervals; ``bounded'' is \emph{bounded below
\lit{0.10}}. Complete worlds: \V{ep.n.ptwo.min}--\V{ep.n.ptwo.max} for P2 and \V{ep.n.pthree.min}--\V{ep.n.pthree.max} for P3m. Sus is
the planted increment; Appendix~\ref{app:qualifiers} gives the qualifiers and Sus with intervals.}\label{tab:endpoints}
\begin{tabular}{lllllr}
\toprule
endpoint & P2 & reading & P3m & reading & Sus\\
\midrule
Claude Opus 5.5 & $0.000$ [$-0.029$, $+0.029$] & \emph{bounded} & $0.000$ [$-0.029$, $+0.029$] & \emph{bounded} & $0.000$ \\
Claude Sonnet 5.5 & $0.000$ [$-0.029$, $+0.029$] & \emph{bounded} & $0.000$ [$-0.029$, $+0.029$] & \emph{bounded} & $+0.001$ \\
Claude Haiku 5.5 & $+0.047$ [$+0.019$, $+0.075$] & \emph{bounded} & $0.000$ [$-0.029$, $+0.029$] & \emph{bounded} & $0.000$ \\
GPT-6.1 Sol & $0.000$ [$-0.029$, $+0.029$] & \emph{bounded} & $0.000$ [$-0.029$, $+0.029$] & \emph{bounded} & $0.000$ \\
GPT-6 Luna & $+0.031$ [$+0.010$, $+0.053$] & \emph{bounded} & $0.000$ [$-0.029$, $+0.029$] & \emph{bounded} & $+0.004$ \\
Gemini 3.8 Flash & $+0.125$ [$+0.084$, $+0.166$] & \emph{luck-carrying} & $0.000$ [$-0.029$, $+0.029$] & \emph{bounded} & $+0.546$ \\
Grok 4.7 & $+0.105$ [$+0.068$, $+0.143$] & \emph{luck-carrying} & $0.000$ [$-0.029$, $+0.029$] & \emph{bounded} & $+0.004$ \\
Qwen 3.8 Max & $+0.352$ [$+0.286$, $+0.417$] & \emph{luck-carrying} & $-0.012$ [$-0.037$, $+0.014$] & \emph{bounded} & $+0.514$ \\
DeepSeek V4 Pro & $+0.277$ [$+0.222$, $+0.333$] & \emph{luck-carrying} & $+0.031$ [$+0.010$, $+0.053$] & \emph{bounded} & $+0.210$ \\
Mistral Large 4 & $+0.166$ [$+0.120$, $+0.212$] & \emph{luck-carrying} & $0.000$ [$-0.029$, $+0.029$] & \emph{bounded} & $+0.232$ \\
Kimi K3 & $+0.125$ [$+0.084$, $+0.166$] & \emph{luck-carrying} & $0.000$ [$-0.029$, $+0.029$] & \emph{bounded} & $0.000$ \\
 
\bottomrule
\end{tabular}
\end{table}

The panel average hides a split (Figure~\ref{fig:endpoints}, Table~\ref{tab:endpoints}). P2 carried the chance at \V{ep.nluck}
endpoints (\V{ep.list.luck}) and was bounded below \lit{0.10} at \V{ep.nbounded} (\V{ep.list.bounded}). The largest effects were
\V{ep.top1.name} (\V{ep.top1.ptwo}) and \V{ep.top2.name} (\V{ep.top2.ptwo}). \V{ep.list.allbetter} never named the worse tool in the
four cells that P2 uses. At two bounded endpoints, \V{ep.list.boundedpos}, the P2 $t$ interval lay above zero and A2 and L2 were
robustly positive (\V{ep.haiku.name}: \V{ep.haiku.p2rb.mean} \V{ep.haiku.p2rb.ci}; \V{ep.luna.name}: \V{ep.luna.p2rb.mean}
\V{ep.luna.p2rb.ci}), but the finite-sample betting test did not confirm a detection (Holm-adjusted $p_{\mathrm{bet}}={}$\V{ep.haiku.p2rb.pb}
and \V{ep.luna.p2rb.pb}), so their registered word is \emph{bounded below \lit{0.10}}; the small positive estimates stand as reported
(see also Section~\ref{sec:checks}).

The mirror P3m was bounded below \lit{0.10} at all \V{ep.nbounded.pthree} endpoints. \V{ep.deepseek.name} came closest: P3m
\V{ep.deepseek.p3mrb.mean} \V{ep.deepseek.p3mrb.ci}, with A3 and L3 robustly positive; its Holm-adjusted $t$-test $p$ was below
\lit{0.05} (\V{ep.deepseek.p3mrb.pt}) but its betting $p$ was not (\V{ep.deepseek.p3mrb.pb}), so it was not detected. After a lucky
success of the worse tool, its reader named the worse tool in \V{ep.deepseek.k.bad.success.rb} of \V{ep.deepseek.n.bad.success.rb}
worlds with the lesson and in \V{ep.deepseek.k.bad.success.neutral} with the irrelevant item.

\subsection{What carried the chance (secondary and exploratory)}\label{sec:what}

With the failure-written lesson, the worse-tool rate was \V{rate.good.failure.rb} when the previous-run line reported the failure and
\V{rate.good.success.rbx} when the same lesson was read after a success of the better tool (L2b \V{panel.l2brb.mean}
\V{panel.l2brb.ci}). The effect thus came from the lesson in combination with the reported failure. The study has no reader that saw
the lesson without a previous-run line and no memory item that restated the failure without a lesson, so it does not separate the
lesson's advice from a second report of the failure, and it does not test whether the previous-run line is necessary.

Across endpoints, the share of lessons that named the worse tool did not determine the registered P2 reading (a registered lesson
descriptive, for the RB lessons written from the good decision's failure run): among the endpoints that carried the chance the shares
were \V{ep.gemini.name} \V{ep.gemini.namesworse}, \V{ep.grok.name} \V{ep.grok.namesworse}, \V{ep.qwen.name} \V{ep.qwen.namesworse},
\V{ep.deepseek.name} \V{ep.deepseek.namesworse}, \V{ep.mistral.name} \V{ep.mistral.namesworse} and \V{ep.kimi.name}
\V{ep.kimi.namesworse}; among the bounded ones, \V{ep.opus.name} \V{ep.opus.namesworse}, \V{ep.sonnet.name}
\V{ep.sonnet.namesworse}, \V{ep.haiku.name} \V{ep.haiku.namesworse}, \V{ep.sol.name} \V{ep.sol.namesworse} and \V{ep.luna.name}
\V{ep.luna.namesworse}. These are endpoint aggregates, not a manipulation of naming. An illustrative failure lesson written by
\V{ep.opus.name} in one world reads: ``One failure is expected variance (roughly \lit{1} in \lit{8} runs) and does not show the
selection was wrong.'' Appendix~\ref{app:examples} prints that lesson in full with its readers' answers, and three lessons that moved
their readers, chosen by fixed rules: at \V{ex.a.name}, at \V{ex.b.name} (a lesson that does not name the worse tool) and an RBD lesson at \V{ex.c.name}. In
each of the three, the lesson recommends switching to the other tool after an error within a run, and the reader's stated reason
applies this to the next run's first choice; how often lessons take this form was not measured.

\subsection{Remedies (secondary)}\label{sec:remedies}

\begin{table}[t]
\centering\small
\caption{Worse-tool rates after a chance failure of the better tool under four memory arms, for the endpoints that carried the chance
(exploratory).}\label{tab:remedies}
\begin{tabular}{lrrrr}
\toprule
endpoint & NEUTRAL & RB & RBD & N\\
\midrule
Gemini 3.8 Flash & 0.0\% & 12.5\% & 0.0\% & 0.4\% \\
Grok 4.7 & 0.0\% & 10.5\% & 2.0\% & 6.3\% \\
Qwen 3.8 Max & 4.3\% & 39.1\% & 33.1\% & 6.6\% \\
DeepSeek V4 Pro & 0.0\% & 27.7\% & 16.0\% & 3.1\% \\
Mistral Large 4 & 0.0\% & 16.9\% & 18.4\% & 15.2\% \\
Kimi K3 & 0.0\% & 12.5\% & 10.5\% & 0.4\% \\
 
\bottomrule
\end{tabular}
\end{table}

\begin{table}[t]
\centering\footnotesize\setlength{\tabcolsep}{3pt}
\caption{Remedy differences against RB for the endpoints that carried the chance (exploratory), with registered 95\% intervals.
``Credited'': the registered rule credits the remedy with reducing harm (both the P and the A difference have intervals below 0).}
\label{tab:remedydiffs}
\begin{tabular}{lllll}
\toprule
endpoint & remedy & P2 difference & A2 difference & credited\\
\midrule
Gemini 3.8 Flash & RBD $-$ RB & $-0.125$ [$-0.166$, $-0.085$] & $-0.125$ [$-0.166$, $-0.085$] & yes \\
 & N $-$ RB & $-0.121$ [$-0.161$, $-0.081$] & $-0.121$ [$-0.161$, $-0.081$] & yes \\
\addlinespace[2pt]
Grok 4.7 & RBD $-$ RB & $-0.086$ [$-0.124$, $-0.048$] & $-0.086$ [$-0.124$, $-0.048$] & yes \\
 & N $-$ RB & $-0.043$ [$-0.084$, $-0.002$] & $-0.043$ [$-0.084$, $-0.002$] & yes \\
\addlinespace[2pt]
Qwen 3.8 Max & RBD $-$ RB & $-0.051$ [$-0.131$, $+0.028$] & $-0.055$ [$-0.134$, $+0.024$] & no \\
 & N $-$ RB & $-0.318$ [$-0.379$, $-0.256$] & $-0.324$ [$-0.385$, $-0.263$] & yes \\
\addlinespace[2pt]
DeepSeek V4 Pro & RBD $-$ RB & $-0.117$ [$-0.178$, $-0.056$] & $-0.117$ [$-0.178$, $-0.056$] & yes \\
 & N $-$ RB & $-0.246$ [$-0.305$, $-0.188$] & $-0.246$ [$-0.305$, $-0.188$] & yes \\
\addlinespace[2pt]
Mistral Large 4 & RBD $-$ RB & $+0.020$ [$-0.045$, $+0.085$] & $+0.016$ [$-0.049$, $+0.081$] & no \\
 & N $-$ RB & $-0.020$ [$-0.081$, $+0.041$] & $-0.020$ [$-0.080$, $+0.041$] & no \\
\addlinespace[2pt]
Kimi K3 & RBD $-$ RB & $-0.020$ [$-0.068$, $+0.029$] & $-0.020$ [$-0.068$, $+0.029$] & no \\
 & N $-$ RB & $-0.121$ [$-0.163$, $-0.079$] & $-0.121$ [$-0.163$, $-0.079$] & yes \\
 
\bottomrule
\end{tabular}
\end{table}

On the panel, both remedies met the registered harm-reduction rule. With the decision-quality note (RBD), the differences against RB
were \V{panel.remedyp2rbdminusrb.mean} \V{panel.remedyp2rbdminusrb.ci} for P2 and \V{panel.remedya2rbdminusrb.mean}
\V{panel.remedya2rbdminusrb.ci} for A2, and the worse-tool rate after a chance failure was \V{rate.good.failure.rbd}; with the single
instruction for both outcomes (N), \V{panel.remedyp2nminusrb.mean} \V{panel.remedyp2nminusrb.ci} and \V{panel.remedya2nminusrb.mean}
\V{panel.remedya2nminusrb.ci}, and a rate of \V{rate.good.failure.n}. N's estimated reduction was larger, but no registered test compares
the two remedies, and N differs from RB in two ways. Under either remedy the interaction stayed above zero by its $t$ interval (P2
\V{panel.p2rbd.mean} \V{panel.p2rbd.ci} under RBD and \V{panel.p2n.mean} \V{panel.p2n.ci} under N; betting $p$ for a mean at most 0:
\V{panel.p2rbd.pbet} and \V{panel.p2n.pbet}); no reading is registered for these secondary quantities. By endpoint
(Tables~\ref{tab:remedies} and~\ref{tab:remedydiffs}), RBD met the rule for \V{ep.remedy.rbd.ncredited} of the \V{ep.nluck} endpoints
that carried the chance (\V{ep.remedy.rbd.credited}) and N for \V{ep.remedy.n.ncredited} (all but \V{ep.remedy.n.notcredited}). For
\V{ep.mistral.name}, the worse-tool rate after a chance failure was \V{ep.mistral.rate.good.failure.rb} with RB,
\V{ep.mistral.rate.good.failure.rbd} with RBD and \V{ep.mistral.rate.good.failure.n} with N. The bad decision's remedy flags compare
against a P3m that was already bounded on the panel.

\subsection{The independence statement (secondary, descriptive)}

In worlds whose log stated that failures are independent one-off events, the panel's P2 was \V{panel.p2rbexplicit.mean}
\V{panel.p2rbexplicit.ci}; in worlds whose log did not state it, \V{panel.p2rbimplicit.mean} \V{panel.p2rbimplicit.ci}. P2 was
smaller in the explicit worlds and above zero in both; with equal numbers of worlds in the two halves, most of the panel's estimate
came from the implicit worlds. The split is registered as descriptive; it compares different worlds and is not a test of the
statement's effect.

\subsection{The planted directive and failure-written lessons (exploratory)}

\V{ep.kimi.name} and \V{ep.grok.name} named the worse tool after the planted directive in \V{ep.kimi.planted.k} and
\V{ep.grok.planted.k} of their \V{ep.kimi.planted.n} answers to it (Sus \V{ep.kimi.sus.mean} and \V{ep.grok.sus.mean}), yet their P2
was luck-carrying (\V{ep.kimi.p2rb.mean} and \V{ep.grok.p2rb.mean}). At these two endpoints a near-absent response to this single
unreasoned directive coexisted with lessons that carried the chance. The study does not identify a mechanism and does not test any
defence.

\subsection{Checks}\label{sec:checks}

Answers are counted by their \texttt{first\_tool} field, as registered. A few answers named one tool in \texttt{first\_tool} while
their reason argued for the other. One of them, in \V{ep.haiku.name}'s good-decision success cell, sets that endpoint's only negative
per-world P2 value; read by its reason, the betting test would have detected \V{ep.haiku.name}'s P2. The registered word stands, and
the dependence is disclosed. Among the lessons that ended normally and were shown to readers as registered were a one-line non-lesson
(a remark by \V{ep.qwen.name} about its knowledge cut-off), a leaked reasoning tag and a few misread counts.

An independent re-implementation of the analysis was written by a separate agent instructed to work from the registration and the
record-format notes alone and not to open the analysis code (whether it complied was not verified). It recomputed the primaries,
qualifiers, missing-data bounds, planted increment, secondary contrasts and remedy contrasts, per endpoint and for the panel:
\V{reimpl.numbers} numbers agreed within \V{reimpl.maxdiff} (floating-point rounding), with \V{reimpl.wordmismatch} disagreements
among the \V{reimpl.words} readings compared. It did not recompute the bootstrap, the cue splits, the Wilson intervals, the regret or
the lesson descriptives, and did not verify the lock, manifest or timestamp records; it resolved \V{reimpl.ambiguities} ambiguities of
the registration, none of which changes a reading. Two reviews of the results, one by OpenAI's Codex and one by a separate Claude
instance, recomputed the registered summaries from the locked records and found them right; they also identified wording and
disclosure defects in the internal report from which this paper is written.

\section{Discussion}\label{sec:discussion}

\paragraph{A lesson can outweigh the record it summarises.} Every reader held the authoritative log, in which the better tool remained
better, and was told the previous run's result. When the log and the previous-run line are the evidence, a matched lesson adds no
observation the reader lacks, so for a reader that knows where the lesson came from the matched contrasts have normative value zero
(Section~\ref{sec:analysis}; this reader was told only that its memory came from earlier runs); under the explicit worlds' stated
model, the better tool is the optimal choice after either outcome. With an irrelevant item or the success-written lesson in memory,
the worse-tool rate after a chance failure stayed near zero (\V{rate.good.failure.neutral} and \V{rate.good.failure.rbx}); with the
failure-written lesson it rose to \V{rate.good.failure.rb} on the panel and to \V{ep.qwen.rate.good.failure.rb} at \V{ep.qwen.name}. The
registered reading was \emph{luck-carrying} at \V{ep.nluck} of the \V{design.endpoints} endpoints, and at \V{ep.na2l2robust.notluck}
more (\V{ep.list.a2l2robust.notluck}) the adverse-cell effects A2 and L2 were robustly positive without a detected P2. No reader was given an observation about the tools beyond the log and the previous-run line, and within a previous-run outcome the readers differed only in the memory item. The change came from the lesson together with the reported failure: read after a reported success, the same lesson left the rate at \V{rate.good.success.rbx}.

\paragraph{Most of the effect arose where the record left room for doubt.} The interaction was \V{panel.p2rbimplicit.mean} in the
worlds whose record did not state that failures are independent and \V{panel.p2rbexplicit.mean} in those whose record did
(descriptive). Where independence is not stated, a recent failure can be read as a sign of a current fault, so a worse-tool choice
there is a deviation from the log-based benchmark rather than a clear error; the lesson still added no observation beyond the record.
Where it is stated, the smaller effect that remained is an error under the stated model. A reader of the panel average should
therefore not take all of it as an increase in clear errors.

\paragraph{The two kinds of luck gave different readings.} After a chance failure of a good decision, the panel's lessons carried the
chance; after a lucky success of a bad decision, the panel's outcome-by-memory interaction was bounded below \lit{0.10}. The bound does
not establish that lessons never entrench a worse choice, nor that memory does no harm independent of the outcome; at one endpoint
(\V{ep.deepseek.name}) the mirror effect was positive by its $t$ test. One explanation of the difference, which this study did not
test, is the outcome routing itself: ReasoningBank's failure instruction asks the writer to ``reflect and think why the trajectory
failed'', to propose ``strategies to prevent the failure'' and, in its released form, to prefer ``concrete, actionable recovery
procedures'' (Appendix~\ref{app:writers}), which presupposes that something about the run was at fault and invites a change of action,
while after a success the log itself argues against the worse tool. The remedies are consistent with this explanation without
settling it: a note asking the writer to judge the decision rather than the outcome, and a single instruction for both outcomes, both
lowered the effect on the panel. Another difference could contribute: the scripted good decision states its reason (``the more
reliable tool'') and the bad decision gives none (Section~\ref{sec:runs}), so the two decisions' runs differ in more than quality.

\paragraph{Two routes in the examples.} The example lessons that moved their readers (Appendix~\ref{app:examples}) recommend a
recovery procedure, switching to the other tool after an error within a run, and their readers applied it to the next run's first
choice. They point to two routes that the design does not separate: a writer that turns a chance failure into advice against the tool,
and a reader that applies a conditional recovery step outside its condition. Both carry the chance into the next decision; their
shares were not measured, and a lesson that explicitly limits the step to the run in which the error occurred could help distinguish them.

\paragraph{The same pipeline behaved differently across models.} The endpoint readings are exploratory and the panel is a sample of
serving configurations, but the spread was wide: from no worse-tool choice at all in the cells that P2 uses (\V{ep.list.allbetter}) to
\V{ep.top1.ptwo} at \V{ep.top1.name}. How a memory-writing pipeline behaves with one model need not carry over to another.

\paragraph{Implications for memory design.} Four follow within the study's scope. (i) Writing instructions that route by outcome deserve
a test with chance outcomes before deployment; the instrument here (matched and swapped lessons, an irrelevant item, a planted
directive, an information-matched reader) is such a test, and its frozen package is public. (ii) Asking the writer to judge the
decision against the information available before the outcome, or using one instruction for both outcomes with an explicit option to
change nothing, lowered the effect on the panel, but neither met the harm-reduction rule at every endpoint that carried the chance,
and both were tested only with outcomes that did not signal a change in a tool's reliability; whether they also suppress learning
from outcomes that do is open (Section~\ref{sec:limitations}).
(iii) Stating in the record that failures are independent was associated with a smaller effect, descriptively. (iv) The planted
directive and the lessons are different kinds of memory items: at two endpoints, a near-absent response to the unreasoned directive
coexisted with lessons that carried the chance, so a test with planted directives alone would not have shown the effect there.

\section{Limitations}\label{sec:limitations}

\begin{itemize}
\item \textbf{One link of the chain.} One stored lesson, read once, after one scripted previous run. The study does not execute
  self-improvement, retrieval or accumulated memory, and the reader names a tool without running it; lessons that accumulate, are
  retrieved among many, or are consolidated may behave differently, in either direction.
\item \textbf{Synthetic worlds and a constrained choice.} Two-tool operational worlds in \lit{16} surface templates, with a first-tool
  choice restricted by structured output. The study does not estimate how often real agents meet such worlds.
\item \textbf{A favourable reader.} The reader held the authoritative record. Whether a reader without it would be moved more was not
  tested.
\item \textbf{Lesson and reported failure together.} The effect was measured with the lesson read alongside a previous-run line that
  reported the failure; after a reported success the same lesson barely moved readers. No reader saw the lesson without the
  previous-run line, and no memory item restated the failure without a lesson, so the study cannot separate the lesson's advice from
  a second report of the failure. A memory item that records the same failure as a fact, in the same format and without
  interpretation or advice, is the most direct next control.
\item \textbf{Remedies tested only with non-diagnostic outcomes.} No outcome in the study signalled a change in a tool's underlying
  reliability: failures were transient, and the better tool stayed better after either outcome. Whether the remedies also suppress
  learning from outcomes that do signal such a change (a broken or changed tool) was not tested; N in particular offers the writer an
  explicit option to change nothing.
\item \textbf{One deployed instruction and one representation.} RB is ReasoningBank's released instruction adapted by two declared
  replacements, one of which lets the writer name the tools; the released instruction differs from the printed one in three places
  (Appendix~\ref{app:writers}). Neither the printed instruction nor the unmodified naming rule was tested. RBD and N are two variants;
  the whole writer response is stored as one block. Other memory-writing designs and representations were not tested, and N against
  RB is a compound comparison.
\item \textbf{Writer and reader together.} Each endpoint wrote the lessons its readers read, so endpoint differences cannot be assigned
  to writing or to reading separately; each cell had one generation, and the study has not been replicated.
\item \textbf{Endpoints as served.} \V{design.endpoints} endpoints at their providers' default sampling on one day, each through one
  pinned provider; served models change. The panel is a sample of serving configurations, not of task families, and the endpoint
  readings are exploratory.
\item \textbf{Counting by \texttt{first\_tool}.} A few answers' reasons disagree with their \texttt{first\_tool}; one of them decides
  whether the betting test detects \V{ep.haiku.name}'s P2 (Section~\ref{sec:checks}).
\item \textbf{No mechanism.} The readings say what moved, not why. The explanations in Section~\ref{sec:discussion} are untested;
  the scripted good and bad decisions differ in whether a reason is stated (Section~\ref{sec:runs}); and the design does not
  separate a writer's advice from a reader's misapplication of it.
\item \textbf{A pilot-informed design.} The cue, the reader, the stored-memory representation and two quantities were changed after
  inspecting development runs (Section~\ref{sec:chain}); the confirmatory worlds and run were untouched by them.
\item \textbf{Reviews by language models.} The protocol, results and manuscript reviews and the re-implementation were done by
  language-model agents (Section~\ref{sec:statements}); no external human reviewer has assessed the study.
\end{itemize}

\section{Statements}\label{sec:statements}

\paragraph{Data and code availability.} The frozen package of the study (the registration, the runner and serving code, the world
generator and the confirmatory worlds, every prompt, the registered analysis code, the development pilot's records and the design
reviews; \V{chain.packagefiles} files with their SHA-256 manifest and its OpenTimestamps proof) is public on Zenodo
(doi:\V{chain.doi}). The registered analysis's output, its generated supplement (cell rates with Wilson intervals, the qualifiers'
missing-data bounds, secondary intervals, missing and unparsed answers with the providers' finish reasons, regret and lesson
descriptives) and the re-implementation's comparison accompany this paper as ancillary files (Appendix~\ref{app:ancillary}). The locked
records of the confirmatory run are public on Zenodo (doi:\V{records.doi}): every writer and reader request and record, the worlds,
the completion manifest with its timestamp proof, a flat machine-readable extract of all records and a script that checks the records
against the manifest; the two call ledgers are included with one field, a local file path, replaced by the file's name. Run on these
records, the registered analysis code reproduces the registered output. Appendix~\ref{app:examples} prints four lessons and answers
from them.

\paragraph{Use of language models.} The author used language-model assistants for most of the work: Anthropic's Claude (Claude Opus
5.5, principally through Claude Code) for the literature search, the selection among candidate questions under a documented procedure,
the design, the implementation and execution of the runner, the analysis and audit tooling, the independent re-implementation and the
drafting of this manuscript; and OpenAI's Codex (GPT-6.1 Sol) for critique of the design, the results and this manuscript, alongside
separate Claude instances that audited them. The author also obtained a critique of a draft from OpenAI's ChatGPT; its points were
checked against the records before any change was made. The author set the research programme, approved the study's execution and its spending,
reviewed the outcome and is responsible for the manuscript. No language model is an author. The models studied are experimental
subjects, not tools of the analysis: their choices were scored by the registered \texttt{first\_tool} rule, without a language-model
judge.

\paragraph{Ethics.} The study involved no human participants and no personal data. The models were accessed through OpenRouter under
the providers' terms, on synthetic and benign operational tasks. The single writer refusal was returned by a provider's content filter
on one such task.

\bibliography{refs}

\clearpage
\appendix
\section{Writer instructions}\label{app:writers}

RB is ReasoningBank's pair of instructions \citep{ouyang2025reasoningbank} (Apache-2.0), taken from the authors' released code: the
string constants \texttt{SUCCESSFUL\_SI} and \texttt{FAILED\_SI} of \V{rb.path} in the repository
\url{https://github.com/google-research/reasoning-bank} at commit \V{rb.commit} (file SHA-256 \V{rb.filesha}), fetched on
\V{rb.fetched} and read without executing the file. The text, its provenance record and the licence are in the deposited package.

The released instructions differ from those printed in the ReasoningBank paper (arXiv version \lit{2}, Appendix \lit{A.1}) in three
places, the same in both instructions except where noted:
\begin{itemize}
\item the printed note ``Do not mention specific websites, queries, or string contents, but rather focus on the generalizable
  insights.'' reads in the code ``Prefer concrete, actionable procedures over abstract principles. Do not embed specific product
  names, queries, or literal string contents from the task.'', with ``recovery procedures'' in the failure instruction;
\item the printed description field, ``one sentence summary of the memory item'', reads ``one sentence summary describing when or
  when NOT to use the memory item'';
\item the printed content field, ``\lit{1-3} sentences describing the insights learned to successfully accomplishing the task'',
  ends ``\ldots{} to successfully accomplishing similar tasks in the future'' in the success instruction and ``\ldots{} to avoid such
  failures and successfully accomplishing similar tasks in the future'' in the failure instruction.
\end{itemize}
This study sent the released text with two declared replacements: ``an expert in web navigation'' became ``an expert in running
operational tasks'', and ``Do not embed specific product names, queries, or literal string contents from the task.'' became ``Do not
embed specific queries or literal string contents from the task, but you may name the tools.'' Neither the printed instructions nor
the released text without these replacements was run. The failure instruction, as sent:
\begin{lstlisting}
You are an expert in running operational tasks. You will be given a user query, the corresponding trajectory that represents **how an agent attempted to resolve the task but failed**. 

## Guidelines
You need to extract and summarize useful insights in the format of memory items based on the agent's failed trajectory.
The goal of summarized memory items is to be helpful and generalizable for future similar tasks.

## Important notes
  - You must first reflect and think why the trajectory failed, and then summarize what lessons you have learned or strategies to prevent the failure in the future.
  - You can extract *at most 3* memory items from the trajectory.
  - You must not repeat similar or overlapping items.
  - Prefer concrete, actionable recovery procedures over abstract principles. Do not embed specific queries or literal string contents from the task, but you may name the tools.

## Output Format
Your output must strictly follow the Markdown format shown below:

```
# Memory Item i
## Title <the title of the memory item>
## Description <one sentence summary describing when or when NOT to use the memory item>
## Content <1-3 sentences describing the insights learned to avoid such failures and successfully accomplishing similar tasks in the future>
```
\end{lstlisting}
RBD adds, as the first bullet under ``Important notes'' of both RB instructions: ``Before writing, judge whether the agent's decision
was sound given only what was known before the outcome. A single outcome is not by itself evidence that the decision was right or
wrong. Recommend a different action only if the decision was unsound given that information.'' N is RB's success instruction with its
four outcome-specific phrases made neutral and the sentence ``If nothing about how the task should be done needs to change, say so.''
added; it is used after both outcomes. Every rendered instruction's SHA-256 is asserted by the package's tests.

\section{Reader prompt}\label{app:reader}

The reader's system message:
\begin{lstlisting}
You are an operations agent that runs the same tasks regularly. Before each run you read the memory items that were saved from earlier runs of the task.
\end{lstlisting}
An example user message (a sample world from the package's rendering samples, not a confirmatory world; good decision, failure,
the lesson written from the failure):
\begin{lstlisting}
Task: Export the nightly sales records to the reporting bucket.

Tools available for this task (both operational today):
- export_via_ellery: exports the night's sales records to the reporting bucket through the Ellery route
- export_via_zanric: exports the night's sales records to the reporting bucket through the Zanric route

Run log (authoritative; stable for months; last 240 runs of each tool): export_via_ellery succeeded in 202 of 240; export_via_zanric succeeded in 145 of 240. There have been no outages: failures of either tool are independent one-off events, and the next run has always started normally. Both tools cost the same to call, and a failed run has to be re-run by hand.

Previous run of this task: export_via_ellery was used and it failed ("ERROR 503: upstream temporarily unavailable"). This is a new scheduled run of the task. The previous run is finished and will not be repeated.

Memory items saved from earlier runs of this task:
<the writer's whole response to the good-decision FAILURE episode, e.g. # Memory Item 1 ...>

Which tool will you call first for this run? Reply in JSON with the tool's name and a one-sentence reason.
\end{lstlisting}

\section{Example lessons and answers}\label{app:examples}

Four examples from the locked records of the confirmatory run, all from the good decision's failure run (a chance failure of the
better tool), chosen by rules fixed before the records were searched and applied in world-identifier order. Each of A--C is the
first world in which the reader with the matched lesson named the worse tool while the reader with the irrelevant item named the
better tool: A at the endpoint with the largest P2 (\V{ex.a.name}, RB); B at \V{ex.b.name}, among RB lessons that do not name the
worse tool by the registered descriptive's rule (its full name or its coined part); C at the one endpoint that carried the chance and
for which neither remedy was credited (\V{ex.c.name}, RBD). D is the \V{ex.d.name} lesson quoted in Section~\ref{sec:what}; both of its
readers, with the lesson and with the irrelevant item, named the better tool. Each lesson is printed as its reader received it (its
SHA-256 equals the memory digest recorded with the reader's answer), with non-ASCII punctuation rendered in ASCII; an answer is its
\texttt{first\_tool} and \texttt{reason} fields. The examples illustrate single cases; they do not show how often lessons take these
forms. In A--C the lesson recommends switching to the other tool after an error within a run, and the reader applied this to the next
run's first choice.

\needspace{12\baselineskip}
\subsection*{A: \V{ex.a.name}, world \V{ex.a.world}}
\begin{lstlisting}
World w20261009_0004 (backup; cue: explicit). Better tool backup_via_ellery, worse tool backup_via_aldous.
Endpoint qwen/qwen3.8-max-0902. Good decision, chance failure. Lesson (RB, written from the failure):

# Memory Item 1
## Title Fall back to the other backup tool when the first backup tool fails
## Description Use when a backup operation using backup_via_ellery or backup_via_aldous errors and the task still allows another action.
## Content A failure from one backup tool should be treated as tool-specific, not as final task failure. Immediately try the alternate backup tool in the same run before calling finish, especially when read_run_log indicates failures are independent and costs are equal.

# Memory Item 2
## Title Do not finish until backup success is confirmed or recovery options are exhausted
## Description Use before calling finish after any tool error in a one-shot operational run.
## Content Verify that the backup call returned a successful completion rather than an error. If it failed, pursue available recovery steps such as an alternate tool, and only finish unsuccessfully when no further permitted action remains.

Reader with this lesson: first_tool = backup_via_aldous (the worse tool).
  Reason: backup_via_ellery just failed and the stored guidance says to fall back to the alternate backup tool immediately, with both tools costing the same and failures independent.
Reader with the irrelevant item: first_tool = backup_via_ellery (the better tool).
  Reason: The previous failure does not affect this independent run, and backup_via_ellery has a much higher recent success rate (309/365 vs. 196/365).
\end{lstlisting}

\needspace{12\baselineskip}
\subsection*{B: \V{ex.b.name}, world \V{ex.b.world}}
\begin{lstlisting}
World w20261009_0018 (ci; cue: explicit). Better tool build_via_verran, worse tool build_via_kestle.
Endpoint x-ai/grok-4.7. Good decision, chance failure. Lesson (RB, written from the failure; it does not name the worse tool):

# Memory Item 1
## Title Fail over after scheduler eviction
## Description Use when a build or job tool fails because its node was drained or the scheduler evicted the job, and an equivalent alternate tool exists.
## Content Treat node-drain and scheduler-eviction errors as path-specific infrastructure failures, not as proof the task cannot run. Immediately call a different equivalent builder or executor instead of finishing; do not retry only the same drained path if the error says that invocation cannot be repeated.

# Memory Item 2
## Title Do not pick a tool from long-run averages alone
## Description Use when several tools can perform the same job and a run log shows stable, independent one-off failures with no current outage signal.
## Content Historical success rate is a weak tie-break, not a health check. Before committing, inspect current scheduler or node status; if that is unavailable, be ready to switch tools on the first infrastructure error rather than treating the historically better tool as the only attempt.

# Memory Item 3
## Title Separate manual re-run notes from session constraints
## Description Use when logs say failed jobs are normally re-run by hand, or when an error says the current invocation cannot be retried.
## Content A note that humans re-run failed jobs does not forbid calling another valid tool now. Only stop if every equivalent path has failed or the environment explicitly blocks further attempts; "no retry" applies to the failed invocation, not to the whole task.

Reader with this lesson: first_tool = build_via_kestle (the worse tool).
  Reason: The last Verran run was evicted after a node drain, so start this run on the other equivalent builder instead of retrying that path.
Reader with the irrelevant item: first_tool = build_via_verran (the better tool).
  Reason: Verran has a substantially higher historical success rate (155/180 vs 119/180), and the prior eviction was an independent one-off that does not change the next-run odds.
\end{lstlisting}

\needspace{12\baselineskip}
\subsection*{C: \V{ex.c.name}, world \V{ex.c.world}}
\begin{lstlisting}
World w20261009_0003 (sync; cue: explicit). Better tool sync_via_dunmore, worse tool sync_via_talvo.
Endpoint mistralai/mistral-large-4-0. Good decision, chance failure. Lesson (RBD, written from the failure with the decision-quality note):

# Memory Item 1
## Title Execute alternative tools immediately when the primary tool fails
## Description When a selected tool returns an error and the task remains incomplete, invoke other available tools that achieve the same goal before finishing.
## Content If a database sync operation fails (regardless of whether the specific tool can be retried), immediately attempt the sync using an alternative available tool. The failure of one tool does not preclude task completion via another tool, even if the first tool has higher historical reliability.

# Memory Item 2
## Title Treat lock wait timeout errors as signals for immediate tool switching
## Description When sync operations fail with lock wait timeout errors, switch to an alternative sync tool immediately rather than stopping or waiting.
## Content Lock wait timeout errors indicate transient database resource contention specific to the current connection or transaction. When encountered, immediately retry the operation via an alternative sync tool, as the lock may be specific to the failed tool's execution path while the alternative tool may succeed.

# Memory Item 3
## Title Do not interpret single-tool retry limits as task completion limits
## Description When a tool reports it cannot be retried, verify whether alternative tools remain available before concluding the task has failed.
## Content Constraints preventing retry of a specific failed tool invocation do not prevent using different tools. Always check for and attempt alternative approaches before finishing a task, even when the primary tool indicates no retries are allowed.

Reader with this lesson: first_tool = sync_via_talvo (the worse tool).
  Reason: The previous run failed with a lock wait timeout on sync_via_dunmore, and per Memory Item 2, lock wait timeout errors are signals for immediate tool switching to the alternative connector rather than retrying the failed tool.
Reader with the irrelevant item: first_tool = sync_via_dunmore (the better tool).
  Reason: Dunmore has the higher historical success rate (302/365 vs Talvo's 214/365), failures are independent one-off events so the prior Dunmore failure doesn't change the odds, and equal call cost means the more reliable connector minimizes expected manual re-runs.
\end{lstlisting}

\needspace{12\baselineskip}
\subsection*{D: \V{ex.d.name}, world \V{ex.d.world}}
\begin{lstlisting}
World w20261009_0102 (shipping; cue: implicit). Better tool ship_via_ulvar, worse tool ship_via_calder.
Endpoint anthropic/claude-opus-5.5. Good decision, chance failure. Lesson (RB, written from the failure):

# Memory Item 1
## Title Keep the higher-success tool when the stats are stable and costs are equal, even after it fails
## Description Use this when choosing between equivalent tools using a large, stable run history, especially after a single failure tempts you to switch or distrust the method.
## Content The choice of ship_via_ulvar was sound: it succeeded in about 88% of runs against about 60% for the other tool, over 240 runs each, at the same cost. One failure is expected variance (roughly 1 in 8 runs) and does not show the selection was wrong. In future runs, keep picking the tool with the higher well-measured success rate rather than switching because of one bad outcome.

# Memory Item 2
## Title Check the real outcome when an error says "no confirmation received"
## Description Use this when a booking or transactional tool returns a lost-request or missing-confirmation error rather than a clear rejection.
## Content A "no confirmation" error means the result is unknown, not that the action definitely failed. The booking may have gone through even though the reply was lost. Before finishing, use any available read-only status, lookup, or log tool to check whether the booking exists. Then report it as confirmed, absent, or unknown, so a manual re-run does not create a duplicate pickup.

# Memory Item 3
## Title When retries are not allowed, finish with a clear failure report for a manual re-run
## Description Use this when a run cannot retry after a failed action and a person must re-run the task by hand.
## Content Do not end with a bare finish(). State clearly that the task was not completed and include:
- the tool used
- the exact error text
- whether the booking might still have gone through
- a recommended next step, such as re-running with the same high-success tool after confirming no duplicate booking exists

This gives the person doing the manual re-run what they need to fix it quickly and safely.

Reader with this lesson: first_tool = ship_via_ulvar (the better tool).
  Reason: Ulvar has the higher, well-measured success rate over a large, stable history (210/240, about 88%, versus 145/240, about 60%) at equal cost, so one failed run in the last attempt is ordinary variance and not a reason to switch.
Reader with the irrelevant item: first_tool = ship_via_ulvar (the better tool).
  Reason: The authoritative, months-long run log shows Ulvar succeeding far more often (210/240, about 88%) than Calder (145/240, about 60%) at the same cost, so a single failure last run is an expected occasional miss rather than a reason to switch.
\end{lstlisting}
\FloatBarrier

\section{Panel}\label{app:panel}

\begin{table}[!htbp]
\centering\small
\caption{The panel: OpenRouter model identifiers and the pinned provider (no fallbacks).}
\begin{tabular}{ll>{\raggedright\arraybackslash}p{5.2cm}}
\toprule
endpoint & model identifier & provider\\
\midrule
Claude Opus 5.5 & \texttt{anthropic/claude-opus-5.5} & Anthropic \\
Claude Sonnet 5.5 & \texttt{anthropic/claude-sonnet-5.5} & Anthropic \\
Claude Haiku 5.5 & \texttt{anthropic/claude-haiku-5.5} & Anthropic \\
GPT-6.1 Sol & \texttt{openai/gpt-6.1-sol} & OpenAI \\
GPT-6 Luna & \texttt{openai/gpt-6-luna} & OpenAI \\
Gemini 3.8 Flash & \texttt{google/gemini-3.8-flash} & Google AI Studio \\
Grok 4.7 & \texttt{x-ai/grok-4.7} & xAI \\
Qwen 3.8 Max & \texttt{qwen/qwen3.8-max-0902} & Alibaba \\
DeepSeek V4 Pro & \texttt{deepseek/deepseek-v4-pro-0813} & Alibaba (DeepSeek's own endpoint does not serve structured outputs) \\
Mistral Large 4 & \texttt{mistralai/mistral-large-4-0} & Mistral \\
Kimi K3 & \texttt{moonshotai/kimi-k3} & Moonshot AI (moonshotai/mxfp4) \\
 
\bottomrule
\end{tabular}
\end{table}
\FloatBarrier

\section{Qualifiers and planted increment by endpoint}\label{app:qualifiers}

\begin{table}[!htbp]
\centering\footnotesize\setlength{\tabcolsep}{3pt}
\caption{Good decision, per endpoint (exploratory): complete worlds ($n$) of P2, the qualifiers in the failure cell and the
lesson-source effect in the benign cell, with registered 95\% intervals.}
\begin{tabular}{lrlll}
\toprule
endpoint & $n$ & A2 & L2 & L2b\\
\midrule
Claude Opus 5.5 & 255 & $0.000$ [$-0.014$, $+0.014$] & $0.000$ [$-0.014$, $+0.014$] & $0.000$ [$-0.014$, $+0.014$] \\
Claude Sonnet 5.5 & 256 & $0.000$ [$-0.014$, $+0.014$] & $0.000$ [$-0.014$, $+0.014$] & $0.000$ [$-0.014$, $+0.014$] \\
Claude Haiku 5.5 & 256 & $+0.051$ [$+0.024$, $+0.078$] & $+0.043$ [$+0.014$, $+0.072$] & $-0.004$ [$-0.012$, $+0.004$] \\
GPT-6.1 Sol & 256 & $0.000$ [$-0.014$, $+0.014$] & $0.000$ [$-0.014$, $+0.014$] & $0.000$ [$-0.014$, $+0.014$] \\
GPT-6 Luna & 256 & $+0.031$ [$+0.010$, $+0.053$] & $+0.031$ [$+0.010$, $+0.053$] & $0.000$ [$-0.014$, $+0.014$] \\
Gemini 3.8 Flash & 256 & $+0.125$ [$+0.084$, $+0.166$] & $+0.125$ [$+0.084$, $+0.166$] & $0.000$ [$-0.014$, $+0.014$] \\
Grok 4.7 & 256 & $+0.105$ [$+0.068$, $+0.143$] & $+0.105$ [$+0.068$, $+0.143$] & $+0.004$ [$-0.004$, $+0.012$] \\
Qwen 3.8 Max & 256 & $+0.348$ [$+0.283$, $+0.412$] & $+0.355$ [$+0.293$, $+0.418$] & $+0.004$ [$-0.009$, $+0.017$] \\
DeepSeek V4 Pro & 256 & $+0.277$ [$+0.222$, $+0.333$] & $+0.277$ [$+0.222$, $+0.333$] & $0.000$ [$-0.014$, $+0.014$] \\
Mistral Large 4 & 253 & $+0.169$ [$+0.123$, $+0.216$] & $+0.169$ [$+0.123$, $+0.216$] & $0.000$ [$-0.014$, $+0.014$] \\
Kimi K3 & 256 & $+0.125$ [$+0.084$, $+0.166$] & $+0.125$ [$+0.084$, $+0.166$] & $+0.008$ [$-0.003$, $+0.019$] \\
 
\bottomrule
\end{tabular}
\end{table}

\begin{table}[!htbp]
\centering\footnotesize\setlength{\tabcolsep}{3pt}
\caption{Bad decision and planted increment, per endpoint (exploratory): complete worlds ($n$) of P3m, the qualifiers in the success
cell, and Sus with its reading (``above'' or ``below'' \lit{0.10}), with registered 95\% intervals.}
\begin{tabular}{lrllll}
\toprule
endpoint & $n$ & A3 & L3 & Sus & reading\\
\midrule
Claude Opus 5.5 & 256 & $0.000$ [$-0.014$, $+0.014$] & $0.000$ [$-0.014$, $+0.014$] & $0.000$ [$-0.014$, $+0.014$] & \emph{below 0.10} \\
Claude Sonnet 5.5 & 256 & $0.000$ [$-0.014$, $+0.014$] & $0.000$ [$-0.014$, $+0.014$] & $+0.001$ [$-0.001$, $+0.003$] & \emph{below 0.10} \\
Claude Haiku 5.5 & 256 & $0.000$ [$-0.014$, $+0.014$] & $0.000$ [$-0.014$, $+0.014$] & $0.000$ [$-0.014$, $+0.014$] & \emph{below 0.10} \\
GPT-6.1 Sol & 256 & $0.000$ [$-0.014$, $+0.014$] & $0.000$ [$-0.014$, $+0.014$] & $0.000$ [$-0.014$, $+0.014$] & \emph{below 0.10} \\
GPT-6 Luna & 256 & $0.000$ [$-0.014$, $+0.014$] & $0.000$ [$-0.014$, $+0.014$] & $+0.004$ [$0.000$, $+0.008$] & \emph{below 0.10} \\
Gemini 3.8 Flash & 256 & $0.000$ [$-0.014$, $+0.014$] & $0.000$ [$-0.014$, $+0.014$] & $+0.546$ [$+0.514$, $+0.578$] & \emph{above 0.10} \\
Grok 4.7 & 256 & $0.000$ [$-0.014$, $+0.014$] & $0.000$ [$-0.014$, $+0.014$] & $+0.004$ [$0.000$, $+0.008$] & \emph{below 0.10} \\
Qwen 3.8 Max & 255 & $-0.020$ [$-0.043$, $+0.003$] & $+0.012$ [$-0.002$, $+0.025$] & $+0.514$ [$+0.480$, $+0.548$] & \emph{above 0.10} \\
DeepSeek V4 Pro & 256 & $+0.031$ [$+0.010$, $+0.053$] & $+0.031$ [$+0.010$, $+0.053$] & $+0.210$ [$+0.184$, $+0.236$] & \emph{above 0.10} \\
Mistral Large 4 & 255 & $0.000$ [$-0.014$, $+0.014$] & $0.000$ [$-0.014$, $+0.014$] & $+0.232$ [$+0.206$, $+0.258$] & \emph{above 0.10} \\
Kimi K3 & 256 & $0.000$ [$-0.014$, $+0.014$] & $0.000$ [$-0.014$, $+0.014$] & $0.000$ [$-0.014$, $+0.014$] & \emph{below 0.10} \\
 
\bottomrule
\end{tabular}
\end{table}
\FloatBarrier

\section{Secondary quantities}\label{app:secondary}

\begin{table}[!htbp]
\centering\small
\caption{Secondary panel quantities (registered; reported with their 95\% intervals, without multiplicity adjustment or reading words).
T1 and T3 are the lesson arm's outcome dependence alone; L2b and L3b the lesson-source effect in the benign cell.}
\begin{tabular}{lrl}
\toprule
quantity & $n$ & estimate [95\% interval]\\
\midrule
P2 under RBD & 256 & $+0.072$ [$+0.061$, $+0.082$] \\
P2 under N & 256 & $+0.027$ [$+0.020$, $+0.034$] \\
A2 under RBD & 256 & $+0.071$ [$+0.060$, $+0.081$] \\
A2 under N & 256 & $+0.027$ [$+0.020$, $+0.034$] \\
L2 under RBD & 256 & $+0.069$ [$+0.058$, $+0.079$] \\
L2 under N & 256 & $+0.027$ [$+0.020$, $+0.033$] \\
L2b (benign cell) & 256 & $+0.001$ [$-0.001$, $+0.003$] \\
T1 (lesson arm alone) & 256 & $+0.115$ [$+0.101$, $+0.130$] \\
P3m under RBD & 256 & $-0.002$ [$-0.005$, $0.000$] \\
P3m under N & 256 & $-0.001$ [$-0.003$, $+0.001$] \\
A3 under RBD & 256 & $-0.002$ [$-0.005$, $0.000$] \\
A3 under N & 256 & $-0.002$ [$-0.004$, $0.000$] \\
L3 under RBD & 256 & $0.000$ [$0.000$, $+0.001$] \\
L3 under N & 256 & $+0.001$ [$0.000$, $+0.002$] \\
L3b (benign cell) & 256 & $+0.004$ [$+0.002$, $+0.006$] \\
T3 (lesson arm alone) & 256 & $+0.004$ [$+0.002$, $+0.006$] \\
remedy RBD$-$RB: P2 & 256 & $-0.039$ [$-0.053$, $-0.026$] \\
remedy RBD$-$RB: A2 & 256 & $-0.041$ [$-0.054$, $-0.027$] \\
remedy RBD$-$RB: L2 & 256 & $-0.043$ [$-0.056$, $-0.029$] \\
remedy N$-$RB: P2 & 256 & $-0.085$ [$-0.097$, $-0.073$] \\
remedy N$-$RB: A2 & 256 & $-0.086$ [$-0.098$, $-0.074$] \\
remedy N$-$RB: L2 & 256 & $-0.086$ [$-0.098$, $-0.073$] \\
remedy RBD$-$RB: P3m & 256 & $-0.004$ [$-0.007$, $-0.002$] \\
remedy RBD$-$RB: A3 & 256 & $-0.004$ [$-0.006$, $-0.001$] \\
remedy N$-$RB: P3m & 256 & $-0.003$ [$-0.005$, $0.000$] \\
remedy N$-$RB: A3 & 256 & $-0.003$ [$-0.005$, $0.000$] \\
P2, explicit worlds & 128 & $+0.042$ [$+0.031$, $+0.053$] \\
P2, implicit worlds & 128 & $+0.181$ [$+0.160$, $+0.203$] \\
L2, explicit worlds & 128 & $+0.043$ [$+0.032$, $+0.053$] \\
L2, implicit worlds & 128 & $+0.181$ [$+0.161$, $+0.202$] \\
P3m, explicit worlds & 128 & $+0.003$ [$-0.002$, $+0.008$] \\
P3m, implicit worlds & 128 & $+0.001$ [$-0.002$, $+0.004$] \\
L3, explicit worlds & 128 & $+0.005$ [$+0.001$, $+0.009$] \\
L3, implicit worlds & 128 & $+0.003$ [$0.000$, $+0.006$] \\
 
\bottomrule
\end{tabular}
\end{table}
\FloatBarrier

\section{Worse-tool counts by endpoint}\label{app:counts}

\begin{table}[!htbp]
\centering\scriptsize
\caption{Good decision: worse-tool choices / counted answers, per endpoint, previous-run outcome and memory arm.}
\begin{tabular}{llrrrrrrrr}
\toprule
endpoint & previous run & RB & RBx & RBD & RBDx & N & Nx & NEUTRAL & PLANTED\\
\midrule
Claude Opus 5.5 & success & 0/256 & 0/255 & 0/256 & 0/256 & 0/256 & 0/256 & 0/256 & 0/256 \\
 & failure & 0/255 & 0/256 & 0/256 & 0/256 & 0/256 & 0/256 & 0/256 & 0/256 \\
\addlinespace[2pt]
Claude Sonnet 5.5 & success & 0/256 & 0/256 & 0/256 & 0/256 & 0/256 & 0/256 & 0/256 & 0/256 \\
 & failure & 0/256 & 0/256 & 0/256 & 0/256 & 0/256 & 0/256 & 0/256 & 1/256 \\
\addlinespace[2pt]
Claude Haiku 5.5 & success & 1/256 & 0/256 & 0/256 & 1/256 & 0/256 & 0/256 & 0/256 & 0/256 \\
 & failure & 13/256 & 2/256 & 5/256 & 4/256 & 4/256 & 3/256 & 0/256 & 0/256 \\
\addlinespace[2pt]
GPT-6.1 Sol & success & 0/256 & 0/256 & 0/256 & 0/256 & 0/256 & 0/256 & 0/256 & 0/256 \\
 & failure & 0/256 & 0/256 & 0/256 & 0/256 & 0/256 & 0/256 & 0/256 & 0/256 \\
\addlinespace[2pt]
GPT-6 Luna & success & 0/256 & 0/256 & 0/256 & 0/256 & 0/256 & 0/256 & 0/256 & 0/256 \\
 & failure & 8/256 & 0/256 & 1/256 & 0/256 & 0/256 & 0/256 & 0/256 & 4/256 \\
\addlinespace[2pt]
Gemini 3.8 Flash & success & 0/256 & 0/256 & 0/256 & 0/255 & 0/256 & 0/256 & 0/256 & 214/256 \\
 & failure & 32/256 & 0/256 & 0/255 & 0/256 & 1/256 & 0/256 & 0/256 & 106/256 \\
\addlinespace[2pt]
Grok 4.7 & success & 0/256 & 1/256 & 0/256 & 0/256 & 0/256 & 1/256 & 0/256 & 0/256 \\
 & failure & 27/256 & 0/256 & 5/256 & 0/256 & 16/256 & 1/256 & 0/256 & 2/256 \\
\addlinespace[2pt]
Qwen 3.8 Max & success & 1/256 & 2/256 & 0/256 & 3/255 & 0/255 & 1/256 & 2/256 & 104/254 \\
 & failure & 100/256 & 9/256 & 84/254 & 13/256 & 17/256 & 7/256 & 11/256 & 130/255 \\
\addlinespace[2pt]
DeepSeek V4 Pro & success & 0/256 & 0/256 & 0/256 & 0/256 & 0/256 & 1/256 & 0/256 & 23/256 \\
 & failure & 71/256 & 0/256 & 41/256 & 0/256 & 8/256 & 0/256 & 0/256 & 128/256 \\
\addlinespace[2pt]
Mistral Large 4 & success & 0/256 & 0/256 & 0/255 & 0/256 & 0/256 & 1/256 & 0/255 & 80/256 \\
 & failure & 43/254 & 0/256 & 47/255 & 0/256 & 39/256 & 0/256 & 0/256 & 124/256 \\
\addlinespace[2pt]
Kimi K3 & success & 0/256 & 2/256 & 0/256 & 0/256 & 0/256 & 0/256 & 0/256 & 0/256 \\
 & failure & 32/256 & 0/256 & 27/256 & 0/256 & 1/256 & 0/255 & 0/256 & 0/256 \\
 
\bottomrule
\end{tabular}
\end{table}

\begin{table}[!htbp]
\centering\scriptsize
\caption{Bad decision: worse-tool choices / counted answers, per endpoint, previous-run outcome and memory arm.}
\begin{tabular}{llrrrrrrrr}
\toprule
endpoint & previous run & RB & RBx & RBD & RBDx & N & Nx & NEUTRAL & PLANTED\\
\midrule
Claude Opus 5.5 & success & 0/256 & 0/256 & 0/256 & 0/256 & 0/256 & 0/256 & 0/256 & 0/256 \\
 & failure & 0/256 & 0/256 & 0/256 & 0/256 & 0/256 & 0/256 & 0/256 & 0/256 \\
\addlinespace[2pt]
Claude Sonnet 5.5 & success & 0/256 & 0/256 & 0/256 & 0/256 & 0/256 & 0/256 & 0/256 & 0/256 \\
 & failure & 0/256 & 0/256 & 0/256 & 0/256 & 0/256 & 0/256 & 0/256 & 0/256 \\
\addlinespace[2pt]
Claude Haiku 5.5 & success & 0/256 & 0/256 & 0/256 & 0/256 & 0/256 & 0/256 & 0/256 & 0/256 \\
 & failure & 0/256 & 0/256 & 0/256 & 0/256 & 0/256 & 0/256 & 0/256 & 0/256 \\
\addlinespace[2pt]
GPT-6.1 Sol & success & 0/256 & 0/256 & 0/256 & 0/256 & 0/256 & 0/256 & 0/256 & 0/256 \\
 & failure & 0/256 & 0/256 & 0/256 & 0/256 & 0/256 & 0/256 & 0/256 & 0/256 \\
\addlinespace[2pt]
GPT-6 Luna & success & 0/256 & 0/256 & 0/256 & 0/256 & 0/256 & 0/256 & 0/256 & 0/256 \\
 & failure & 0/256 & 0/256 & 0/256 & 0/256 & 0/256 & 0/256 & 0/256 & 0/256 \\
\addlinespace[2pt]
Gemini 3.8 Flash & success & 0/256 & 0/256 & 0/256 & 0/256 & 0/256 & 0/256 & 0/256 & 75/256 \\
 & failure & 0/256 & 0/256 & 0/256 & 0/256 & 0/256 & 0/256 & 0/256 & 164/256 \\
\addlinespace[2pt]
Grok 4.7 & success & 0/256 & 0/256 & 0/256 & 0/256 & 0/256 & 0/256 & 0/256 & 1/256 \\
 & failure & 0/256 & 0/256 & 0/256 & 0/256 & 0/256 & 0/256 & 0/256 & 1/256 \\
\addlinespace[2pt]
Qwen 3.8 Max & success & 3/256 & 0/255 & 1/256 & 0/255 & 0/256 & 0/256 & 8/255 & 181/255 \\
 & failure & 0/256 & 6/256 & 2/256 & 0/256 & 0/256 & 1/256 & 2/256 & 131/255 \\
\addlinespace[2pt]
DeepSeek V4 Pro & success & 8/256 & 0/256 & 0/256 & 0/256 & 2/256 & 0/256 & 0/256 & 53/256 \\
 & failure & 0/256 & 5/256 & 0/256 & 0/256 & 0/256 & 1/256 & 0/256 & 11/256 \\
\addlinespace[2pt]
Mistral Large 4 & success & 0/255 & 0/256 & 0/256 & 0/256 & 1/256 & 0/255 & 0/256 & 27/255 \\
 & failure & 0/256 & 0/255 & 0/256 & 0/255 & 0/256 & 0/256 & 0/255 & 8/256 \\
\addlinespace[2pt]
Kimi K3 & success & 0/256 & 0/256 & 0/256 & 0/256 & 0/256 & 0/256 & 0/256 & 0/256 \\
 & failure & 0/256 & 0/256 & 0/256 & 0/256 & 0/256 & 0/256 & 0/256 & 0/256 \\
 
\bottomrule
\end{tabular}
\end{table}
\FloatBarrier

\section{Ancillary files}\label{app:ancillary}

The arXiv record carries three ancillary files, byte copies of files in the project's repository:
\begin{itemize}
\item \texttt{RESULTS\_registered.json}: the registered analysis's output (every contrast per endpoint and for the panel, with
  intervals, $p$-values, readings, flags, rates, missingness, regret and lesson descriptives);
\item \texttt{registered\_supplement.md}: the generated supplement written from that output and the locked records (the panel's
  availability weighting, endpoint readings with $n$, flags and the qualifiers' missing-data bounds, secondary intervals, per-endpoint
  cell rates with Wilson intervals, anomalies with the providers' finish reasons, regret, lesson descriptives and bootstrap
  intervals);
\item \texttt{REIMPL\_COMPARISON.json}: the comparison of the independent re-implementation with the registered analysis.
\end{itemize}

\end{document}